\documentclass{article}
\usepackage[T1]{fontenc}

\usepackage{preprint,times}

\usepackage{xcolor}
\usepackage{microtype}
\usepackage{amsmath,amssymb,amsthm,mathtools}

\usepackage{graphicx}
\usepackage{booktabs}
\usepackage{tabularx}
\usepackage{array}
\usepackage{algorithm}
\usepackage{algpseudocode}
\usepackage{placeins}
\usepackage{tikz}
\usepackage{bm}
\usepackage{wrapfig}
\usepackage{needspace}
\usetikzlibrary{arrows.meta,positioning,fit}
\usepackage{xspace}

\definecolor{accent}{HTML}{245B78}

\usepackage{enumitem}
\setlist[itemize]{leftmargin=2em}
\setlist[enumerate]{leftmargin=2em}

\newcommand{\R}{\mathbb{R}}
\newcommand{\V}{V}
\newcommand{\Pset}{\mathcal{P}}
\newcommand{\ctx}{\mathrm{ctx}}
\newcommand{\sg}{\operatorname{sg}}
\newcommand{\cat}{\operatorname{cat}}
\newcommand{\1}{\mathbf{1}}

\makeatletter
\@ifundefined{theorem}{\newtheorem{theorem}{Theorem}[section]}{}
\@ifundefined{lemma}{}{}
\@ifundefined{proposition}{}{}
\@ifundefined{corollary}{}{}
\@ifundefined{definition}{}{}
\@ifundefined{remark}{}{}
\@ifundefined{theHALG@line}
  {\newcommand{\theHALG@line}{\thealgorithm.\arabic{ALG@line}}}
  {\renewcommand{\theHALG@line}{\thealgorithm.\arabic{ALG@line}}}
\makeatother

\newcommand{\meanstd}[2]{\ensuremath{#1\,{\scriptstyle\pm #2}}}
\newcommand{\bestmeanstd}[2]{\ensuremath{\mathbf{#1}\,{\scriptstyle\pm\mathbf{#2}}}}
\newcommand{\meanCI}[2]{\shortstack{$#1$\\[-1pt]{\scriptsize$[#2]$}}}

\input{visualizations/setup.tex}

\usepackage{hyperref}
\usepackage{url}

\ExplSyntaxOn
\NewExpandableDocumentCommand{\LTwoGIfDoiUrlTF}{m m m}
  {\str_if_eq:eeTF {\str_range:nnn {#1} {1} {4}} {http} {#2} {#3}}
\ExplSyntaxOff

\makeatletter
\let\LTwoG@NAT@hyper\NAT@hyper@
\renewcommand{\NAT@hyper@}[1]{\mbox{\LTwoG@NAT@hyper{#1}}}

\AddToHook{env/tabular/begin}{\let\NAT@hyper@\LTwoG@NAT@hyper}
\AddToHook{env/tabular*/begin}{\let\NAT@hyper@\LTwoG@NAT@hyper}
\AddToHook{env/tabularx/begin}{\let\NAT@hyper@\LTwoG@NAT@hyper}
\makeatother

\title{Learning Field Reconstruction \\ from Incomplete Data by Globally \\ Correcting Local Estimates}

\author{Renhao Zhong\textsuperscript{1}, Zihan Zhou\textsuperscript{1},
Chiyuan Ma\textsuperscript{1}, Tianshu Yu\thanks{Corresponding author.}\textsuperscript{,1,2}\\
{\normalfont\textsuperscript{1}The Chinese University of Hong Kong, Shenzhen}\\
{\normalfont\textsuperscript{2}Shanghai Artificial Intelligence Laboratory}\\
{\normalfont\texttt{\{renhaozhong, zihanzhou1, chiyuanma\}@link.cuhk.edu.cn}}\\
{\normalfont\texttt{yutianshu@cuhk.edu.cn}}}
\hypersetup{
  pdftitle={Learning Field Reconstruction from Incomplete Data by Globally Correcting Local Estimates},
  pdfauthor={Renhao Zhong, Zihan Zhou, Chiyuan Ma, Tianshu Yu},
  pdfsubject={Preprint},
  pdfkeywords={Incomplete Data, Prior, Diffusion Models}
}

\begin{document}

\maketitle

\begin{abstract}
Reconstructing physical fields from training samples that are always incomplete requires learning spatial structure from fragmented observations.
Existing context--query work establishes how held-out observations provide valid training targets, but this does not make the complete-field distribution identifiable when every training field is incomplete.
With finite data, weak evidence of sharp transitions and localized variations can further favor averaged predictions that attenuate local detail.
A structural prior is therefore needed to favor plausible completions; local spatial relationships offer one grounded in the observations.
We propose a locally constructed, globally revisable estimator that explicitly learns local field estimates and subsequently corrects them using full-domain observations.
A shared coordinate-conditioned predictor learns from incomplete patches, allowing relatively well-observed neighborhoods to provide direct supervision of local structure.
Its overlapping predictions are reconciled into an observation-conditioned consensus field.
A full-domain estimator retains the original observations and learns a residual correction around this frozen field estimate, allowing locally constructed structure to be revised by broader evidence.
The local estimate serves as both an explicit input, accompanied by its discrepancies with the observations, and a prediction starting point that the global model can revise.
On three real-world ocean datasets with authentic observation gaps, our estimator achieves the lowest MSE and highest PSNR on withheld source-supported values, reducing MSE by 28.9\%--34.5\% against the strongest external baseline.
\end{abstract}

\section{Introduction}
\label{sec:introduction}

\paragraph{Learning from never-complete fields.}
Environmental monitoring relies on spatially resolved physical fields, yet satellite observations contain irregular gaps caused by sampling and retrieval constraints \citep{barth2024dindiff,ma2026observation}.
Reconstructing the unobserved regions requires inferring their values from measurements elsewhere in the field and spatial relationships learned across samples \citep{beckers2003dineof,alveraazcarate2005dineof}.
In many practical datasets, however, every recorded field is incomplete: each training example contains only the measurements on its observed support, with no corresponding complete field available as a target \citep{zhou2026incomplete}.
The learning task is therefore to acquire a reconstruction rule from a collection of incomplete fields and apply it to the missing regions of a new, partially observed field.

\paragraph{Limited supervision can obscure local detail.}
Missingness limits both the evidence available for each prediction and the supervision available to learn that prediction.
A model cannot be trained directly against the unknown values in a sample's missing regions; it must instead learn from relationships among the values that are observed together.
Context--query methods such as IDCD and OAMP create such learning examples by withholding some available measurements as targets and using the remaining observations as context \citep{zhou2026incomplete,ma2026observation}.
This provides a training signal, but its coverage of spatial relationships remains limited by the original observations.
The full-domain direct estimator considered here relies on these examples to learn local structure and the use of distant observations jointly.
With extensive missingness and finite data, sharp transitions and localized variations may be sparsely represented in the context--target pairs.
Where the available context leaves several configurations plausible, squared-error prediction favors their conditional average, potentially attenuating local detail \citep{pmlr-v80-lehtinen18a,barth2024dindiff}.
These difficulties motivate giving local structure a dedicated learning role within the reconstruction process.

\Needspace{0.42\textheight}
\begin{wrapfigure}{R}{0.50\textwidth}
    \centering
    \setlength{\abovecaptionskip}{6pt}
    \input{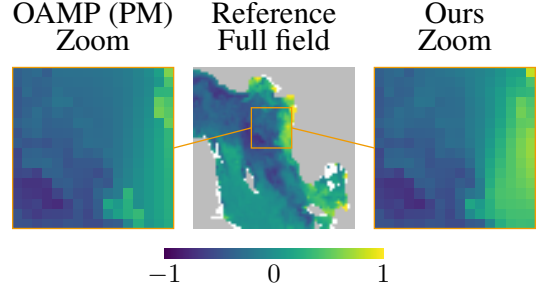}
    \vspace{-15pt}
    \caption{Fine-scale detail in full-mask Baltic Sea NANO reconstruction at $64\times64$. OAMP (PM) is the parameter-matched OAMP control, denoted Full Direct. The yellow box marks the held-out region enlarged in the side panels. Gray denotes outside-domain locations; white in Reference denotes source-missing values.}
    \label{fig:intro-detail}
\end{wrapfigure}

\paragraph{Local learning provides structure for global revision.}
Local views offer a useful starting point: even a heavily incomplete field can contain patches with relatively dense observations, providing more direct supervision of spatial variation within those neighborhoods.
A shared patch predictor can pool these local learning examples across fields, while coordinate conditioning accommodates variation in their spatial statistics \citep{wang2023patch,hu2024padis}.
Our hypothesis is that organizing the available supervision around smaller prediction problems helps finite models learn local relationships explicitly.
Yet local estimates can remain ambiguous, disagree across overlapping views, or miss information outside their support.
This suggests constructing a field estimate from local predictions and then learning its revision from full-domain observations.
The intended division of labor is to establish locally supported detail first, reconcile overlapping predictions, and use broader evidence to correct their errors and inconsistencies.

\paragraph{A consensus prior anchors the full-domain estimator.}
The proposed estimator first trains a coordinate-conditioned patch predictor and freezes it to construct an observation-conditioned local consensus prior---an intermediate field estimate for the current sample.
Overlapping predictions are reconciled by least squares, building on position-aware patch modeling \citep{wang2023patch,hu2024padis} and view reconciliation \citep{bartal2023multidiffusion}.
A full-domain model retains the original observation context and learns a residual correction around this estimate under the same context--query supervision.
Compared with OAMP (PM), our reconstruction preserves more distinct fine-scale variations in the enlarged Baltic Sea NANO region (Figure~\ref{fig:intro-detail}).

\paragraph{Component controls test the role of the prior.}
The reference field and its discrepancies reorganize the retained observations into an explicit representation for full-domain prediction.
Our hypothesis concerns how finite models learn with this local-to-global organization; estimator comparisons and component controls test the contributions of prior conditioning and residual correction.

\paragraph{Results support revisable local estimates.}
Experiments on three real-world ocean products yield the lowest MSE and highest PSNR among the compared methods at $64\times64$, with metrics computed on withheld source-supported values.
On the ocean products, prior-conditioned direct prediction reduces MSE by 8.3\%--9.0\% relative to parameter-matched no-prior direct prediction, and residual correction yields a further 9.8\%--16.2\% reduction.
These comparisons show that a local estimate can be valuable as a revisable intermediate representation even when it is a weaker standalone predictor.

\paragraph{Contributions.} We make three contributions:
\begin{itemize}
    \item We introduce a locally constructed, globally revisable estimator that explicitly learns local field structure from incomplete patches and uses full-domain observations to revise the resulting consensus estimate.
    \item We construct an explicit reference field by overlap consensus and couple it with prior--observation discrepancy features and a residual output anchored at the frozen reference, using only context--query supervision.
    \item We demonstrate reconstruction gains on ocean fields with authentic observation gaps and show, through estimator comparisons and targeted controls, the value of local estimates as revisable intermediate representations.
\end{itemize}

\section{Preliminaries}
\label{sec:preliminaries}

Appendix~\ref{app:related-work} reviews related work on incomplete-data learning, observation-aligned reconstruction, and local--global estimation (Sections~\ref{app:related-incomplete}--\ref{app:related-local-global}). We now define the observation and supervision interfaces used throughout the paper.

\subsection{Incomplete Observations}
\label{sec:supplied-support}

Following the notation of IDCD \citep{zhou2026incomplete}, let $\bm{x}_0$ denote a complete physical field and $\bm{M}$ a binary mask indicating its observed locations.
The available data are $\bm{x}_{\mathrm{obs}}=\bm{M}\odot \bm{x}_0$, where $\odot$ denotes elementwise multiplication.
Each field has $d_x$ channels on a spatial grid $\Omega$; masks apply to all channels, and a validity mask $\bm{V}$ excludes locations outside the physical domain, such as land.
Training uses only incomplete pairs $(\bm{x}_{\mathrm{obs}},\bm{M})$ with $\bm{M}\leq \bm{V}$ and $\bm{M}\neq \bm{V}$.
Each sample's available support stays fixed across training visits.
Our goal is point reconstruction of the valid missing values from these incomplete training pairs.

\subsection{Observation-Aligned Supervision}
\label{sec:context-query-supervision}

\paragraph{Context--query supervision.}
Originally missing values cannot serve as supervision, so we temporarily withhold some available measurements as targets.
Following IDCD and OAMP \citep{zhou2026incomplete,ma2026observation}, we divide the available observations into a context mask $\bm{M}_{\mathrm{ctx}}\leq \bm{M}$ and a nonempty query mask $\bm{M}_{\mathrm{qry}}=\bm{M}\odot(1-\bm{M}_{\mathrm{ctx}})$.
Context values are supplied to the model, while query values are withheld as training targets.
The internal partition varies across visits; the original support $\bm{M}$ remains fixed.

\paragraph{Training context--query construction.}
The training partition respects the acquisition structure: individual sites for pixel observations and whole blocks for block observations \citep{zhou2026incomplete}.
For Black Sea and Baltic, dataset-specific frozen mask-only Bayesian Flow Networks (BFNs; \citealp{graves2023bfn}) generate binary proposals using OAMP's observation-aligned guidance \citep{ma2026observation}; Global Ocean SSS instead uses an empirical authentic-track mask bank. Before feasibility handling, each proposal $\widetilde{\bm{M}}$ is intersected with the available support:
\begin{equation}
\bm{M}_{\mathrm{ctx}}=\bm{M}\odot\widetilde{\bm{M}},
\qquad
\bm{M}_{\mathrm{qry}}=\bm{M}\odot(1-\widetilde{\bm{M}}).
\label{eq:authentic-mask-intersection}
\end{equation}
The constructor selects observations without supplying field values or pseudo-labels. Appendices~\ref{app:aligned-design}--\ref{app:aligned-datasets} specify the partition requirements, dataset-specific constructors, and feasibility handling.

\paragraph{Denoising query prediction.}
At diffusion time $t$, we perturb the observations as
\begin{equation}
\bm{x}_{\mathrm{obs},t}=\bm{M}\odot(\alpha_t \bm{x}_{\mathrm{obs}}+\sigma_t\bm{\epsilon}),
\qquad \bm{\epsilon}\sim\mathcal N(0,I).
\label{eq:observable-state}
\end{equation}
A clean-field predictor $\bm{x}_{\bm{\theta}}$ receives the context values $\bm{M}_{\mathrm{ctx}}\odot \bm{x}_{\mathrm{obs},t}$ and their mask, and is trained by
\begin{equation}
\mathcal L_{\mathrm{qry}}(\bm{\theta})
=\mathbb E\!\left[
\left\|\bm{M}_{\mathrm{qry}}\odot
\bigl(\bm{x}_{\bm{\theta}}(t,\bm{M}_{\mathrm{ctx}}\odot \bm{x}_{\mathrm{obs},t},\bm{M}_{\mathrm{ctx}})-\bm{x}_{\mathrm{obs}}\bigr)
\right\|_2^2\right].
\label{eq:shared-query-loss}
\end{equation}
Here $\bm{\theta}$ denotes the parameters of the chosen predictor; fixed geometry is suppressed in this generic signature.
The query mask selects the loss terms without entering the model; predictions are constrained on states with positive query exposure.
Both stages use this supervision, restricted to patches for local training and applied over the full domain for global training.
Appendices~\ref{app:diffusion-dynamics} and~\ref{app:masked-objectives} specify the diffusion parameterization and query-conditioned target; Section~\ref{sec:method-inference} describes reconstruction from the supplied observations.

\section{Method}
\label{sec:method}
\label{sec:method-overview}
\label{sec:complete-law-identifiability}

Never-complete observations need not identify an unrestricted complete-field distribution (Theorem~\ref{prop:never-complete-nonidentifiability}; Appendix~\ref{app:distribution-identifiability}). We focus on point reconstruction with finite data and model capacity, learning recurring local relationships through shared patch-level tasks and revising their estimates using full-domain observations.

\begin{figure}[t]
\centering
\includegraphics[width=\linewidth]{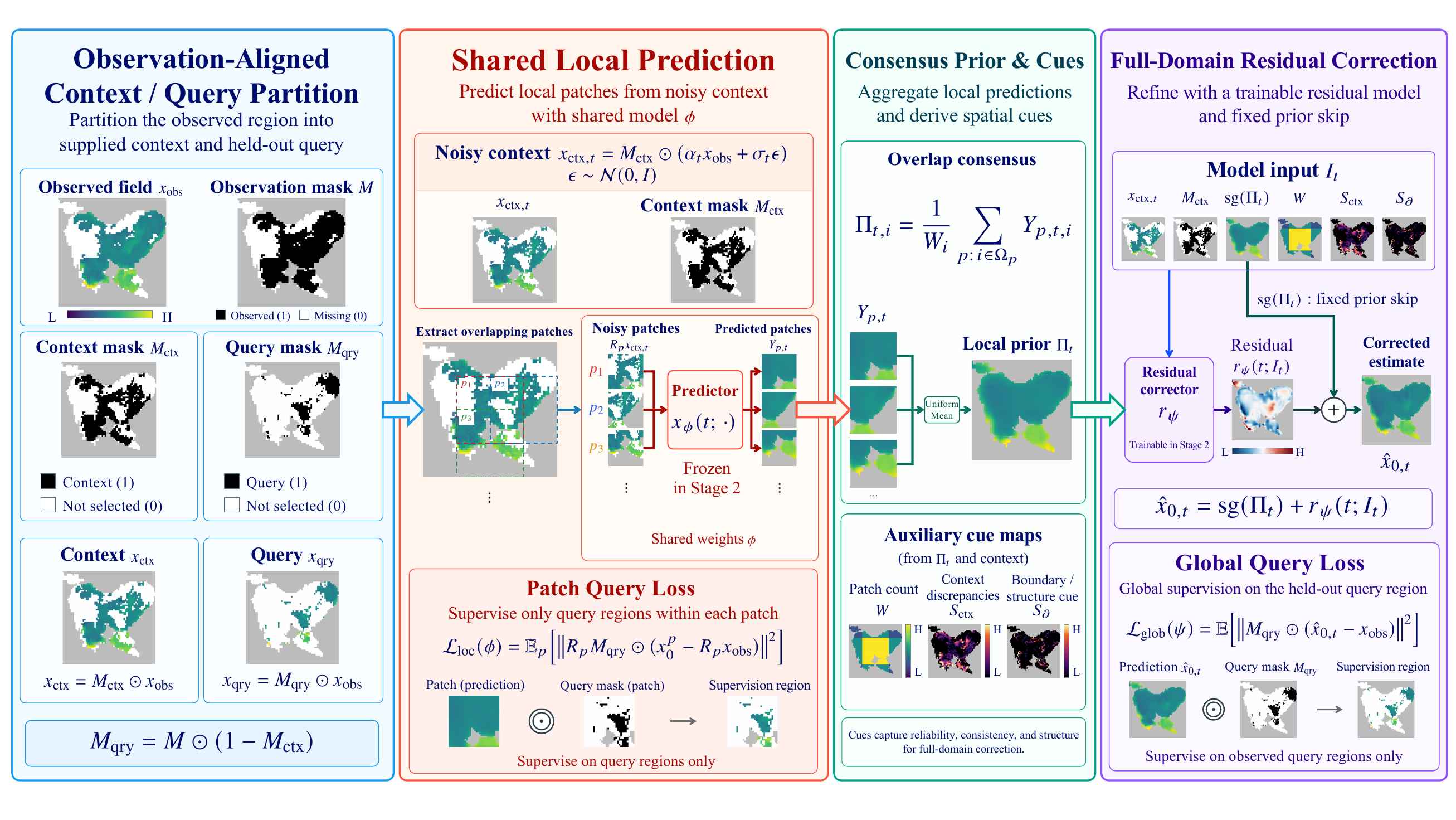}
\caption{Local prior construction and full-domain correction for one context. Overlapping patch predictions form a consensus prior that conditions the corrector and is added to its learned residual. The patch predictor is frozen during global training; query values enter only the training losses.}
\label{fig:method-overview}
\end{figure}

Both stages use the observation-aligned supervision from Section~\ref{sec:preliminaries}. Given noisy context $\bm{x}_{\ctx,t}=\bm{M}_{\mathrm{ctx}}\odot\bm{x}_{\mathrm{obs},t}$, we express reconstruction as a local field prior $\bm{\Pi}_t$ plus a learned full-domain revision:
\begin{equation}
\widehat{\bm{x}}_{0,t}=\sg(\bm{\Pi}_t)+r_{\bm{\psi}}(t;I_t),
\label{eq:prior-anchored-output}
\end{equation}
Here $I_t$ contains the context, prior, and discrepancy features, while $\sg$ denotes stop-gradient through the frozen local predictor.
The prior both conditions the corrector and anchors its output (Figure~\ref{fig:method-overview}).

\subsection{Learning and Constructing a Local Prior}
\label{sec:method-stage1}
\label{sec:coordinate-local-operator}
\label{sec:field-consensus}

The first stage learns to infer withheld values from the measurements within a patch, then assembles these local estimates into a field.
Cropping reorganizes the available supervision into smaller prediction tasks without adding independent observations.

\paragraph{Coordinate-conditioned local learning.}
We share one predictor across patches so that local relationships learned from one neighborhood can help reconstruct another.
Absolute coordinates tell the predictor where each patch lies, allowing its predictions to reflect spatially varying field statistics.
Let $R_p$ crop patch $p$ with support $\Omega_p$ from a fixed covering $\mathcal P$, and let $\bm{\Gamma}$ contain normalized absolute coordinates.
A shared predictor receives the patch's noisy context, mask, and coordinate crop:
\begin{equation}
\widehat{\bm{x}}_0^p
=\bm{x}_{\bm{\phi}}\!\left(t,R_p\bm{x}_{\ctx,t},R_p\bm{M}_{\mathrm{ctx}},R_p\bm{\Gamma}\right).
\label{eq:local-prediction-main}
\end{equation}
This follows coordinate-aware patch modeling \citep{liu2018coordconv,wang2023patch,hu2024padis}.
Within each patch, we ask the predictor to recover the withheld measurements from its supplied context, using the shared query objective restricted to that patch:
\begin{equation}
\mathcal L_{\mathrm{loc}}(\bm{\phi})
=\mathbb E\!\left[
\left\|R_p\bm{M}_{\mathrm{qry}}\odot
\bigl(\widehat{\bm{x}}_0^p-R_p\bm{x}_{\mathrm{obs}}\bigr)\right\|_2^2\right].
\label{eq:stage1-loss}
\end{equation}
For each training item, a patch is sampled uniformly from those with nonempty query support, and only its withheld observed values contribute to the loss.
Each patch prediction uses its own context and coordinates, with no exchange of current-sample information between patch evaluations.
Together, parameter sharing and coordinate conditioning let the model pool local examples without treating all regions as statistically interchangeable.
This gives local variation a dedicated learning task; the full-domain stage can subsequently revise its estimates using evidence outside the patch.

\paragraph{From patch predictions to a reference field.}
After local training, we freeze the trained predictor $\bm{x}_{\bm{\phi}}$ and evaluate it on every patch in the covering, including patches with no query values.
The parameters $\bm{\phi}$ remain fixed throughout global training and reconstruction.
The resulting predictions $\bm{Y}_{p,t}$ may disagree at overlapping locations because they use different neighborhoods.
We reconcile them by averaging all predictions that cover each valid location:
\begin{equation}
\bm{\Pi}_{t,i}=\frac{1}{\bm{W}_i}\sum_{p:i\in\Omega_p}\bm{Y}_{p,t,i},
\qquad
\bm{W}_i=\sum_{p\in\mathcal P}\mathbf1[i\in\Omega_p],\qquad \bm{V}_i=1.
\label{eq:uniform-consensus-solution}
\end{equation}
Every valid location is covered, and $\bm{\Pi}_t$ is set to zero outside the valid domain.
This uniform consensus is the unique solution of an equal-weight least-squares reconciliation problem (Appendix~\ref{app:formal-estimator-definitions}), following the use of overlap reconciliation in regional prediction \citep{bartal2023multidiffusion}.
The resulting field has the same spatial domain, channels, and units as the target, so subsequent correction can operate directly in field space.
We call $\bm{\Pi}_t$ the \emph{local field prior}: an observation-conditioned estimate that provides a reference for reconstruction, distinct from a probability distribution or the mask prior used to select context and query.

Consensus produces a full-sized field, but each value still uses only the neighborhoods that contributed to it.
The next stage lets observations beyond those neighborhoods revise the estimate; Appendix~\ref{app:local-support} formalizes this local-support boundary.

\subsection{Global Residual Correction}
\label{sec:prior-to-global-connection}
\label{sec:information-support-representation}
\label{sec:observable-discrepancies}

The local prior is a starting estimate with two reasons to need revision: distant observations may favor a different completion, and the local predictor may make errors even where nearby evidence is available.
The global stage learns both kinds of revision through Equation~\eqref{eq:prior-anchored-output}.
Its design connects three elements: the evidence available for correction, an output anchored at the prior, and a supervised target for the required change.

\paragraph{Give the corrector the estimate and the evidence.}
We retain the full-domain context so that the corrector can use observations outside any contributing patch, and supply the prior so that it can inspect the current estimate.
We also expose where this estimate disagrees with the available measurements.
The context discrepancy $\bm{S}_{\ctx}$ measures absolute differences between the prior and noisy values at context locations; the boundary discrepancy $\bm{S}_{\partial}$ makes the analogous comparison at adjacent non-context locations using the mean of neighboring noisy context values.
The overlap count $\bm{W}$ records how many patch predictions contribute to each location.
Together these inputs form
\begin{equation}
I_t=\left(\bm{x}_{\ctx,t},\bm{M}_{\mathrm{ctx}},\sg(\bm{\Pi}_t),\bm{W},\bm{S}_{\ctx},\bm{S}_{\partial}\right).
\label{eq:full-domain-input-main}
\end{equation}
These maps make disagreement and construction geometry explicit, helping a finite model learn where and how to revise the field.
The discrepancy maps are observable cues rather than clean error targets, and $\bm{W}$ is not a calibrated uncertainty estimate.
All features use the same realized noisy context, without access to query values; Appendix~\ref{app:implementation-inputs} gives their definitions.

\paragraph{Make the output a revision of the prior.}
Conditioning on the prior tells the corrector what the local stage predicts; adding its output to that prior makes the learning task explicitly one of revision.
We freeze the local predictor and initialize the correction head at zero, so the initial full-field prediction equals $\bm{\Pi}_t$.
Training then learns departures from this context-specific reference, with the correction free to change every predicted value.
At each query location, the required change is simply the withheld measurement minus the prior estimate.
Applying the shared query loss to Equation~\eqref{eq:prior-anchored-output} therefore gives
\begin{equation}
\mathcal L_{\mathrm{glob}}(\bm{\psi})
=\mathbb E\!\left[
\left\|\bm{M}_{\mathrm{qry}}\odot
\left(r_{\bm{\psi}}(t;I_t)-\bigl[\bm{x}_{\mathrm{obs}}-\sg(\bm{\Pi}_t)\bigr]\right)\right\|_2^2\right].
\label{eq:stage2-loss}
\end{equation}
Thus the observed queries directly teach the model how to correct its starting estimate, without requiring complete-field targets.
Appendix~\ref{app:residual-representation} formalizes the two roles of this residual under squared loss: incorporating evidence beyond local support and correcting remaining local-estimation error.

\paragraph{Separating the design choices.}
Section~\ref{sec:exp-estimator-organizations} compares prior conditioning, prior-anchored residual prediction, and alternative references; Section~\ref{sec:exp-coordinates-discrepancies} examines local-prior construction and correction cues. These are choices about finite-model learning: the prior and features are deterministic functions of the retained context, so unrestricted direct and residual predictors have the same optimal query risk (Appendix~\ref{app:residual-representation}).

\subsection{Training and Reconstruction}
\label{sec:method-inference}

\begin{wrapfigure}{R}{0.45\linewidth}
\vspace{-0.7\baselineskip}
\refstepcounter{algorithm}
\label{alg:overall-training-sampling}
\addcontentsline{loa}{algorithm}{\protect\numberline{\thealgorithm}Local prior learning, global correction, and reconstruction.}
\hrule\vspace{0.35\baselineskip}
\textbf{Algorithm \thealgorithm:} Local prior learning, global correction, and reconstruction.
\vspace{0.15\baselineskip}\hrule
\vspace{0.25\baselineskip}
\footnotesize
\begin{algorithmic}[1]
\Statex \textbf{Stage 1: learn the local prior}
\State Sample context--query partitions and perturb contexts.
\State Train the coordinate-conditioned patch predictor with $\mathcal L_{\mathrm{loc}}$; freeze it.
\Statex \textbf{Stage 2: learn global correction}
\State For each sampled context, fuse local predictions into $\bm{\Pi}_t$ and compute $I_t$.
\State Train the zero-initialized residual corrector with $\mathcal L_{\mathrm{glob}}$.
\Statex \textbf{Reconstruction}
\State Draw one near-zero-noise perturbation; select the full supplied mask ($K=1$) or $K$ subcontexts according to the inference mode.
\For{each selected context $k$}
  \State Construct its prior $\bm{\Pi}_\delta^{(k)}$ and features; predict $\bm{\Pi}_\delta^{(k)}+r_{\bm{\psi}}(\delta;I_\delta^{(k)})$.
\EndFor
\State Average the corrected fields and restore supplied observations.
\end{algorithmic}
\vspace{0.15\baselineskip}\hrule
\vspace{-0.5\baselineskip}
\end{wrapfigure}

\paragraph{Two-stage training.}
We first establish the local reconstruction rule by optimizing $\mathcal L_{\mathrm{loc}}$, then freeze it and learn its revision by optimizing $\mathcal L_{\mathrm{glob}}$ for the corrector.
Both stages use the same stored incomplete observations and observation-aligned partition rule.
For each forward pass, the frozen predictor rebuilds the prior from that pass's context.
Query values enter only the losses.
Algorithm~\ref{alg:overall-training-sampling} summarizes the procedure.

\paragraph{Reconstruction from supplied observations.}
Following IDCD's single-step procedure \citep{zhou2026incomplete}, inference uses a fixed near-zero diffusion time $0<\delta\ll1$.
Given $\bm{x}_{\mathrm{in}}=\bm{M}_{\mathrm{in}}\odot\bm{x}_0$, we draw one perturbation $\bm{x}_{\mathrm{in},\delta}=\alpha_\delta\bm{x}_{\mathrm{in}}+\sigma_\delta\bm{\epsilon}$.
Full-mask inference sets $K=1$ and $\bm{M}_{\mathrm{ctx}}^{(1)}=\bm{M}_{\mathrm{in}}$, using all supplied observations rather than a complete reference field.
Context-ensemble inference instead uses $K$ subcontexts sampled within $\bm{M}_{\mathrm{in}}$ by the training partition rule, with the same perturbation shared across members.
For each selected context, we construct its prior and discrepancy features and apply the full-domain correction.
\par\WFclear
Reconstruction combines the resulting predictions and restores the supplied observations:
\begin{equation}
\widetilde{\bm{x}}_\delta=\frac1K\sum_{k=1}^K\widehat{\bm{x}}_{0,\delta}^{(k)},
\qquad
\widehat{\bm{x}}_{\mathrm{out}}=\bm{x}_{\mathrm{in}}+(\bm{V}-\bm{M}_{\mathrm{in}})\odot\widetilde{\bm{x}}_\delta.
\label{eq:proximal-output}
\end{equation}
Validation selects full-mask inference for ocean data and context ensembling for PDE data (Appendix~\ref{app:full-support-inference}). Appendix~\ref{app:complete-algorithms} gives the complete training and reconstruction procedure in Algorithm~\ref{alg:detailed-training-sampling}.

\FloatBarrier

\raggedbottom
\section{Experiments}
\label{sec:experiments}

We evaluate reconstruction under authentic ocean gaps, compare matched estimators and alternative references, and isolate prior construction and corrector inputs. We then intervene outside local support to test whether global correction responds to evidence unavailable to the local predictor.

\subsection{Baselines}
\label{sec:exp-baselines}

On the ocean products, we compare with DINEOF \citep{beckers2003dineof,alveraazcarate2005dineof}, AmbientDiff \citep{daras2023ambient}, DINDiff \citep{barth2024dindiff}, MissDiff \citep{ouyang2023missdiff}, MSM \citep{park2026msm}, and OAMP \citep{ma2026observation}. Controlled PDE comparisons use AmbientDiff, DINDiff, MissDiff, MSM, and IDCD \citep{zhou2026incomplete}. Methods share the available training observations, evaluation items, supplied supports, and scoring coordinates, while external baselines retain their method-specific objectives. Full Direct denotes the parameter-matched OAMP and IDCD controls on ocean and PDE data, respectively (Appendix~\ref{app:retraining-reporting}). Appendix~\ref{app:pde-comparisons} compares MSM's reconstruction mechanism with ours and discusses its behavior under Pixel 30\%.

\subsection{Datasets and Evaluation}
\label{sec:exp-datasets}
\label{sec:exp-protocol}
\label{sec:exp-design}

The ocean products preserve realistic, structured gaps and reveal reference values only where the source product observed them.

\paragraph{Ocean products.}
We use daily Black Sea CHL, Baltic Sea NANO, and Global Ocean SSS observations, retaining their cloud-related and track- or swath-structured gaps.
Fields and masks are jointly resampled to $64\times64$ and $128\times128$, with every retained field remaining incomplete on its valid ocean domain.
Samples are split chronologically into 70\%/15\%/15\% training, validation, and test partitions before resolution-specific filtering.

\paragraph{Controlled PDE fields.}
We additionally evaluate Shallow Water height and Navier--Stokes vorticity under controlled pixel- and block-observation protocols.
Complete PDE references permit evaluation on valid unsupplied coordinates, while training uses the stored incomplete observations.
Appendices~\ref{app:dataset-settings} and~\ref{app:observation-scoring-protocols} give the data preparation and observation protocols.

\paragraph{Evaluation.}
Ocean validation and testing follow OAMP's real-mask overlay protocol \citep{ma2026observation}. A mask from another retained sample of the same product, split, and resolution is intersected with the current source support $\bm{M}_{\mathrm{raw}}$. After feasibility handling, the input and scoring supports are cached and shared across methods (Appendix~\ref{app:observation-scoring-protocols}). Given $\bm{M}_{\mathrm{in}}$, we score only withheld source-supported values:
\begin{equation}
\bm{M}_{\mathrm{eval}}=\bm{M}_{\mathrm{raw}}\odot(1-\bm{M}_{\mathrm{in}}).
\label{eq:evaluation-supports}
\end{equation}
Source-missing ocean locations are not scored. MSE pools squared errors over scored spatial--channel values, whereas PSNR averages sample-level scores; both use the normalized, clipped evaluation space (Appendix~\ref{app:pooled-mse}). Preprocessing statistics are fitted only to available training observations (Appendix~\ref{app:dataset-settings}).

\subsection{Reconstruction Accuracy and Estimator Organization}
\label{sec:exp-results}
\label{sec:exp-reconstruction}
\label{sec:exp-ocean}
\label{sec:exp-estimator-organizations}
\label{sec:direct-residual-parameterizations}
\label{sec:estimator-organizations}
\label{sec:prior-estimator-classes}
\label{sec:exp-resolution}

\paragraph{Main results.}
Our first question is whether local-to-global reconstruction improves accuracy across three authentic ocean products with different observation geometries. At $64\times64$, our estimator achieves the lowest MSE and highest PSNR among the compared methods on all three ocean products (Table~\ref{tab:ocean64}).
MSE decreases by 28.9\%--34.5\% relative to the lowest-MSE external baseline for each product.
Complete comparisons under the controlled PDE observation settings of prior work \citep{zhou2026incomplete} are reported in Tables~\ref{tab:pde64} and~\ref{tab:pde-block89} in Appendix~\ref{app:pde-comparisons}.

\begin{table*}[!t]
\centering
\footnotesize
\caption{Ocean reconstruction at $64\times64$ on withheld source-supported values. Learned-model results are means $\pm$ standard deviations across independent training seeds; DINEOF results are point estimates. Bold marks the best result in each column.}
\label{tab:main-ocean}
\label{tab:ocean64}

\vspace{3pt}
\setlength{\tabcolsep}{1.5pt}
\renewcommand{\arraystretch}{1.15}
\begin{tabular*}{\linewidth}{@{\extracolsep{\fill}}lcccccc@{}}
\toprule
Method & \multicolumn{2}{c}{Black Sea CHL} & \multicolumn{2}{c}{Baltic Sea NANO} & \multicolumn{2}{c}{Global Ocean SSS} \\
\cmidrule(lr){2-3}\cmidrule(lr){4-5}\cmidrule(lr){6-7}
 & MSE$\downarrow$\,($\times10^{-3}$) & PSNR$\uparrow$\,(dB) & MSE$\downarrow$\,($\times10^{-3}$) & PSNR$\uparrow$\,(dB) & MSE$\downarrow$\,($\times10^{-3}$) & PSNR$\uparrow$\,(dB) \\
\midrule
DINEOF & $21.457$ & $16.89$ & $37.398$ & $14.37$ & $44.451$ & $13.68$ \\
AmbientDiff & \meanstd{12.204}{0.156} & \meanstd{19.17}{0.03} & \meanstd{28.304}{0.206} & \meanstd{15.72}{0.03} & \meanstd{47.458}{1.718} & \meanstd{14.07}{0.12} \\
DINDiff & \meanstd{43.830}{1.515} & \meanstd{14.77}{0.09} & \meanstd{53.153}{1.775} & \meanstd{13.22}{0.04} & \meanstd{61.082}{0.960} & \meanstd{12.73}{0.07} \\
MissDiff & \meanstd{25.636}{0.300} & \meanstd{15.92}{0.08} & \meanstd{47.818}{1.704} & \meanstd{13.29}{0.16} & \meanstd{41.935}{0.173} & \meanstd{14.01}{0.03} \\
MSM & \meanstd{15.077}{0.184} & \meanstd{18.27}{0.04} & \meanstd{30.206}{1.170} & \meanstd{15.37}{0.19} & \meanstd{40.901}{0.277} & \meanstd{14.13}{0.03} \\
OAMP & \meanstd{10.263}{0.162} & \meanstd{19.84}{0.10} & \meanstd{27.220}{0.458} & \meanstd{15.60}{0.24} & \meanstd{44.082}{0.472} & \meanstd{14.04}{0.05} \\
\midrule
Ours & \bestmeanstd{7.164}{0.060} & \bestmeanstd{21.79}{0.04} & \bestmeanstd{19.345}{0.741} & \bestmeanstd{17.37}{0.07} & \bestmeanstd{26.798}{0.533} & \bestmeanstd{15.93}{0.05} \\
\bottomrule
\end{tabular*}
\end{table*}

\begin{table*}[!t]
\centering
\footnotesize
\caption{Estimator organization and prior replacement for full-mask ocean reconstruction at $64\times64$. Results are means $\pm$ standard deviations across independent training seeds. Bold marks the best result in each column.}
\label{tab:estimator-ablation-ocean}
\label{tab:matched}

\setlength{\tabcolsep}{0.5pt}
\renewcommand{\arraystretch}{1.05}
\begin{tabular*}{\linewidth}{@{\extracolsep{\fill}}>{\raggedright\arraybackslash}p{0.20\linewidth}cccccc@{}}
\toprule
Estimator & \multicolumn{2}{c}{Black Sea CHL} & \multicolumn{2}{c}{Baltic Sea NANO} & \multicolumn{2}{c}{Global Ocean SSS} \\
\cmidrule(lr){2-3}\cmidrule(lr){4-5}\cmidrule(lr){6-7}
 & MSE$\downarrow$\,($\times10^{-3}$) & PSNR$\uparrow$\,(dB) & MSE$\downarrow$\,($\times10^{-3}$) & PSNR$\uparrow$\,(dB) & MSE$\downarrow$\,($\times10^{-3}$) & PSNR$\uparrow$\,(dB) \\
\midrule
Local Prior & \meanstd{15.227}{0.103} & \meanstd{18.15}{0.03} & \meanstd{30.576}{0.479} & \meanstd{15.17}{0.04} & \meanstd{32.053}{0.276} & \meanstd{15.13}{0.04} \\
Full Direct & \meanstd{9.350}{0.306} & \meanstd{20.32}{0.12} & \meanstd{23.559}{0.834} & \meanstd{16.41}{0.16} & \meanstd{32.895}{0.967} & \meanstd{14.99}{0.15} \\
Low-rank2Global & \meanstd{11.067}{0.243} & \meanstd{19.79}{0.11} & \meanstd{22.841}{0.171} & \meanstd{16.43}{0.04} & \meanstd{30.282}{0.649} & \meanstd{15.46}{0.08} \\
Global2Global & \meanstd{9.175}{0.499} & \meanstd{20.55}{0.16} & \meanstd{25.097}{1.442} & \meanstd{16.23}{0.21} & \meanstd{41.695}{0.670} & \meanstd{14.30}{0.04} \\
Prior Direct & \meanstd{8.546}{0.062} & \meanstd{20.82}{0.03} & \meanstd{21.445}{0.417} & \meanstd{16.89}{0.09} & \meanstd{30.151}{0.197} & \meanstd{15.40}{0.03} \\
\midrule
Prior Residual (Ours) & \bestmeanstd{7.164}{0.060} & \bestmeanstd{21.79}{0.04} & \bestmeanstd{19.345}{0.741} & \bestmeanstd{17.37}{0.07} & \bestmeanstd{26.798}{0.533} & \bestmeanstd{15.93}{0.05} \\
\bottomrule
\end{tabular*}
\end{table*}

\paragraph{Matched estimator organization.}
Table~\ref{tab:matched} compares Local Prior with three full-domain estimators trained on the same incomplete pairs and observation-aligned query supervision. Full Direct predicts from the context and coordinates and matches the total parameter count of Prior Direct and Prior Residual; its training time is comparable to our combined two-stage training time (Appendix~\ref{app:retraining-reporting}). Prior Direct and Prior Residual share the frozen local predictor, full-domain backbone, and augmented input $I_t$. Their zero-output heads give initial full-field predictions of zero and $\bm{\Pi}_t$, respectively, so their comparison evaluates residual parameterization together with its induced initialization (Appendix~\ref{app:implementation-inputs}).

The Local Prior alone has higher MSE than Full Direct on Black Sea CHL and Baltic Sea NANO, yet Prior Direct reduces MSE relative to Full Direct by 8.3\%--9.0\% across the three products. Prior Residual yields a further 9.8\%--16.2\% reduction over Prior Direct. These results distinguish the value of a local estimate as an intermediate representation from its standalone prediction accuracy.

  \begin{wrapfigure}[21]{R}{0.50\textwidth}
  \centering
  \vspace{-0.9\baselineskip}
  \setlength{\abovecaptionskip}{3pt}
  \input{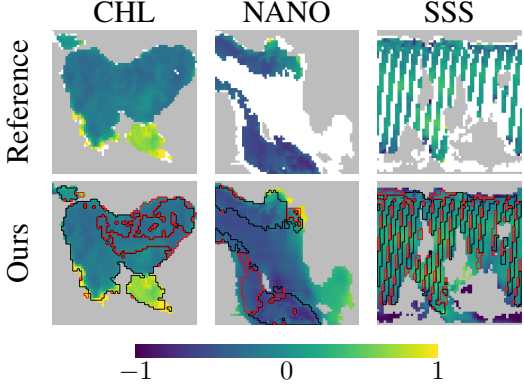}
  \par\vspace{-0.9\baselineskip}
  \caption{Full-mask ocean reconstructions at $64\times64$, with supplied observations restored in Ours. Black and red outlines mark supplied observations and withheld source-supported scoring locations, respectively. Gray denotes outside-domain locations; white in Reference denotes source-missing values.}
  \label{fig:local-global-reconstruction}
  \end{wrapfigure}%

\paragraph{Prior replacement.}
We next ask whether the result follows from any two-stage correction pipeline or from the particular locally learned reference.
We replace the learned local consensus prior with a full-domain OAMP prediction (Global2Global) or a low-rank DINEOF reconstruction (Low-rank2Global), preserving the prior--corrector connection design and retraining a separate global residual corrector for each replacement.
Relative to the standalone OAMP and DINEOF results in Table~\ref{tab:ocean64}, their respective cascades improve both MSE and PSNR (Table~\ref{tab:matched}). Prior Residual nevertheless achieves lower MSE and higher PSNR than both cascades on all three products, reducing MSE by 21.9\%--35.7\% relative to Global2Global and 11.5\%--35.3\% relative to Low-rank2Global.
Thus a corrector can improve alternative references, but the choice of reference still matters within the tested two-stage designs.

Figure~\ref{fig:local-global-reconstruction} shows full-mask reconstructions of the three ocean products. Figure~\ref{fig:ocean-decomposition} in Appendix~\ref{app:ocean64-visualizations} decomposes these predictions into local priors and learned residuals, while Figures~\ref{fig:ocean64-chl}--\ref{fig:ocean64-sss} compare Ours with Full Direct.

\begin{wrapfigure}[16]{R}{0.50\textwidth}
    \centering
    \vspace{-0.5\baselineskip}
    \setlength{\abovecaptionskip}{3pt}
    \input{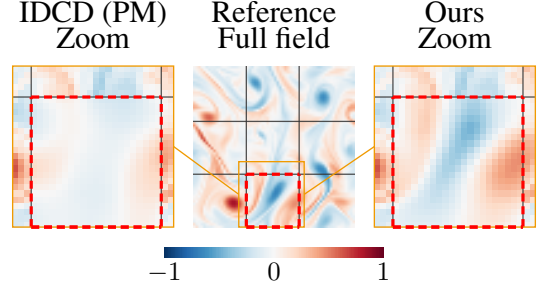}
    \par\vspace{-0.5\baselineskip}
    \caption{Navier--Stokes Block 8/9 reconstruction at $64\times64$. IDCD (PM) is the parameter-matched IDCD control, denoted Full Direct. Red dashes mark the held-out block, black lines the acquisition partition, and orange boxes the zoomed region.}
    \label{fig:ns-block-detail}
\end{wrapfigure}

\paragraph{Fine-scale structure under block missingness.}
In the highlighted missing block of Navier--Stokes Block 8/9, Ours retains a more distinct elongated vorticity structure than IDCD (PM), with local contrast closer to the reference (Figure~\ref{fig:ns-block-detail}). Appendices~\ref{app:pde-pixel-visualizations}--\ref{app:pde-block-visualizations} provide additional PDE reconstruction examples.

\paragraph{Inference cost.}
At the evaluated common reconstruction-latency ceilings, our estimator retains lower MSE than parameter-matched Full Direct on Black Sea CHL and Shallow Water Pixel 30\% (Table~\ref{tab:inference-cost}). Appendix~\ref{app:inference-cost} reports the inference-mode comparison and the corresponding observation-retention and spatial-coverage statistics (Table~\ref{tab:observation-retention}).

\paragraph{Resolution.}
With each learned method trained and its checkpoint selected separately at $128\times128$, our estimator achieves lower MSE and higher PSNR than all six ocean baselines under the same source-supported scoring protocol (Appendix~\ref{app:resolution-additional-settings}, Table~\ref{tab:ocean128}). Appendix~\ref{app:ocean128-visualizations} shows reconstructions at this resolution (Figures~\ref{fig:ocean128-chl}--\ref{fig:ocean128-sss}).

\subsection{Prior Construction and Connection Ablations}
\label{sec:exp-coordinates-discrepancies}
\label{sec:exp-ablations}

We examine how the local reference is constructed and connected to the corrector. Removing local coordinates retrains both stages, whereas input-map controls fix the prior constructor and retrain only the corrector (Appendix~\ref{app:retraining-reporting}). Both controls evaluate the final full-mask reconstruction.

\paragraph{Constructing the local prior.}
Removing local coordinates raises final MSE on all three products (Table~\ref{tab:construction-controls}, Panel A). For the local prior constructed from the full supplied mask before correction, uniform fusion gives the lowest MSE among tested weighting rules applied to identical overlapping predictions, while half-stride overlap gives lower MSE than non-overlap (Table~\ref{tab:main-fusion}). Final reconstruction favors $32\times32$ patches on CHL and NANO but $16\times16$ patches on SSS; patch scale is selected per product on validation data (Table~\ref{tab:patch-scale}). Appendix~\ref{app:prior-construction-connection-ablations} gives the fusion definitions and explains the coupled changes in the covering comparison.

\paragraph{Connecting the prior to the corrector.}
Removing any tested input map raises final MSE on all three products. Removing both discrepancy maps gives the largest increase among these controls: 36.8\%, 22.9\%, and 17.4\% on CHL, NANO, and SSS, respectively (Appendix~\ref{app:prior-construction-connection-ablations}, Table~\ref{tab:construction-controls}, Panel B). These results support explicit prior--evidence features for a finite corrector, not access to new measurements. Because each corrector is retrained, the ablations do not isolate an independent causal contribution for each map.

\subsection{Does Global Correction Use Evidence Beyond Local Support?}
\label{sec:remote-evidence}
\label{sec:exp-remote-evidence}

Does global correction respond to observations that cannot affect the local prior at a target? For a scored block $B$, $U(B)$ is the union of contributing patch supports, so deleting observations outside it cannot change the prior on $B$ (Appendix~\ref{app:local-support}). We pair full-mask ocean reconstruction with deletion of the farthest third of supplied observations outside $U(B)$, fixing the checkpoint and perturbation and recomputing the prior, features, and final reconstruction.

The prior on $B$ remains unchanged for every eligible pair, while the corrected prediction has nonzero mean squared displacement on all three products (Table~\ref{tab:remote-displacement}). This establishes responsiveness beyond local support, but not whether the removed evidence improves accuracy. We test that separately using matched-count deletions from three remote distance bands and a Local Near control, comparing paired block MSE of the final reconstructions. Remote Mid and Remote Far deletion increase mean error on CHL and NANO, with 95\% confidence intervals above zero. On SSS, Remote Near and Local Near deletion increase mean error, Remote Far deletion decreases it, and the Remote Mid interval crosses zero (Table~\ref{tab:remote-error}). Thus the accuracy effect of evidence beyond local support depends on the product and distance band. Appendix~\ref{app:remote-evidence} specifies the diagnostic supports and paired deletion protocols.

\FloatBarrier
\begin{samepage}
\section{Conclusion}
\label{sec:conclusion}

We studied an explicit structural organization for learning from incomplete physical fields: construct a local consensus estimate, then learn its full-domain residual correction.
Estimator comparisons and retrained controls support this organization, alongside reconstruction gains on withheld source-supported ocean values.
The central finding is that a local estimate can be valuable as a revisable intermediate representation even when it is not the strongest standalone predictor, allowing local structure to be learned explicitly while retaining full-domain observations for correction.

Remaining boundary artifacts under block missingness are discussed in Appendix~\ref{app:limitations}.

\end{samepage}

\flushbottom

\clearpage
\subsection*{AI use statement}
\phantomsection\label{app:llm-usage}\label{sec:ai-use}

Generative AI tools were used for language editing, manuscript organization,
LaTeX assistance, and preliminary feedback on mathematical exposition and
experimental design. They did not execute experiments, access unreleased data,
or generate reported measurements. The authors independently verified all
definitions, proofs, citations, protocols, and numerical claims and take full
responsibility for the final manuscript.

\subsection*{Reproducibility statement}
\phantomsection\label{sec:reproducibility}

To support reproducibility, we plan to make the experimental code publicly
available. The release will include the model
implementations, training and evaluation procedures, experimental
configurations, and documentation. The datasets used in our experiments are
available from their original providers and remain subject to the providers'
access and licensing terms. The ocean data are provided by the E.U. Copernicus
Marine Environment Monitoring Service \citep{letraon2017cmems}.
Appendix~\ref{app:copernicus-provenance} records the
corresponding product identifiers and DOIs and describes the preprocessing
used in our experiments.
The use of generative AI tools is documented in the \hyperref[sec:ai-use]{AI Use Statement}.

\bibliographystyle{plainnat}
\bibliography{references}

@article{alveraazcarate2005dineof,
  title = {Reconstruction of incomplete oceanographic data sets using empirical orthogonal functions: application to the Adriatic Sea surface temperature},
  journal = {Ocean Modelling},
  volume = {9},
  number = {4},
  pages = {325-346},
  year = {2005},
  issn = {1463-5003},
  doi = {https://doi.org/10.1016/j.ocemod.2004.08.001},
  url = {https://www.sciencedirect.com/science/article/pii/S1463500304000472},
  author = {A. Alvera-Azcárate and A. Barth and M. Rixen and J.M. Beckers}
}

@InProceedings{bartal2023multidiffusion,
  title = {{M}ulti{D}iffusion: Fusing Diffusion Paths for Controlled Image Generation},
  author = {Bar-Tal, Omer and Yariv, Lior and Lipman, Yaron and Dekel, Tali},
  booktitle = {Proceedings of the 40th International Conference on Machine Learning},
  pages = {1737--1752},
  year = {2023},
  editor = {Krause, Andreas and Brunskill, Emma and Cho, Kyunghyun and Engelhardt, Barbara and Sabato, Sivan and Scarlett, Jonathan},
  volume = {202},
  series = {Proceedings of Machine Learning Research},
  month = {23--29 Jul},
  publisher = {PMLR},
  url = {https://proceedings.mlr.press/v202/bar-tal23a.html}
}

@Article{barth2024dindiff,
  AUTHOR = {Barth, A. and Brajard, J. and Alvera-Azc\'arate, A. and Mohamed, B. and Troupin, C. and Beckers, J.-M.},
  TITLE = {Ensemble reconstruction of missing satellite data using a denoising diffusion model: application to chlorophyll $a$ concentration in the Black Sea},
  JOURNAL = {Ocean Science},
  VOLUME = {20},
  YEAR = {2024},
  NUMBER = {6},
  PAGES = {1567--1584},
  URL = {https://os.copernicus.org/articles/20/1567/2024/},
  DOI = {10.5194/os-20-1567-2024}
}

@article{beckers2003dineof,
  author = "J. M.  Beckers and M.  Rixen",
  title = "EOF Calculations and Data Filling from Incomplete Oceanographic Datasets",
  journal = "Journal of Atmospheric and Oceanic Technology",
  year = "2003",
  publisher = "American Meteorological Society",
  address = "Boston MA, USA",
  volume = "20",
  number = "12",
  doi = "10.1175/1520-0426(2003)020<1839:ECADFF>2.0.CO;2",
  pages = "1839 - 1856",
  url = "https://journals.ametsoc.org/view/journals/atot/20/12/1520-0426_2003_020_1839_ecadff_2_0_co_2.xml"
}

@misc{cao2024navierstokes,
  author = {Shuhao Cao},
  title = {Navier-stokes dataset of isotropic turbulence in a periodic box},
  year = {2024},
  url = {https://huggingface.co/datasets/scaomath/navier-stokes-dataset},
  note = {Funded by National Science Foundation: NSF award DMS-2309778.}
}

@inproceedings{cao2025spectralrefiner,
  author = {Cao, Shuhao and Brarda, Francesco and Li, Ruipeng and Xi, Yuanzhe},
  booktitle = {International Conference on Learning Representations},
  editor = {Y. Yue and A. Garg and N. Peng and F. Sha and R. Yu},
  pages = {60348--60384},
  title = {Spectral-Refiner: Accurate Fine-Tuning of Spatiotemporal Fourier Neural Operator for Turbulent Flows},
  url = {https://proceedings.iclr.cc/paper_files/paper/2025/file/972eea97af7fe9a7588f5af136edef0c-Paper-Conference.pdf},
  volume = {2025},
  year = {2025}
}

@techreport{letraon2017cmems,
  TITLE = {{THE COPERNICUS MARINE ENVIRONMENTAL MONITORING SERVICE: MAIN SCIENTIFIC ACHIEVEMENTS AND FUTURE PROSPECTS}},
  AUTHOR = {Letraon, Py and Ali, A. and Fanjul, E. Alvarez and Aouf, L. and Axell, L. and Aznar, R. and Ballarotta, M. and Behrens, A. and Benkiran, M. and Bentamy, A. and Bertino, L. and Bowyer, P. and Brando, V. and Breivik, L. A. and Nardelli, B. Buongiorno and Cailleau, S. and Ciliberti, S. A. and Colella, S. and Connell, N. Mc and Coppini, G. and Cossarini, G. and Dabrowski, T. and Alonsomu{\~n}oyerro, M. de Alfonso and O'dea, E. and Desportes, C. and Dinessen, F. and Drevillon, M. and Drillet, Y. and Drudi, M. and Dussurget, R. and Faug{\`e}re, Y. and Forneris, V. and Fratianni, C. and Galloudec, O. Le and Hermosa, I. Garcia and Sotillo, M. Garc{\'i}a and Garnesson, P. and Garric, G. and Golbeck, I. and Gourrion, J. and Gr{\'e}goire, M. L. and Guinehut, S. and Gutknecht, E. and Harris, C. and Hernandez, F. and Huess, V. and Johannessen, J.A. and Kay, S. and Killick, R. and King, R. and Kloe, J. De and Korres, G. and Lagemaa, P. and Lecci, R. and Legeais, J.F. and Lellouche, J. M. and Levier, B. and Lorente, P. and Mangin, A. and Martin, M. and Melet, A. and Murawski, J. and {\"O}zsoy, E. and Palazov, A. and Pardo, S. and Parent, L. and Pascual, A. and Paul, J. and Peneva, E. and Perruche, C. and Peterson, D. and Villeon, L. Petit de La and Pinardi, N. and Pouliquen, S. and Pujol, M. I. and Rainaud, R. and Rampal, Pierre and Reffray, G. and Regnier, C. and Reppucci, A. and Ryan, A. and Salon, S. and Samuelsen, A. and Santoleri, R. and Saulter, Andrew and She, J. and Solidoro, C. and Stanev, E. and Staneva, J. and Stoffelen, A. and Storto, A. and Sykes, P. and Szekely, T. and Taburet, G. and Taylor, B. and Tintore, J. and Toledano, C. and Tonani, M. and Tuomi, L. and Volpe, G. and Wedhe, H. and Williams, T. and Vandendbulcke, L. and Zanten, D. Van and Schuckmann, K. Von and Xie, J. and Zacharioudaki, A. and Zuo, H.},
  URL = {https://univ-grenoble-alpes.hal.science/hal-03405376},
  TYPE = {Research Report},
  NUMBER = {56},
  INSTITUTION = {{MERCATOR OCEAN}},
  YEAR = {2017},
  MONTH = Sep,
  DOI = {10.25575/56},
  HAL_ID = {hal-03405376},
  HAL_VERSION = {v1}
}

@Article{zupancicmuc2025criter,
  AUTHOR = {Zupan\v{c}i\v{c} Muc, M. and Zavrtanik, V. and Barth, A. and Alvera-Azcarate, A. and Li\v{c}er, M. and Kristan, M.},
  TITLE = {CRITER 1.0: a coarse reconstruction with iterative refinement network for sparse spatio-temporal satellite data},
  JOURNAL = {Geoscientific Model Development},
  VOLUME = {18},
  YEAR = {2025},
  NUMBER = {17},
  PAGES = {5549--5573},
  URL = {https://gmd.copernicus.org/articles/18/5549/2025/},
  DOI = {10.5194/gmd-18-5549-2025}
}

@inproceedings{daras2023ambient,
  author = {Daras, Giannis and Shah, Kulin and Dagan, Yuval and Gollakota, Aravind and Dimakis, Alex and Klivans, Adam},
  booktitle = {Advances in Neural Information Processing Systems},
  doi = {10.52202/075280-0015},
  editor = {A. Oh and T. Naumann and A. Globerson and K. Saenko and M. Hardt and S. Levine},
  pages = {288--313},
  publisher = {Curran Associates, Inc.},
  title = {Ambient Diffusion: Learning Clean Distributions from Corrupted Data},
  url = {https://proceedings.neurips.cc/paper_files/paper/2023/file/012af729c5d14d279581fc8a5db975a1-Paper-Conference.pdf},
  volume = {36},
  year = {2023}
}

@InProceedings{givens2025score,
  title = {Score Matching with Missing Data},
  author = {Givens, Josh and Liu, Song and Reeve, Henry},
  booktitle = {Proceedings of the 42nd International Conference on Machine Learning},
  pages = {19523--19561},
  year = {2025},
  editor = {Singh, Aarti and Fazel, Maryam and Hsu, Daniel and Lacoste-Julien, Simon and Berkenkamp, Felix and Maharaj, Tegan and Wagstaff, Kiri and Zhu, Jerry},
  volume = {267},
  series = {Proceedings of Machine Learning Research},
  month = {13--19 Jul},
  publisher = {PMLR},
  url = {https://proceedings.mlr.press/v267/givens25a.html}
}

@misc{graves2023bfn,
  title = {Bayesian Flow Networks},
  author = {Alex Graves and Rupesh Kumar Srivastava and Timothy Atkinson and Faustino Gomez},
  year = {2025},
  eprint = {2308.07037},
  archivePrefix = {arXiv},
  primaryClass = {cs.LG},
  url = {https://arxiv.org/abs/2308.07037}
}

@inproceedings{hu2024padis,
  author = {Hu, Jason and Song, Bowen and Xu, Xiaojian and Shen, Liyue and Fessler, Jeffrey},
  booktitle = {Advances in Neural Information Processing Systems},
  doi = {10.52202/079017-0052},
  editor = {A. Globerson and L. Mackey and D. Belgrave and A. Fan and U. Paquet and J. Tomczak and C. Zhang},
  pages = {1625--1660},
  publisher = {Curran Associates, Inc.},
  title = {Learning Image Priors Through Patch-Based Diffusion Models for Solving Inverse Problems},
  url = {https://proceedings.neurips.cc/paper_files/paper/2024/file/03261886741f1f21f52f2a2d570616a2-Paper-Conference.pdf},
  volume = {37},
  year = {2024}
}

@InProceedings{pmlr-v80-lehtinen18a,
  title = {{N}oise2{N}oise: Learning Image Restoration without Clean Data},
  author = {Lehtinen, Jaakko and Munkberg, Jacob and Hasselgren, Jon and Laine, Samuli and Karras, Tero and Aittala, Miika and Aila, Timo},
  booktitle = {Proceedings of the 35th International Conference on Machine Learning},
  pages = {2965--2974},
  year = {2018},
  editor = {Dy, Jennifer and Krause, Andreas},
  volume = {80},
  series = {Proceedings of Machine Learning Research},
  month = {10--15 Jul},
  publisher = {PMLR},
  url = {https://proceedings.mlr.press/v80/lehtinen18a.html}
}

@InProceedings{li2024patchfusion,
  author = {Li, Zhenyu and Bhat, Shariq Farooq and Wonka, Peter},
  title = {PatchFusion: An End-to-End Tile-Based Framework for High-Resolution Monocular Metric Depth Estimation},
  booktitle = {Proceedings of the IEEE/CVF Conference on Computer Vision and Pattern Recognition (CVPR)},
  month = {June},
  year = {2024},
  pages = {10016-10025}
}

@InProceedings{li2024patchrefiner,
  author = "Li, Zhenyu
and Bhat, Shariq Farooq
and Wonka, Peter",
  editor = "Leonardis, Ale{\v{s}}
and Ricci, Elisa
and Roth, Stefan
and Russakovsky, Olga
and Sattler, Torsten
and Varol, G{\"u}l",
  title = "PatchRefiner: Leveraging Synthetic Data for Real-Domain High-Resolution Monocular Metric Depth Estimation",
  booktitle = "Computer Vision -- ECCV 2024",
  year = "2025",
  publisher = "Springer Nature Switzerland",
  address = "Cham",
  pages = "250--267",
  isbn = "978-3-031-72855-6"
}

@inproceedings{li2026patchrefinerv2,
  author = {Li, Zhenyu and Cui, Wenqing and Bhat, Shariq and Wonka, Peter},
  booktitle = {International Conference on Learning Representations},
  editor = {C. Vondrick and B. Hariharan and C. Raffel and L. Pinto and D. Yang and A. Faust},
  pages = {117738--117753},
  title = {PatchRefiner V2: Fast and Lightweight Real-Domain High-Resolution Metric Depth Estimation},
  url = {https://proceedings.iclr.cc/paper_files/paper/2026/file/bf2c36a946443a6f5b9e696a7b224ffb-Paper-Conference.pdf},
  volume = {2026},
  year = {2026}
}

@inproceedings{liu2018coordconv,
  author = {Liu, Rosanne and Lehman, Joel and Molino, Piero and Petroski Such, Felipe and Frank, Eric and Sergeev, Alex and Yosinski, Jason},
  booktitle = {Advances in Neural Information Processing Systems},
  editor = {S. Bengio and H. Wallach and H. Larochelle and K. Grauman and N. Cesa-Bianchi and R. Garnett},
  pages = {},
  publisher = {Curran Associates, Inc.},
  title = {An intriguing failing of convolutional neural networks and the CoordConv solution},
  url = {https://proceedings.neurips.cc/paper_files/paper/2018/file/60106888f8977b71e1f15db7bc9a88d1-Paper.pdf},
  volume = {31},
  year = {2018}
}

@misc{ma2026observation,
  title = {Observation-Aligned Mask Priors for Learning Physical Dynamics from Authentic Occlusions},
  author = {Chiyuan Ma and Zihan Zhou and Tianshu Yu},
  year = {2026},
  eprint = {2605.16818},
  archivePrefix = {arXiv},
  primaryClass = {cs.CV},
  url = {https://arxiv.org/abs/2605.16818}
}

@inproceedings{ouyang2023missdiff,
  title = {MissDiff: Training Diffusion Models on Tabular Data with Missing Values},
  author = {Yidong Ouyang and Liyan Xie and Chongxuan Li and Guang Cheng},
  booktitle = {ICML 2023 Workshop on Structured Probabilistic Inference {\&} Generative Modeling},
  year = {2023},
  url = {https://openreview.net/forum?id=S435pkeAdT}
}

@inproceedings{park2026msm,
  author = {Park, Chicago Y. and Shoushtari, Shirin and An, Hongyu and Kamilov, Ulugbek},
  booktitle = {International Conference on Learning Representations},
  editor = {C. Vondrick and B. Hariharan and C. Raffel and L. Pinto and D. Yang and A. Faust},
  pages = {96428--96461},
  title = {Measurement Score-Based Diffusion Model},
  url = {https://proceedings.iclr.cc/paper_files/paper/2026/file/9bcc5e97c6343b52dc4ff7b034cc640c-Paper-Conference.pdf},
  volume = {2026},
  year = {2026}
}

@inproceedings{song2021scoresde,
  title = {Score-Based Generative Modeling through Stochastic Differential Equations},
  author = {Yang Song and Jascha Sohl-Dickstein and Diederik P Kingma and Abhishek Kumar and Stefano Ermon and Ben Poole},
  booktitle = {International Conference on Learning Representations},
  year = {2021},
  url = {https://openreview.net/forum?id=PxTIG12RRHS}
}

@inproceedings{takamoto2022pdebench,
  author = {Takamoto, Makoto and Praditia, Timothy and Leiteritz, Raphael and MacKinlay, Daniel and Alesiani, Francesco and Pfl\"{u}ger, Dirk and Niepert, Mathias},
  booktitle = {Advances in Neural Information Processing Systems},
  doi = {10.52202/068431-0117},
  editor = {S. Koyejo and S. Mohamed and A. Agarwal and D. Belgrave and K. Cho and A. Oh},
  pages = {1596--1611},
  publisher = {Curran Associates, Inc.},
  title = {PDEBench: An Extensive Benchmark for Scientific Machine Learning},
  url = {https://proceedings.neurips.cc/paper_files/paper/2022/file/0a9747136d411fb83f0cf81820d44afb-Paper-Datasets_and_Benchmarks.pdf},
  volume = {35},
  year = {2022}
}

@inproceedings{wang2023patch,
  author = {Wang, Zhendong and Jiang, Yifan and Zheng, Huangjie and Wang, Peihao and He, Pengcheng and Wang, Zhangyang "Atlas" and Chen, Weizhu and Zhou, Mingyuan},
  booktitle = {Advances in Neural Information Processing Systems},
  doi = {10.52202/075280-3158},
  editor = {A. Oh and T. Naumann and A. Globerson and K. Saenko and M. Hardt and S. Levine},
  pages = {72137--72154},
  publisher = {Curran Associates, Inc.},
  title = {Patch Diffusion: Faster and More Data-Efficient Training of Diffusion Models},
  url = {https://proceedings.neurips.cc/paper_files/paper/2023/file/e4667dd0a5a54b74019b72b677ed8ec1-Paper-Conference.pdf},
  volume = {36},
  year = {2023}
}

@inproceedings{zhang2025diffputer,
  author = {Zhang, Hengrui and Fang, Liancheng and Wu, Qitian and Yu, Philip},
  booktitle = {International Conference on Learning Representations},
  editor = {Y. Yue and A. Garg and N. Peng and F. Sha and R. Yu},
  pages = {63164--63185},
  title = {DiffPuter: Empowering Diffusion Models for Missing Data Imputation},
  url = {https://proceedings.iclr.cc/paper_files/paper/2025/file/9e2ad37b8899366767df9319efc6f4de-Paper-Conference.pdf},
  volume = {2025},
  year = {2025}
}

@inproceedings{zhou2026incomplete,
  author = {Zhou, Zihan and Wang, Chenguang and Ye, Hongyi and Guan, Yongtao and Yu, Tianshu},
  booktitle = {International Conference on Learning Representations},
  editor = {C. Vondrick and B. Hariharan and C. Raffel and L. Pinto and D. Yang and A. Faust},
  pages = {10918--10952},
  title = {Incomplete Data, Complete Dynamics: A Diffusion Approach},
  url = {https://proceedings.iclr.cc/paper_files/paper/2026/file/123a18dfd821c8b440f42a00a27648d6-Paper-Conference.pdf},
  volume = {2026},
  year = {2026}
}

@inproceedings{xu2025ppidm,
  author = {Xu, Qianxun and Li, Zuchuan},
  booktitle = {Advances in Neural Information Processing Systems},
  doi = {10.52202/085713-5230},
  editor = {D. Belgrave and C. Zhang and H. Lin and R. Pascanu and P. Koniusz and M. Ghassemi and N. Chen},
  pages = {156490--156507},
  publisher = {Curran Associates, Inc.},
  title = {Partial Physics Informed Diffusion Model for Ocean Chlorophyll Concentration Reconstruction},
  url = {https://proceedings.neurips.cc/paper_files/paper/2025/file/e57cd5735565b737b2c2afe56daca99b-Paper-Conference.pdf},
  volume = {38, Main Conference},
  year = {2025}
}

@article{iizuka2017globally,
  author = {Iizuka, Satoshi and Simo-Serra, Edgar and Ishikawa, Hiroshi},
  title = {Globally and locally consistent image completion},
  year = {2017},
  issue_date = {August 2017},
  publisher = {Association for Computing Machinery},
  address = {New York, NY, USA},
  volume = {36},
  number = {4},
  issn = {0730-0301},
  url = {https://doi.org/10.1145/3072959.3073659},
  doi = {10.1145/3072959.3073659},
  journal = {ACM Trans. Graph.},
  month = jul,
  articleno = {107},
  numpages = {14}
}

@InProceedings{yu2018contextual,
  author = {Yu, Jiahui and Lin, Zhe and Yang, Jimei and Shen, Xiaohui and Lu, Xin and Huang, Thomas S.},
  title = {Generative Image Inpainting With Contextual Attention},
  booktitle = {Proceedings of the IEEE Conference on Computer Vision and Pattern Recognition (CVPR)},
  month = {June},
  year = {2018}
}

@INPROCEEDINGS{nazeri2019edgeconnect,
  author = {Nazeri, Kamyar and Ng, Eric and Joseph, Tony and Qureshi, Faisal and Ebrahimi, Mehran},
  booktitle = {2019 IEEE/CVF International Conference on Computer Vision Workshop (ICCVW)},
  title = {EdgeConnect: Structure Guided Image Inpainting using Edge Prediction},
  year = {2019},
  volume = {},
  number = {},
  pages = {3265-3274},
  doi = {10.1109/ICCVW.2019.00408}
}

\clearpage
\appendix

\raggedbottom
\section{Related Work}
\label{app:related-work}

\begin{table}[!b]
\centering
\caption{Original settings, reconstruction mechanisms, and experimental roles of the evaluated baselines.}
\label{tab:closest-prior-comparison}
\scriptsize
\setlength{\tabcolsep}{2.2pt}
\renewcommand{\arraystretch}{1.08}
\begin{tabularx}{\textwidth}{@{}
  >{\raggedright\arraybackslash}p{0.13\textwidth}
  >{\raggedright\arraybackslash}p{0.145\textwidth}
  >{\raggedright\arraybackslash}X
  >{\raggedright\arraybackslash}p{0.17\textwidth}@{}}
\toprule
Method & Original setting & Principal mechanism & Use in this paper \\
\midrule
\multicolumn{4}{@{}l}{\textit{Evaluated baselines}} \\
DINEOF \citep{beckers2003dineof,alveraazcarate2005dineof}
& Incomplete ocean fields
& Low-rank EOF reconstruction
& Ocean baseline; low-rank-prior control \\
AmbientDiff \citep{daras2023ambient}
& Corrupted images
& Further corruption; prediction supervised on available measurements
& Ocean and PDE baseline \\
DINDiff \citep{barth2024dindiff}
& Satellite ocean chlorophyll
& Noise prediction on additionally withheld observations
& Ocean and PDE baseline \\
MissDiff \citep{ouyang2023missdiff}
& Incomplete tabular data
& Masked denoising score matching
& Ocean and PDE baseline \\
MSM \citep{park2026msm}
& Natural and medical imaging
& Partial measurement scores; stochastic sampling
& Ocean and PDE baseline \\
IDCD \citep{zhou2026incomplete}
& PDE-oriented physical dynamics
& Observation-structured context--query prediction
& PDE baseline \\
OAMP \citep{ma2026observation}
& Authentically occluded ocean fields
& Learned mask prior; aligned context--query prediction
& Ocean baseline; full-domain-prior control \\
\bottomrule
\end{tabularx}
\end{table}

\begin{table}[!t]
\centering
\caption{Comparison with related local--global and staged reconstruction methods in their original training settings.}
\label{tab:cv-differences}
\scriptsize
\setlength{\tabcolsep}{3pt}
\renewcommand{\arraystretch}{1.13}
\begin{tabularx}{\textwidth}{@{}
  >{\raggedright\arraybackslash}p{0.15\textwidth}
  >{\raggedright\arraybackslash}p{0.20\textwidth}
  >{\raggedright\arraybackslash}p{0.24\textwidth}
  >{\raggedright\arraybackslash}X@{}}
\toprule
Method & Original training setting & Construction and revision & Distinction from our estimator \\
\midrule
GLCIC \citep{iizuka2017globally} & Artificially masked complete images & A completion network trained with global and local discriminators & Local--global consistency is imposed through training losses, rather than a local-value consensus followed by a learned full-domain corrector. \\
\addlinespace[3pt]
Contextual Attention \citep{yu2018contextual} & Artificially masked complete images & Coarse completion followed by contextual-attention refinement & Uses a coarse-to-fine image-completion pipeline rather than a reference learned from incomplete spatial patches. \\
\addlinespace[3pt]
EdgeConnect \citep{nazeri2019edgeconnect} & Images and derived edge targets & Predicted edges condition subsequent image completion & The intermediate representation is an edge map, not a field-valued estimate used as a residual anchor. \\
\addlinespace[3pt]
Patch Diffusion \citep{wang2023patch} & Clean images with coordinate-conditioned, multiscale training & Learns scores from patch and full-image training views & Patch learning serves diffusion-model training rather than held-out-query prediction followed by a separate full-domain corrector. \\
\addlinespace[3pt]
PaDIS \citep{hu2024padis} & Clean image patches for prior learning & Patch scores define a prior used by an inverse-problem solver & Global measurement constraints enter the solver; there is no separately trained full-domain residual corrector around an observed-query-trained local consensus. \\
\addlinespace[3pt]
MultiDiffusion \citep{bartal2023multidiffusion} & A pretrained text-to-image diffusion model & Reconciles regional diffusion updates during sampling & Fusion operates on sampling updates, rather than constructing a frozen clean-value reference for a learned corrector. \\
\addlinespace[3pt]
PatchFusion \citep{li2024patchfusion} & Image--depth training data & Fuses global coarse and tiled depth predictions & Global information guides high-resolution tile fusion, rather than revising a consensus constructed from incomplete local field observations. \\
\addlinespace[3pt]
PatchRefiner \citep{li2024patchrefiner} & Image--depth data with synthetic-to-real supervision & Frozen global coarse depth predictor followed by patchwise residual refinement & Shares a frozen reference and residual refinement; the reference origin, depth supervision, and global-coarse-to-patch direction differ. \\
\addlinespace[3pt]
PatchRefiner V2 \citep{li2026patchrefinerv2} & Image--depth data with synthetic-to-real supervision & Lightweight patch refinement and cross-scale fusion; coarse network and refiner jointly fine-tuned & Jointly fine-tunes global-coarse-to-patch refinement; ours freezes the local predictor before learning full-domain correction. \\
\addlinespace[3pt]
CRITER \citep{zupancicmuc2025criter} & Sparse spatiotemporal satellite fields & Global low-frequency reconstruction followed by iterative refinement & Begins with global spatiotemporal reconstruction, whereas ours begins with spatially local estimates and then applies full-domain revision. \\
\addlinespace[3pt]
\textbf{Ours} & \textbf{Never-complete field observations with context--query supervision} & \textbf{Coordinate-conditioned local predictions $\to$ overlap consensus $\to$ learned full-domain residual correction} & \textbf{The same observation-conditioned field is used as an explicit input and the residual output anchor.} \\
\bottomrule
\end{tabularx}
\end{table}

\subsection{Learning from Incomplete Training Data}
\label{app:related-incomplete}

Learning from incomplete training samples is distinct from filling test-time gaps with a model learned from complete data \citep{zhou2026incomplete}.
AmbientDiff introduces additional measurement corruption and trains predictions against the available measurements \citep{daras2023ambient}.
MissDiff adapts denoising score matching by masking its regression loss to observed entries \citep{ouyang2023missdiff}, while MSM learns partial measurement scores from noisy, subsampled observations and combines them through stochastic sampling \citep{park2026msm}.

Alternative formulations include DiffPuter's expectation--maximization procedure, which alternates diffusion-model learning and imputation \citep{zhang2025diffputer}, and the importance-weighted and variational score-matching objectives of \citet{givens2025score}.
These approaches develop ways to learn and reconstruct without complete training samples.
Our study concerns the organization of a point estimator around a locally constructed field and its full-domain revision. Unlike MSM's sampling-time composition of partial measurement scores \citep{park2026msm}, we reconcile patchwise clean-value predictions into a frozen reference and train a separate residual corrector under held-out-query supervision. Appendix~\ref{app:pde-comparisons} (``MSM under pixel observations'') provides a mechanism-level comparison and discusses MSM under the Pixel 30\% protocol.

\subsection{Physical-Field Reconstruction and Observation-Aligned Supervision}
\label{app:related-observation-aligned}

Geophysical reconstruction has long used structural assumptions to infer missing values.
DINEOF reconstructs incomplete ocean fields through low-rank empirical orthogonal functions \citep{beckers2003dineof,alveraazcarate2005dineof}.
DINDiff trains a denoising diffusion model directly on cloud-obscured satellite chlorophyll, using additionally withheld observations to supervise noise prediction \citep{barth2024dindiff}.
PPIDM incorporates a trusted partial physics operator into diffusion training for ocean chlorophyll reconstruction \citep{xu2025ppidm}.

Within context--query learning, the observation pattern determines which relationships can receive supervision.
IDCD develops a PDE-oriented framework that aligns context--query sampling with known pixel- or block-structured observation patterns and analyzes the role of query exposure \citep{zhou2026incomplete}.
OAMP is designed for ocean reconstruction under authentic occlusions: it learns a prior over binary observation masks with a Bayesian Flow Network \citep{graves2023bfn} and uses guided sampling and mask intersection to construct context--query pairs \citep{ma2026observation}.
We build on their observation-aligned supervision and adopt IDCD's single-step prediction, with inference modes specified in Section~\ref{sec:method-inference}.
The inherited mask construction selects available observations for context and supervision; our local field prior is a different object, supplying sample-conditioned values for a full-domain estimator to revise.

Table~\ref{tab:closest-prior-comparison} summarizes the original settings, principal mechanisms, and experimental roles of the evaluated baselines.

\subsection{Patch Priors and Local--Global Estimation}
\label{app:related-local-global}

Local--global organization and staged prediction have a long history in image completion.
Globally and Locally Consistent Image Completion uses global and local discriminators to train a completion network \citep{iizuka2017globally}.
Contextual Attention employs a coarse-to-fine completion pipeline \citep{yu2018contextual}, while EdgeConnect predicts edges before conditioning image completion on them \citep{nazeri2019edgeconnect}.

Patch Diffusion learns position-conditioned patch scores \citep{wang2023patch}, and PaDIS constructs image priors from position-aware patch scores for inverse-problem solvers \citep{hu2024padis}.
MultiDiffusion reconciles regional diffusion updates through least-squares fusion during sampling \citep{bartal2023multidiffusion}.
Our construction also uses positional information and overlap reconciliation, but its intermediate object is a clean-value field conditioned on the current observations.
The patch predictor is trained on query targets drawn only from observed entries of incomplete fields, and predictions from the frozen local model are reconciled before learned full-domain correction.

Other multiscale estimators use a global coarse prediction to support local detail.
PatchFusion combines a globally consistent coarse depth estimate with tiled predictions \citep{li2024patchfusion}. PatchRefiner freezes a global coarse depth predictor and learns patchwise residual refinements \citep{li2024patchrefiner}. PatchRefiner V2 introduces lightweight encoding and cross-scale feature fusion; after separate pretraining, the coarse network and refiner are jointly fine-tuned end to end \citep{li2026patchrefinerv2}.
For satellite fields, CRITER first reconstructs low-frequency structure using global spatiotemporal correlations and then iteratively refines the estimate \citep{zupancicmuc2025criter}.
Our estimator proceeds from local construction to full-domain revision: a shared local predictor produces a consensus reference, and a separate corrector retains the original context to revise that reference.
The reference field serves both as an input, accompanied by observation-discrepancy features, and as the anchor of the residual output.
When context ensembling is used, construction and correction are performed separately for each context before averaging the corrected fields.
Table~\ref{tab:cv-differences} compares the original supervision, intermediate representations, and reconstruction organization of related local--global and staged methods.

\FloatBarrier

\section{Method Implementation and Reconstruction}
\label{app:architecture-inference}
\label{app:algorithms}

This appendix specifies the diffusion parameterization, local-prior construction, estimator interfaces, reconstruction settings, and complete training and reconstruction procedure.

\subsection{Diffusion Parameterization and Noise Schedule}
\label{app:diffusion-dynamics}
\label{app:diffusion-objectives}
\label{app:schedule}

We use the variance-preserving corruption
\begin{equation}
    \bm{x}_t=\alpha_t\bm{x}_0+\sigma_t\bm{\epsilon},
    \qquad
    \bm{\epsilon}\sim\mathcal N(0,I),
    \qquad
    \alpha_t^2+\sigma_t^2=1,
    \label{eq:forward-corruption}
\end{equation}
and parameterize both local and full-domain networks as clean-data predictors.
Given context $\bm{M}_{\mathrm{ctx}}$,
\begin{equation}
    \bm{x}_{\ctx,t}
    =\bm{M}_{\mathrm{ctx}}\odot\left(\alpha_t\bm{x}_{\mathrm{obs}}+\sigma_t\bm{\epsilon}\right).
    \label{eq:incomplete-diffusion-state}
\end{equation}
The diffusion conventions follow \citet{song2021scoresde}.
We use clean-data prediction so that the local prior, full-domain output, and residual correction are represented in physical-field space.
Training samples diffusion times across the prescribed interval, while inference uses the near-zero time $\delta$ with the reconstruction modes in Section~\ref{sec:method-inference}.

The noise schedule is
\begin{equation}
    \alpha_t=\exp\!\left[
      -\frac14(\beta_{\max}-\beta_{\min})t^2
      -\frac12\beta_{\min}t
    \right],
    \qquad
    \sigma_t^2=1-\alpha_t^2.
    \label{eq:app-vp-schedule}
\end{equation}

\subsection{Local Prediction and Overlap Consensus}
\label{app:formal-estimator-definitions}
\label{app:patch-geometry}

We use the main-text noisy context $\bm{x}_{\ctx,t}=\bm{M}_{\mathrm{ctx}}\odot \bm{x}_{\mathrm{obs},t}$ and abbreviate its observable state by $O_t=(\bm{x}_{\ctx,t},\bm{M}_{\mathrm{ctx}})$.
The crop $R_p$, support $\Omega_p$, and covering $\mathcal P$ are defined in Section~\ref{sec:method-stage1}; let $\bm{\Gamma}_{\mathcal P}=(R_p\bm{\Gamma})_{p\in\mathcal P}$ collect the coordinate crops.
The partition law is denoted by $(\bm{M}_{\mathrm{ctx}},\bm{M}_{\mathrm{qry}})\sim\mathcal P_{\mathrm{align}}(\cdot\mid \bm{M})$.
The compact local input is
\begin{equation}
J_{p,t}=(R_p\bm{x}_{\ctx,t},R_p\bm{M}_{\mathrm{ctx}},R_p\bm{\Gamma}).
\label{eq:local-estimator-input}
\end{equation}
For view $p$, set $\bm{M}_{\mathrm{ctx},p}=R_p\bm{M}_{\mathrm{ctx}}$ and $\bm{M}_{\mathrm{qry},p}=R_p\bm{M}_{\mathrm{qry}}$. The local denoiser receives the
corrupted context values, context mask, and normalized absolute coordinates:
\begin{equation}
    \bm{Z}_{p,t}^{(1)}
    =\cat\!\left[
        R_p\bm{x}_{\ctx,t},
        R_p\bm{M}_{\mathrm{ctx}},
        R_p\bm{\Gamma}
      \right].
    \label{eq:stage1-input}
\end{equation}
The coordinate atlas $\bm{\Gamma}_{\mathcal P}$ is fixed by the covering geometry.
A training item uses one perturbation at its sampled time, while an inference item uses one perturbation to form $\bm{x}_{\mathrm{in},\delta}$ before applying either the full supplied mask or the ensemble subcontext masks.
Within a context evaluation, all local views, discrepancy features, and the full-domain estimator use the same realized corrupted values.

With this notation, $\bm{x}_{\bm{\phi}}(t;J_{p,t})$ is the prediction in Equation~\eqref{eq:local-prediction-main}, trained with Equation~\eqref{eq:stage1-loss}.
After local training, $\bm{\phi}$ is held fixed during prior construction and global correction.
Local training samples views with $\|R_p\bm{M}_{\mathrm{qry}}\|_1>0$; prior construction evaluates the full covering.

The frozen predictor and overlap consensus define
\begin{equation}
\bm{\Pi}_t=\mathcal P_{\bm{\phi}}(t;O_t;\bm{\Gamma}_{\mathcal P}).
\label{eq:local-field-prior-definition}
\end{equation}
Here $\mathcal P_{\bm{\phi}}$ is a deterministic construction operator, not a probability law.
For fixed patch predictions $\bm{Y}_{p,t}=\bm{x}_{\bm{\phi}}(t;J_{p,t})$, overlap consensus solves
\begin{equation}
\bm{\Pi}_t=\arg\min_{\bm{Z}=\bm{V}\odot\bm{Z}}\sum_{p\in\mathcal P}
\left\|R_p\bm{Z}-\bm{Y}_{p,t}\right\|_2^2.
\label{eq:local-prior-consensus}
\end{equation}
At each valid coordinate $i$, differentiating with respect to $\bm{Z}_i$ gives
$\bm{W}_i\bm{Z}_i=\sum_{p:i\in\Omega_p}\bm{Y}_{p,t,i}$.
Every valid coordinate is covered, so $\bm{W}_i>0$ and the unique solution is the pointwise average in Equation~\eqref{eq:uniform-consensus-solution}; the validity constraint fixes the remaining coordinates to zero.

\paragraph{Default patch geometry.}
At $64\times64$, Black Sea CHL, Baltic Sea NANO, and the PDE models use $32\times32$ patches with stride 16, giving nine overlapping views.
The local-prior models for Global Ocean SSS use $16\times16$ patches with stride 8, giving 49 views, throughout the main, estimator-organization, component, fusion, and intervention experiments.
The ocean patch scales are selected on validation data; Appendix~\ref{app:prior-construction-connection-ablations} (Table~\ref{tab:patch_scale_64}) reports the comparison between $32\times32$/stride-16 and $16\times16$/stride-8 coverings.
At $128\times128$, the ocean models use $64\times64$ patches with stride 32, giving nine views.
Every covering includes all valid output coordinates, and the frozen local predictor makes one clean-field prediction per view for each evaluated context.
Each patch is evaluated independently from its local values, mask, and coordinate crop, using fixed preprocessing statistics and no operation that exchanges current-sample information across patch calls.

\subsection{Prior--Evidence Features and Estimator Interfaces}
\label{app:implementation-inputs}
\label{app:deterministic-prior-evidence-features}

The feature operator returns $\bm{D}_t=(\bm{W},\bm{S}_{\ctx},\bm{S}_{\partial})$, where $\bm{W}$ is the overlap multiplicity.
The overlap count $\bm{W}$ is defined in Equation~\eqref{eq:uniform-consensus-solution}.
At supplied coordinates, define
\begin{equation}
\bm{S}_{\ctx,i}=\bm{M}_{\mathrm{ctx},i}\frac1{d_x}\sum_{c=1}^{d_x}
\left|\bm{\Pi}_{t,c,i}-\bm{x}_{\ctx,t,c,i}\right|.
\label{eq:context-discrepancy}
\end{equation}
Let $N_i(\bm{M}_{\mathrm{ctx}})=\{j\in\Omega:\|j-i\|_1=1,\ \bm{M}_{\mathrm{ctx},j}=1\}$ and define the neighboring corrupted-context mean
\[
\bar{\bm{x}}_{\ctx,t,c,i}=
\begin{cases}
\displaystyle\frac1{|N_i(\bm{M}_{\mathrm{ctx}})|}\sum_{j\in N_i(\bm{M}_{\mathrm{ctx}})}\bm{x}_{\ctx,t,c,j},& |N_i(\bm{M}_{\mathrm{ctx}})|>0,\\
0,& |N_i(\bm{M}_{\mathrm{ctx}})|=0.
\end{cases}
\]
At a non-context coordinate adjacent to supplied observations, define
\begin{equation}
\bm{S}_{\partial,i}=(1-\bm{M}_{\mathrm{ctx},i})\mathbf1[|N_i(\bm{M}_{\mathrm{ctx}})|>0]\frac1{d_x}
\sum_{c=1}^{d_x}\left|\bm{\Pi}_{t,c,i}-\bar{\bm{x}}_{\ctx,t,c,i}\right|.
\label{eq:boundary-discrepancy}
\end{equation}
The time dependence of the discrepancy maps is suppressed in their notation, and all comparisons use the realized corrupted context of the corresponding model evaluation.

The full-domain input and its deterministic features can be abbreviated as
\begin{equation}
\begin{aligned}
(\bm{\Pi}_t,\bm{D}_t)&=h(t,O_t),\\
\bm{D}_t&=d_{\mathcal P}(O_t,\bm{\Pi}_t),\\
I_t&=(O_t,\sg(\bm{\Pi}_t),\bm{D}_t).
\end{aligned}
\label{eq:prior-conditioned-state}
\end{equation}
Here $\bm{D}_t=(\bm{W},\bm{S}_{\ctx},\bm{S}_{\partial})$, $d_{\mathcal P}$ is the feature constructor defined above, and $\sg$ denotes stop-gradient.
The nested tuple $I_t$ denotes exactly the full input in Equation~\eqref{eq:full-domain-input-main}.
The residual objective in Equation~\eqref{eq:stage2-loss} is equivalently the query loss on $\sg(\bm{\Pi}_t)+r_{\bm{\psi}}(t;I_t)$; the algorithms denote its minibatch version by $\mathcal L_{\mathrm{glob}}$.
The implementation operator $\mathrm{Cues}_{\mathcal P}(\bm{\Pi}_t,O_t)$ returns the deterministic feature tuple defined above.
The no-prior input tensor is
\begin{equation}
    \bm{Z}_t^{\mathrm{base}}
    =\cat\!\left[
      \bm{x}_{\ctx,t},\bm{M}_{\mathrm{ctx}},\bm{\Gamma}
    \right].
    \label{eq:global-base-input}
\end{equation}
The two prior-conditioned estimators receive the channel-ordered tensor
\begin{equation}
    \bm{Z}_t^{\mathrm{PC}}
    =\cat\!\left[
      \bm{x}_{\ctx,t},\bm{M}_{\mathrm{ctx}},\sg(\bm{\Pi}_t),\bm{W},\bm{S}_\ctx,\bm{S}_\partial
    \right].
    \label{eq:stage2-input}
\end{equation}
The no-prior full-domain direct estimator receives $\bm{Z}_t^{\mathrm{base}}$. The
prior-conditioned full-domain direct estimator and residual corrector receive
$\bm{Z}_t^{\mathrm{PC}}$. These channel orders are used during training and for every inference context.
With a single-channel mask, single-channel $W$, and channel-averaged discrepancy
maps, the input channel counts are
\[
\begin{aligned}
\dim_{\mathrm{channel}}\bm{Z}_t^{\mathrm{base}}&=d_x+3,\\
\dim_{\mathrm{channel}}\bm{Z}_t^{\mathrm{PC}}&=2d_x+4.
\end{aligned}
\]
For a univariate field ($d_x=1$), these are 4 and 6 channels, respectively;
$\bm{\Gamma}$ contributes two normalized absolute-coordinate channels.

Their output forms are
\[
\begin{aligned}
\widehat{\bm{x}}_t^{\mathrm{PD}}&=g_{\bm{\theta}}(t;I_t),\\
g_{\bm{\theta}_0}(t;I_t)&=0,\\[4pt]
\widehat{\bm{x}}_t^{\mathrm{PR}}&=\operatorname{sg}(\bm{\Pi}_t)+r_{\bm{\psi}}(t;I_t),\\
r_{\bm{\psi}_0}(t;I_t)&=0.
\end{aligned}
\]

Zero-output initialization therefore gives initial full-field predictions of zero for Prior Direct and $\bm{\Pi}_t$ for Prior Residual.
Under the shared query loss, the direct branch predicts field values, whereas the residual branch predicts their discrepancy from the frozen reference.

\subsection{Full-Mask and Context-Ensemble Reconstruction}
\label{app:full-support-inference}

For our estimator and internal controls, the validation-selected default settings are full-mask reconstruction on ocean data at both resolutions and $K=8$ context ensembling on PDE data.
Appendix~\ref{app:inference-cost} varies the inference mode and ensemble size.
Full-mask inference sets $K=1$ and $\bm{M}_{\mathrm{ctx}}^{(1)}=\bm{M}_{\mathrm{in}}$; a one-member context ensemble instead uses one subsampled context.
The same near-zero diffusion time $\delta$ is used across datasets, resolutions, validation, and testing.

Algorithm~\ref{alg:detailed-training-sampling} defines the states $O_\delta^{(k)}$, local priors $\bm{\Pi}_\delta^{(k)}$, and augmented inputs $I_\delta^{(k)}$ for the selected contexts.
The local-prior reference and the two direct estimators are evaluated as
\[
\begin{aligned}
\widetilde{\bm{\Pi}}_\delta
&=\frac1K\sum_{k=1}^{K}\bm{\Pi}_\delta^{(k)},\\
\widetilde{\bm{x}}_\delta^{\mathrm{NP}}
&=\frac1K\sum_{k=1}^{K}f_{\bm{\omega}}\!\left(\delta;O_\delta^{(k)},\bm{\Gamma}\right),\\
\widetilde{\bm{x}}_\delta^{\mathrm{PD}}
&=\frac1K\sum_{k=1}^{K}g_{\bm{\theta}}\!\left(\delta;I_\delta^{(k)}\right).
\end{aligned}
\]
One near-zero perturbation is shared by all contexts for each evaluation item. Each context independently completes local-prior construction and full-domain prediction before the resulting fields are averaged.
Each prior-conditioned prediction uses the prior and discrepancy features constructed from its own context.
The residual estimator combines corrected predictions as in Equation~\eqref{eq:proximal-output}, after which the original supplied observations are restored.
\Needspace{5\baselineskip}
\subsection{Training and Reconstruction Algorithm}
\label{app:complete-algorithms}

Algorithm~\ref{alg:detailed-training-sampling} specifies local training, frozen-prior construction, full-domain residual training, and full-mask or context-ensemble reconstruction.

Training accumulates query-masked squared errors over minibatches, while evaluation uses the collection-level metric definitions in Appendix~\ref{app:pooled-mse}.

\begin{algorithm}[!htb]
\caption{Overall Two-Stage Training and Reconstruction}
\label{alg:detailed-training-sampling}
\small
\begin{algorithmic}[1]
\Require Incomplete dataset $\mathcal D$, partition rule $\mathcal P_{\mathrm{align}}$, covering $\mathcal P$, coordinates $\bm{\Gamma}$, valid mask $\bm V$; inference input $(\bm{x}_{\mathrm{in}},\bm M_{\mathrm{in}})$, inference mode, $K$, $\delta$
\Ensure Trained parameters $(\bm\phi,\bm\psi)$ and reconstruction $\widehat{\bm x}_{\mathrm{out}}$
\Statex \textbf{Stage 1: learn local structure}
\While{local training has not converged}
  \State Sample $(\bm{x}_{\mathrm{obs}},\bm M)\sim\mathcal D$ and obtain $(\bm M_{\mathrm{ctx}},\bm M_{\mathrm{qry}})\sim\mathcal P_{\mathrm{align}}(\cdot\mid\bm M)$ using the dataset-specific feasibility handling in Appendix~\ref{app:aligned-partitions}
  \State Draw $t\sim\operatorname{Unif}[t_\varepsilon,1]$ and $\bm\epsilon\sim\mathcal N(0,I)$
  \State $\bm{x}_{\ctx,t}\gets\bm M_{\mathrm{ctx}}\odot(\alpha_t\bm{x}_{\mathrm{obs}}+\sigma_t\bm\epsilon)$
  \State Sample $p$ uniformly from $\{p\in\mathcal P:\|R_p\bm M_{\mathrm{qry}}\|_1>0\}$
  \State Predict $\widehat{\bm x}_0^p$ using Eq.~\eqref{eq:local-prediction-main}; update $\bm\phi$ with the patch query loss~\eqref{eq:stage1-loss}
\EndWhile
\State Freeze $\bm\phi$; initialize the residual head so that $r_{\bm\psi_0}=0$
\Statex \textbf{Stage 2: learn full-domain revision}
\While{global training has not converged}
  \State Sample an incomplete pair, a context--query partition, $t$, and $\bm\epsilon$ as in Stage 1
  \State $\bm{x}_{\ctx,t}\gets\bm M_{\mathrm{ctx}}\odot(\alpha_t\bm{x}_{\mathrm{obs}}+\sigma_t\bm\epsilon)$
  \State $(\bm\Pi_t,I_t)\gets\Call{BuildReference}{t,\bm{x}_{\ctx,t},\bm M_{\mathrm{ctx}}}$
  \State $\widehat{\bm x}_{0,t}\gets\sg(\bm\Pi_t)+r_{\bm\psi}(t;I_t)$; update only $\bm\psi$ with the query loss~\eqref{eq:stage2-loss}
\EndWhile
\State Fix $\bm\psi$ for reconstruction
\Statex \textbf{Reconstruction: full-mask or context ensemble}
\State Draw one $\bm\epsilon\sim\mathcal N(0,I)$
\State Set $\bm{x}_{\mathrm{in},\delta}\gets\alpha_\delta\bm{x}_{\mathrm{in}}+\sigma_\delta\bm\epsilon$
\If{inference mode is full-mask}
  \State Set $K\gets1$ and $\bm M_{\mathrm{ctx}}^{(1)}\gets\bm M_{\mathrm{in}}$
\Else
  \State Use the first $K$ masks from the cached context sequence sampled from the context marginal of $\mathcal P_{\mathrm{align}}(\cdot\mid\bm M_{\mathrm{in}})$
\EndIf
\For{$k=1,\ldots,K$}
  \State $\bm{x}_{\ctx,\delta}^{(k)}\gets\bm M_{\mathrm{ctx}}^{(k)}\odot\bm{x}_{\mathrm{in},\delta}$
  \State $(\bm\Pi_\delta^{(k)},I_\delta^{(k)})\gets\Call{BuildReference}{\delta,\bm{x}_{\ctx,\delta}^{(k)},\bm M_{\mathrm{ctx}}^{(k)}}$
  \State $\widehat{\bm x}_{0,\delta}^{(k)}\gets\bm\Pi_\delta^{(k)}+r_{\bm\psi}(\delta;I_\delta^{(k)})$
\EndFor
\State $\widehat{\bm x}_{\mathrm{out}}\gets\bm{x}_{\mathrm{in}}+(\bm V-\bm M_{\mathrm{in}})\odot K^{-1}\sum_{k=1}^K\widehat{\bm x}_{0,\delta}^{(k)}$
\Statex \textbf{Shared context-only reference construction}
\Function{BuildReference}{$t,\bm{x}_{\ctx,t},\bm M_{\mathrm{ctx}}$}
  \State Evaluate the frozen local predictor on every $p\in\mathcal P$ using Eq.~\eqref{eq:local-prediction-main}
  \State Fuse the patch predictions into $\bm\Pi_t$ by Eq.~\eqref{eq:uniform-consensus-solution}; set $\bm\Pi_t=0$ outside $\bm V$
  \State Compute $\bm S_{\ctx},\bm S_{\partial}$ from this same noisy context; form $I_t$ by Eq.~\eqref{eq:full-domain-input-main}
  \State \Return $(\bm\Pi_t,I_t)$
\EndFunction
\end{algorithmic}
\end{algorithm}

\FloatBarrier

\section{Observation-Aligned Context--Query Construction}
\label{app:aligned-partitions}

This appendix specifies training context--query construction and the subcontext rule used for context-ensemble reconstruction. Ocean validation/test input and scoring supports are constructed separately through the empirical real-mask overlays in Appendix~\ref{app:observation-scoring-protocols}.

\subsection{Partition Requirements}
\label{app:aligned-design}

The estimator forms share acquisition-aligned context--query construction: sitewise partitions for pixel observations, whole-block partitions for block observations, and mask-only proposals for authentic ocean observations \citep{zhou2026incomplete,ma2026observation}. Ocean proposals need not lie within $\bm{M}$; the intersection in Equation~\eqref{eq:authentic-mask-intersection} restricts context and query to available measurements.
The dataset-specific constructors are specified below.
Marginal query frequencies describe the resulting supervision, while Appendix~\ref{app:masked-objectives} defines the conditional target on states with positive query exposure.

\subsection{Controlled Pixel and Block Partitions}
\label{app:controlled-partitions}

At each training visit, candidate context--query partitions use independent retention proposals within the stored outer support:
\begin{equation}
\begin{aligned}
B_i^{(n,k)}&\sim\operatorname{Bernoulli}(0.7),\\
\bm{M}_{\mathrm{ctx}}^{(n,k)}&=\bm{M}^{(n)}\odot B^{(n,k)},\qquad
\bm{M}_{\mathrm{qry}}^{(n,k)}=\bm{M}^{(n)}\odot(1-B^{(n,k)}).
\end{aligned}
\label{eq:aligned-pixel-context-query}
\end{equation}
Before rejecting empty queries, the expected context and query fractions conditional on $\bm{M}$ are $0.7\rho(\bm{M})$ and $0.3\rho(\bm{M})$, respectively, giving nominal fractions of approximately 21\% and 9\%.
Empty queries are handled by resampling the inner partition. Context-ensemble inference reuses this inner context-retention rule.

For Block 3/9, two of the three available blocks form the context and the remaining block forms the query. For Block 8/9, seven of the eight available blocks form the context and the remaining block forms the query.

\subsection{Authentic-Mask Constructors}
\label{app:bfn-construction}
\label{app:bfn-guidance}
\label{app:aligned-datasets}

\paragraph{Guided BFN partitions for Black Sea and Baltic.}
Training context--query partitions for Black Sea and Baltic use dataset-specific mask-only Bayesian Flow Networks \citep{graves2023bfn} with the observation-aligned guided sampler of OAMP \citep{ma2026observation}. Each BFN is trained only on masks from the
corresponding training split, selected by validation-mask prediction accuracy,
and frozen before physical-field training. The current observation mask is
intersected with the guided sample to produce context and query supports.

We expand the inherited OAMP sampling procedure here to distinguish its mask
generation state from the physical-field diffusion state.
Let $\bm{z}_\tau$ denote the continuous mask-sampling logits at time $\tau$,
and let $p_{\mathrm{BFN},i}(\tau,\bm{z}_\tau)$ denote the frozen BFN's predicted
probability of the observed class at spatial location $i$ (with the softmax
mapping of the logits included in this notation).
For each proposal, draw independent $b_i\sim\operatorname{Bernoulli}(\rho)$
and set $y_i=M_i b_i$.
Following \citet[Section 3.4]{ma2026observation}, the guidance objective is
\begin{equation}
\mathcal L_{\mathrm{guide}}(\bm{z}_\tau,\bm{y})
=-\frac{1}{|\Omega|}\sum_{i\in\Omega}
\left[y_i\log p_{\mathrm{BFN},i}(\tau,\bm{z}_\tau)
+(1-y_i)\log\!\left(1-p_{\mathrm{BFN},i}(\tau,\bm{z}_\tau)\right)\right].
\label{eq:mask-guidance-bce}
\end{equation}
The normalization is over the entire spatial grid, not only the retained
anchor sites. Given the base reverse-sampling update
$\bm{z}^{\mathrm{base}}_{\tau_{k+1}}$, apply
\begin{equation}
\bm{z}_{\tau_{k+1}}
=\bm{z}^{\mathrm{base}}_{\tau_{k+1}}
-w_g\nabla_{\bm{z}_{\tau_k}}
\mathcal L_{\mathrm{guide}}(\bm{z}_{\tau_k},\bm{y}).
\label{eq:mask-guidance-update}
\end{equation}
Here $w_g$ controls guidance strength; gradients update the sampling state while the BFN remains fixed. The anchor remains fixed within a
sampling trajectory. Final categorical decoding produces a binary proposal,
which is then intersected with $\bm{M}$ as in
Equation~\eqref{eq:authentic-mask-intersection}.
Neither anchor retention nor soft guidance fixes the final intersection size.

\begin{algorithm}[!t]
\caption{Guided BFN context--query partition for Black Sea CHL and Baltic Sea NANO}
\label{alg:guided-bfn-partition}
\begin{algorithmic}[1]
\Require current mask $\bm{M}$, frozen BFN, $\rho=0.7$, $w_g=80$, $S=20$;
uniform time grid $\tau_0=10^{-3}<\cdots<\tau_S=1$; run-level random-number state initialized with seed 42
\State Set $m_{\mathrm{ctx}}=m_{\mathrm{qry}}=\max\!\left(16,\left\lceil0.05\|\bm{M}\|_1\right\rceil\right)$
\For{$r=0,\ldots,8$} \Comment{Initial proposal and at most eight additional retries}
\State Draw $b_i\sim\operatorname{Bernoulli}(\rho)$ and set $y_i\gets M_i b_i$
\State Initialize mask logits $\bm{z}_{\tau_0}\sim\mathcal N(0,I)$
\For{$k=0,\ldots,S-1$}
  \State Set $w_g^{(k)}\gets0.4w_g$ if $k<0.1S$, and $w_g^{(k)}\gets w_g$ otherwise
  \State Compute the base BFN ODE update $\bm{z}^{\mathrm{base}}_{\tau_{k+1}}$
  \State Evaluate $\mathcal L_{\mathrm{guide}}(\bm{z}_{\tau_k},\bm{y})$ by Equation~\eqref{eq:mask-guidance-bce}
  \State Update $\bm{z}_{\tau_{k+1}}$ by Equation~\eqref{eq:mask-guidance-update} using $w_g^{(k)}$
\EndFor
\State Evaluate the frozen BFN once more at $\tau_S$ and decode its categorical argmax to obtain $\widetilde{\bm{M}}$
\State $\bm{M}_{\mathrm{ctx}}\gets \bm{M}\odot\widetilde{\bm{M}}$
\State $\bm{M}_{\mathrm{qry}}\gets \bm{M}\odot(1-\widetilde{\bm{M}})$
\If{$\|\bm{M}_{\mathrm{ctx}}\|_1>0\;\land\;\|\bm{M}_{\mathrm{qry}}\|_1>0$}
  \State \textbf{break}
\EndIf
\EndFor
\If{$\|\bm{M}_{\mathrm{ctx}}\|_1<m_{\mathrm{ctx}}$ or $\|\bm{M}_{\mathrm{qry}}\|_1<m_{\mathrm{qry}}$}
  \State Reassign pixels between context and query within $\bm{M}$ to meet both minimum sizes
\EndIf
\If{$\|\bm{M}_{\mathrm{ctx}}\|_1<m_{\mathrm{ctx}}$ or $\|\bm{M}_{\mathrm{qry}}\|_1<m_{\mathrm{qry}}$}
  \State Raise an error
\EndIf
\State \Return $(\bm{M}_{\mathrm{ctx}},\bm{M}_{\mathrm{qry}})$
\end{algorithmic}
\end{algorithm}

The retry loop requires both context and query to be nonempty; the subsequent minimum-size check also applies after an early exit. Anchor retention does not set the final context-retention fraction, which is determined by categorical decoding and intersection with the available support. The run-level random-number state advances across proposals without reseeding. The mask sampler's reverse steps are distinct from the near-zero-time, single-step field prediction used for each evaluated context. Table~\ref{tab:bfn-sampling-settings} records the sampling configuration.

\begin{table}[!htb]
\centering
\footnotesize
\setlength{\abovecaptionskip}{0pt}
\caption{Guided BFN sampling configuration for Black Sea CHL and Baltic Sea NANO.}
\label{tab:bfn-sampling-settings}
\setlength{\tabcolsep}{3pt}
\renewcommand{\arraystretch}{1.15}
\begin{tabularx}{\linewidth}{@{}>{\raggedright\arraybackslash}p{0.30\linewidth}>{\raggedright\arraybackslash}X@{}}
\toprule
Setting & Value \\
\midrule
Anchor retention $\rho$ & \textbf{0.7} \\
Guidance scale $w_g$ & \textbf{80}; \textbf{32} for the first two updates \\
Sampling steps $S$ & \textbf{20} \\
Reverse solver / schedule & Guided BFN ODE; uniform time grid $t:10^{-3}\to1$ \\
Sampling seed & \textbf{42}; run-level random-number state \\
Empty-support handling & Initial proposal plus at most 8 retries; post-loop minimum-size repair \\
\bottomrule
\end{tabularx}
\end{table}

\paragraph{Empirical authentic-track partitions for Global Ocean SSS.}
Global Ocean SSS draws a proposal from the empirical authentic-track mask bank and intersects it with the current available support as in Equation~\eqref{eq:authentic-mask-intersection}, producing the context and the remaining query support.

\section{Theoretical Analysis and Proofs}
\label{app:formal-proofs}

We first establish a distributional limitation of never-complete observations, then characterize the query-supervised target, the support of the local consensus prior, and the role of full-domain residual revision.

\subsection{Distributional Non-Identifiability}
\label{app:distribution-identifiability}

This subsection states and proves the distributional limitation discussed in Section~\ref{sec:complete-law-identifiability}.
The result holds even under a fixed, known mask law independent of the field.
This independence assumption applies only to the theorem; ocean experiments retain authentic acquisition masks.
Let $\bm{x}_0\in\R^d$ denote a field vectorized over its valid domain.

\begin{samepage}
\begin{theorem}[Non-identifiability under never-complete masking]
\label{prop:never-complete-nonidentifiability}
Suppose that $\bm{M}$ has a fixed, known distribution independent of $\bm{x}_0$, and that $\Pr(\bm{M}=\mathbf1_d)=0$.
Over the unrestricted class of probability laws on $\R^d$, the complete-field distribution is not identifiable from $(\bm{M}\odot \bm{x}_0,\bm{M})$.
Consequently, no algorithm can consistently recover that distribution for every law in this class.
\end{theorem}
\end{samepage}

The proof constructs two distinct complete-field laws whose marginals agree on every proper subset of coordinates, making their incomplete observations indistinguishable.

\paragraph{Observation operator and identifiability.}
\label{app:observation-identifiability}

Write $\bm{Y}=\bm{M}\odot \bm{x}_0$ and let $q$ be the fixed, known mask law, independent of $\bm{x}_0$.
The observation operator is
\begin{equation}
\mathcal T_q(P_{\bm{x}_0})=\operatorname{Law}_{P_{\bm{x}_0},q}(\bm{M}\odot \bm{x}_0,\bm{M}).
\label{eq:identifiability-operator}
\end{equation}
Complete-law identifiability over a class $\mathcal C$ means
\begin{equation}
\mathcal T_q(P_1)=\mathcal T_q(P_2)\quad\Longrightarrow\quad P_1=P_2
\qquad\text{for all }P_1,P_2\in\mathcal C.
\label{eq:identifiability-definition}
\end{equation}

\paragraph{Parity construction and proper-subset marginals.}

Let $\bm{x}_{0,i}\in\{-1,+1\}$ and define the two parity supports
\begin{equation}
\mathcal X_+=\left\{\bm{x}:\prod_{i=1}^d \bm{x}_i=+1\right\},
\qquad
\mathcal X_-=\left\{\bm{x}:\prod_{i=1}^d \bm{x}_i=-1\right\}.
\label{eq:parity-supports}
\end{equation}
Let $P_+$ and $P_-$ be the uniform distributions on $\mathcal X_+$ and
$\mathcal X_-$, respectively. The supports are disjoint, so $P_+\neq P_-$, and
each support contains $2^{d-1}$ points.

Fix any proper subset $S\subsetneq[d]$ and any assignment $\bm{x}_S$. At least one
coordinate remains unassigned. Under either required parity, choose all but one
of the remaining signs freely; the final sign is then uniquely determined by
the parity constraint. Hence $\bm{x}_S$ has exactly $2^{d-|S|-1}$ valid completions
under each parity law. Dividing by the support size gives
\begin{equation}
P_+(\bm{x}_{0,S}=\bm{x}_S)=P_-(\bm{x}_{0,S}=\bm{x}_S)
=\frac{2^{d-|S|-1}}{2^{d-1}}=2^{-|S|}.
\label{eq:proper-subset-uniform-marginal}
\end{equation}
Therefore every proper-subset marginal is uniform and identical under $P_+$
and $P_-$, even though the complete laws have opposite $d$-way parity.

\paragraph{Equality of the observed laws.}

For a mask value $m$ with $q(m)>0$, let $S_m=\{i:m_i=1\}$. The assumption
$\Pr_q(\bm{M}=\1_d)=0$ implies $S_m\subsetneq[d]$. Conditional on $\bm{M}=m$, the
nonzero coordinates of $\bm{Y}=\bm{M}\odot \bm{x}_0$ have the $\bm{x}_{0,S_m}$ marginal, while all
remaining coordinates are zero. Equation~\eqref{eq:proper-subset-uniform-marginal}
therefore yields
\begin{equation}
\operatorname{Law}_{P_+}(\bm{Y}\mid \bm{M}=m)
=
\operatorname{Law}_{P_-}(\bm{Y}\mid \bm{M}=m).
\label{eq:parity-conditional-observation-law}
\end{equation}
Both worlds use the same mask law $q$, and $\bm{M}$ is independent of $\bm{x}_0$ in each.
Thus, for every measurable set $A$ and every $m$,
\begin{align}
\Pr_{P_+,q}(\bm{Y}\in A,\bm{M}=m)
&=q(m)\Pr_{P_+}(\bm{Y}\in A\mid \bm{M}=m)\nonumber\\
&=q(m)\Pr_{P_-}(\bm{Y}\in A\mid \bm{M}=m)
=\Pr_{P_-,q}(\bm{Y}\in A,\bm{M}=m).
\label{eq:parity-joint-observation-law}
\end{align}
Consequently,
$\mathcal T_q(P_+)=\mathcal T_q(P_-)$ while $P_+\neq P_-$, proving the
non-identifiability claim.

\paragraph{Consequence for universal recovery algorithms.}

Suppose an algorithm estimates the complete-field law by $\widehat P_n$ from $n$ i.i.d. observed pairs.
Its inputs have identical distributions under $P_+$ and $P_-$ for every $n$.
Coupling any internal randomization identically gives the same output distribution in both cases.
Thus $\widehat P_n$ cannot converge to $P_+$ under one law and the distinct $P_-$ under the other, ruling out universal consistency.

\paragraph{Two-dimensional example and missing-coordinate prediction.}
\label{app:two-dimensional-identifiability-example}
The ambiguity already appears when $d=2$. Let $\bm{M}=(1,0)$ or $(0,1)$, each with
probability $1/2$ independently of $\bm{x}_0$, and define
\begin{align}
P_1(\bm{x}_0=(1,1))=P_1(\bm{x}_0=(-1,-1))&=\tfrac12,
\label{eq:two-dimensional-p-one}\\
P_2(\bm{x}_0=(1,-1))=P_2(\bm{x}_0=(-1,1))&=\tfrac12.
\label{eq:two-dimensional-p-two}
\end{align}
The laws have opposite dependence: $\bm{x}_{0,1}\bm{x}_{0,2}=+1$ almost surely under $P_1$ and $-1$ under $P_2$.
Yet each coordinate is uniform on $\{-1,+1\}$ in both laws.
Since each mask reveals only one coordinate and has the same probability in both cases, $\mathcal T_q(P_1)=\mathcal T_q(P_2)$.

When only $\bm{x}_{0,1}$ is observed, the two observationally indistinguishable laws imply different missing-coordinate conditional means: $\mathbb E_{P_1}[\bm{x}_{0,2}\mid \bm{x}_{0,1}]=\bm{x}_{0,1}$ and $\mathbb E_{P_2}[\bm{x}_{0,2}\mid \bm{x}_{0,1}]=-\bm{x}_{0,1}$.
This ambiguity concerns prediction without joint exposure of $\bm{x}_{0,1}$ and $\bm{x}_{0,2}$, whereas the supervised conditional mean in Appendix~\ref{app:masked-objectives} is defined on states with positive query exposure.

\subsection{Query-Supervised Conditional Targets}
\label{app:masked-objectives}

The no-prior direct, prior-conditioned direct, and prior-conditioned residual estimators in Section~\ref{sec:exp-estimator-organizations} share the observation-aligned query support and pooled masked squared-error objective.
Their output forms determine whether the trainable network predicts the total field or its discrepancy from the frozen prior.
The partition constructor is fixed across these comparisons, and the unrestricted target on positive-exposure states is the observed-query conditional mean defined below.

For a predictor receiving $(t,O_t)$, with $O_t=(\bm{x}_{\ctx,t},\bm{M}_{\mathrm{ctx}})$, the unrestricted squared-loss target is
\begin{equation}
\mu_{O,i}(t,O_t)
=\mathbb E_{\mathrm{train}}[\bm{x}_{0,i}\mid t,O_t,\bm{M}_{\mathrm{qry},i}=1],
\label{eq:observed-query-estimand}
\end{equation}
on states with $\Pr(\bm{M}_{\mathrm{qry},i}=1\mid t,O_t)>0$, assuming finite relevant second moments.
The query-selection event remains in this expression: the empirical authentic-mask setting is not assumed to satisfy additional conditions that would allow dropping it.
Here $\bm{x}_{0,i}$ denotes the clean field value (or channel vector) at coordinate $i$.
When query exposure is zero, the loss does not constrain the prediction at that state and coordinate.

\subsection{Local Support of the Consensus Prior}
\label{app:local-support}
\label{sec:local-support-information}
\label{sec:constructed-prior-support}
Fix the trained predictors, patch covering, coordinate atlas, validity mask, and preprocessing operators.
Although consensus assembles a full-sized field, the context used by each output is determined by the supports of its contributing patches.
For a block $B$, the union of contributing patch supports is
\begin{equation}
U(B)=\bigcup_{p:\Omega_p\cap B\neq\varnothing}\Omega_p.
\label{eq:prior-block-support-union}
\end{equation}
For a valid coordinate $i$, write $U_i=U(\{i\})$.
Each contributing prediction depends only on the context inside its patch, so their uniform average $\bm{\Pi}_{t,i}$ depends only on the context in $U_i$.
Consequently, two states at the same diffusion time that agree on $U(B)$ produce identical reference fields on $B$.
Overlap reconciles the local predictions but leaves observations beyond their combined support unused for this prediction.
The full-domain corrector can use those observations to revise the estimate and can also correct errors in the local predictions themselves.
This gives the support boundary for the paired remote-observation interventions in Section~\ref{sec:remote-evidence} and Appendix~\ref{app:remote-evidence}; in particular, deleting supplied observations outside $U(B)$ leaves $\bm{\Pi}_{\delta,B}$ unchanged.

\subsection{Residual Revision and Representation}
\label{app:residual-representation}
\label{sec:full-domain-correction-target}
Let $\mathcal L_i=\sigma(t,R_{U_i}\bm{x}_{\ctx,t},R_{U_i}\bm{M}_{\mathrm{ctx}})$ and $\mathcal G=\sigma(t,\bm{x}_{\ctx,t},\bm{M}_{\mathrm{ctx}})$ denote local and full-domain information.
For a coordinate with positive query exposure, assume finite conditional second moments and define
\begin{equation}
\mu_i^L=\mathbb E[\bm{x}_{0,i}\mid\mathcal L_i,\bm{M}_{\mathrm{qry},i}=1],
\qquad
\mu_i^G=\mathbb E[\bm{x}_{0,i}\mid\mathcal G,\bm{M}_{\mathrm{qry},i}=1].
\label{eq:local-full-conditional-means}
\end{equation}
For a fixed square-integrable reference $\bm{\Pi}_{t,i}$, the conditional mean minimizes squared error under the measure conditioned on $\bm{M}_{\mathrm{qry},i}=1$.
Adding and subtracting $\mu_i^L$ from the optimal unrestricted full-domain correction gives
\begin{equation}
 r_i^\star=\mu_i^G-\bm{\Pi}_{t,i}
 =\underbrace{\mu_i^G-\mu_i^L}_{\text{adjustment beyond local support}}
 +\underbrace{\mu_i^L-\bm{\Pi}_{t,i}}_{\text{remaining local-estimation error}}.
\label{eq:full-domain-correction-target-decomposition}
\end{equation}
The first term accounts for evidence unavailable to the local construction; the second accounts for error that can remain even within its support.
Thus the prior need not be an optimal local predictor to serve as a useful reference for correction.
The query-selection condition is retained because authentic acquisition and partition rules need not expose all prediction states in the same way.
This identity explains that revision can address both missing nonlocal evidence and local-estimation error; it does not guarantee that a trained finite corrector attains this target or outperforms direct prediction.

\paragraph{Scope of the representation.}
With the trained constructor fixed, the prior and discrepancy maps are deterministic functions of $(t,O_t)$, and $O_t$ is retained in $I_t$.
Thus $\sigma(t,I_t)=\sigma(t,O_t)=\mathcal G$: unrestricted direct and residual predictors have the same optimal total-field prediction and query risk.
The empirical comparisons in Sections~\ref{sec:exp-estimator-organizations} and \ref{sec:exp-coordinates-discrepancies} test how finite models learn with this representation and initialization, while the remote interventions diagnose responses to observations beyond local support.

\section{Experimental Setup and Evaluation}
\label{app:experimental-details}

This appendix specifies the datasets, preprocessing, observation and scoring protocols, evaluation metrics, and estimator controls used in Section~\ref{sec:experiments}.

\subsection{Datasets and Preprocessing}
\label{app:dataset-settings}
\label{app:data}

\noindent\textbf{Ocean Products.}
\label{app:ocean-data}
\label{app:copernicus-provenance}
We use daily Level-3 products from the E.U. Copernicus Marine Environment
Monitoring Service (CMEMS) \citep{letraon2017cmems} and derive observation
supports from their quality-controlled source masks. Black Sea CHL and Baltic Sea NANO are multi-sensor
ocean-colour observations at approximately 1-km native spacing; Global Ocean
SSS uses SMOS salinity observations on the global cylindrical EASE-Grid 2.0,
with approximately 25-km spacing at $30^\circ$ latitude. Longitude increments
are approximately $0.259^\circ$, while latitude increments vary with latitude.
The reconstructed variables are chlorophyll-a, nanophytoplankton expressed as
chlorophyll (both in $\mathrm{mg}\,\mathrm{m}^{-3}$), and practical
sea-surface salinity, respectively. The regional
products contain cloud-related gaps, while SSS exhibits orbital-track and
inter-swath gaps. Table~\ref{tab:ocean-sources} identifies the sources and requested geographic extents; Black Sea CHL uses the full regional product domain, and Global Ocean SSS uses the ascending-orbit dataset.
The records span approximately 26 years for Black Sea CHL, 28 years for Baltic Sea NANO, and 16 years for Global Ocean SSS.

This study has been conducted using E.U. Copernicus Marine Service Information;
\href{https://doi.org/10.48670/moi-00303}{\nolinkurl{10.48670/moi-00303}},
\href{https://doi.org/10.48670/moi-00296}{\nolinkurl{10.48670/moi-00296}}, and
\href{https://doi.org/10.48670/mds-00368}{\nolinkurl{10.48670/mds-00368}}.

\begin{table*}[!hbp]
\centering
\footnotesize
\setlength{\tabcolsep}{4pt}
\caption{Ocean data products and geographic subsets.}
\label{tab:copernicus-provenance}
\label{tab:ocean-sources}
\begin{tabularx}{\textwidth}{@{}l *{3}{>{\raggedright\arraybackslash}X}@{}}
\toprule
 & Black Sea CHL & Baltic Sea NANO & Global Ocean SSS \\
\midrule
Product ID
& \path{OCEANCOLOUR_BLK_BGC_L3_MY_009_153}
& \path{OCEANCOLOUR_BAL_BGC_L3_MY_009_133}
& \path{MULTIOBS_GLO_PHY_SSS_L3_MYNRT_015_014} \\
DOI
& \href{https://doi.org/10.48670/moi-00303}{\path{10.48670/moi-00303}}
& \href{https://doi.org/10.48670/moi-00296}{\path{10.48670/moi-00296}}
& \href{https://doi.org/10.48670/mds-00368}{\path{10.48670/mds-00368}} \\
Variable
& \path{CHL} & \path{NANO} & \path{Sea_Surface_Salinity} \\
Spatial subset
& Full Black Sea domain; approximately $26.5^\circ$--$42^\circ$E,
  $40^\circ$--$48^\circ$N
& $15^\circ$--$25^\circ$E, $54^\circ$--$59^\circ$N
& $180^\circ$W--$179.8^\circ$E, $83.6^\circ$S--$83.6^\circ$N \\
Native grid
& Regional grid; approximately 1-km spacing
& Regional grid; approximately 1-km spacing
& Global cylindrical EASE-Grid 2.0 \\
\bottomrule
\end{tabularx}
\end{table*}
\FloatBarrier

\begin{samepage}
After source-level quality control and
before empty-field or sparse-field filtering, the mean observed-ocean
fractions over all dates, including empty fields, are 38.30\%, 26.50\%, and
30.60\% for Black Sea CHL, Baltic Sea NANO, and Global Ocean SSS, respectively.
For $N$ fields, we compute the per-field average
\[
\bar r=\frac{1}{N}\sum_{n=1}^{N}
\frac{|\bm{M}_n\cap \bm{V}_n|}{|\bm{V}_n|},
\]
where $\bm{M}_n$ is the quality-controlled observation support and $\bm{V}_n$ is the
empirical ocean support inferred from locations with valid data at least once
in the temporal record. This support is defined by data availability rather
than an independent coastline map. Each date receives equal weight, and the
fractions are computed before context--query partitioning.
\end{samepage}

Ocean samples are assigned chronologically to 70/15/15 training, validation,
and test splits before resolution-specific filtering. We remove empty samples
and retain fields with at least 128 observed pixels at $64\times64$ or 512
at $128\times128$. Fields and masks are resampled jointly using value-weighted area resampling, and every retained ocean field remains incomplete on its valid domain. SSS retains source values with
\path{Sea_Surface_Salinity_QC} equal to zero. Black Sea
CHL and Baltic Sea NANO use $\log(\bm{x}+10^{-6})$, while SSS uses the original
salinity. At each resolution, normalization statistics are fitted to the available training observations and held fixed for validation, testing, and patch evaluation.
Values withheld by the evaluation protocol or unavailable under the observation masks are outside this fitting support.
Normalized values are clipped to three standard deviations and divided by three, and invalid locations are set to zero.

\paragraph{Controlled PDE systems.}
\label{app:pde-data}
Shallow Water uses water-height fields from the PDEBench two-dimensional radial dam-break benchmark \citep{takamoto2022pdebench}.
Navier--Stokes uses the released $64\times64$ vorticity fields of forced two-dimensional Kolmogorov turbulence at $\mathrm{Re}=1000$ \citep{cao2024navierstokes,cao2025spectralrefiner}, specifically the file \path{Kolmogorov2d_fp32_64x64_N1152_Re1000_T100.pt}.

For Shallow Water, we retain one representative from each of the 81 distinct rasterized initial-state classes in the 1,000-trajectory source and assign 57/12/12 trajectories to training/validation/test splits.
Navier--Stokes uses all 1,152 released trajectories with an 806/173/173 split.
Trajectory selection and splitting use seed 42.
Each trajectory is assigned to a single split before all stored frames are treated as individual spatial reconstruction samples, with no temporal windows or further temporal subsampling.
Shallow Water contributes 101 frames per trajectory and uses height channel 0, bilinearly resized from $128\times128$ to $64\times64$ with antialiasing; Navier--Stokes contributes 100 frames per trajectory at its released $64\times64$ resolution.
PDE normalization and clipping statistics are fitted to observed values within the stored training acquisition masks and held fixed for subsequent patch and full-domain evaluations.

\subsection{Observation and Scoring Protocols}
\label{app:observation-scoring-protocols}
\label{app:authentic-evaluation-target}

\noindent\textbf{Ocean authentic-mask protocol.}
\label{app:ocean-protocol}
Training supplies every method with the source-supported observations of each sample. Our context--query training follows Appendix~\ref{app:bfn-construction}: Black Sea CHL and Baltic Sea NANO use frozen guided BFN constructors, whereas Global Ocean SSS uses an empirical authentic-track mask bank. Ocean validation and testing instead follow OAMP's real-mask overlay protocol \citep{ma2026observation}, drawing authentic observation masks directly from their respective splits rather than using randomly placed blocks.

For a fixed ocean product at resolution $L\times L$, let $\mathcal I_{s,L}$ index the samples retained after the resolution-specific filtering in Appendix~\ref{app:dataset-settings}, where $s\in\{\mathrm{val},\mathrm{test}\}$. For the current sample $n$, each proposal $\widetilde{\bm{M}}^{(r)}$ is drawn from the observation masks indexed by $\mathcal I_{s,L}\setminus\{n\}$. Thus validation and test proposals come from their respective splits, and the current sample's own mask is explicitly excluded. Suppressing the current sample index, the candidate supports are
\begin{equation}
\bm{M}_{\mathrm{in}}^{(r)}
=\bm{M}_{\mathrm{raw}}\odot\widetilde{\bm{M}}^{(r)},
\qquad
\bm{M}_{\mathrm{eval}}^{(r)}
=\bm{M}_{\mathrm{raw}}\odot\bigl(1-\widetilde{\bm{M}}^{(r)}\bigr).
\label{eq:app-ocean-eval-mask}
\end{equation}
The donor sample contributes only its observation mask; all input and scoring values come from the current sample.

\begin{table}[!b]
\centering
\footnotesize
\caption{Available training observations, training context--query partitions, validation/test inputs, and scoring supports by dataset and protocol.}
\label{tab:aligned-mask-settings}
\label{tab:observation-protocols}
\vspace{3pt}
\setlength{\tabcolsep}{3pt}
\renewcommand{\arraystretch}{1.15}
\begin{tabularx}{\linewidth}{@{}>{\raggedright\arraybackslash}p{0.12\linewidth}>{\raggedright\arraybackslash}p{0.20\linewidth}*{4}{>{\raggedright\arraybackslash}X}@{}}
\toprule
Setting & Available training support & Training context rule & Training query & Validation/test input & Scoring support \\
\midrule
Black Sea CHL & All source-supported values & Guided-mask intersection within the current support & Remaining available values & Cached support from a same-split authentic-mask overlay & Withheld source-supported values, $\bm{M}_{\mathrm{raw}}\odot(1-\bm{M}_{\mathrm{in}})$ \\
Baltic Sea NANO & All source-supported values & Guided-mask intersection within the current support & Remaining available values & Cached support from a same-split authentic-mask overlay & Withheld source-supported values, $\bm{M}_{\mathrm{raw}}\odot(1-\bm{M}_{\mathrm{in}})$ \\
Global Ocean SSS & All source-supported values & Empirical-track intersection within the current support & Remaining available values & Cached support from a same-split authentic-mask overlay & Withheld source-supported values, $\bm{M}_{\mathrm{raw}}\odot(1-\bm{M}_{\mathrm{in}})$ \\
Pixel $30\%$ & Fixed $q^{\mathrm{pix}}_{0.3}$ mask; approximately $30\%$ & Retention probability $0.7$; approximately $21\%$ of the field & Approximately $9\%$ of the field & Fixed $q^{\mathrm{pix}}_{0.3}$ mask; approximately $30\%$ & $\V\setminus \bm{M}_{\mathrm{in}}$ \\
Block $3/9$ & Three fixed blocks & Two available blocks & One available block & Three blocks & Remaining six blocks \\
Block $8/9$ & Eight fixed blocks & Seven available blocks & One available block & Eight blocks & Remaining one block \\
\bottomrule
\end{tabularx}
\end{table}

A proposal is rejected if either its input or scoring support is empty. Starting from the initial proposal $r=0$, we allow at most eight additional retries, for at most nine proposals in total, and stop the retry loop as soon as both supports are nonempty.

After the retry loop, including early termination, we apply the same minimum-size check and pixel-reassignment procedure as in Algorithm~\ref{alg:guided-bfn-partition}, with
\[
m_{\mathrm{in}}=m_{\mathrm{eval}}
=\max\!\left(16,\,\left\lceil0.05\|\bm{M}_{\mathrm{raw}}\|_1\right\rceil\right).
\]
If either support is smaller than its minimum, observed pixels are reassigned between the input and scoring supports within $\bm{M}_{\mathrm{raw}}$. The finalized pair is checked against both minima before caching and remains a disjoint partition of the current sample's source-supported observations.

For each evaluation sample, the finalized pair $(\bm{M}_{\mathrm{in}},\bm{M}_{\mathrm{eval}})$ is cached and reused across all compared methods and internal controls. Each method receives $\bm{x}_{\mathrm{in}}=\bm{M}_{\mathrm{in}}\odot\bm{x}_{\mathrm{obs}}$ and $\bm{M}_{\mathrm{in}}$, where $\bm{x}_{\mathrm{obs}}=\bm{M}_{\mathrm{raw}}\odot\bm{x}_0$. Scoring uses $\bm{M}_{\mathrm{eval}}=\bm{M}_{\mathrm{raw}}\odot(1-\bm{M}_{\mathrm{in}})$. This protocol is used for all reported ocean evaluations at both resolutions. In the default ocean reconstruction protocol, our estimator and internal controls use the entire cached $\bm{M}_{\mathrm{in}}$, without a further internal context split.

\paragraph{Controlled Pixel protocol.}
\label{app:pde-protocol}
Each stored pixel-acquisition mask is constructed from Bernoulli proposals with mask-only coverage control, excludes complete observation, and retains approximately 30\% of valid coordinates.
\begin{equation}
\bm{M}^{(n)}\sim q^{\mathrm{pix}}_{0.3},\qquad
q^{\mathrm{pix}}_{0.3}(\bm{M}=\V)=0,\qquad
\rho(\bm{M}^{(n)})=\frac{\|\bm{M}^{(n)}\|_1}{\|\V\|_1}\approx0.3.
\label{eq:aligned-pixel-outer-support}
\end{equation}
Training partitions are sampled within each fixed acquisition support. Validation and test supports are independently constructed from the same acquisition law and then fixed for evaluation. Internal partitions follow Appendix~\ref{app:controlled-partitions}.

\paragraph{Controlled Block protocols.}
Block 3/9 and Block 8/9 retain three and eight blocks of a regular $3\times3$ partition, respectively.
Grid lines in the block-protocol visualizations mark this acquisition partition, whereas the local predictor uses the overlapping covering specified in Appendix~\ref{app:patch-geometry}.
Their internal context--query partitions preserve whole-block units (Appendix~\ref{app:controlled-partitions}).
For both Pixel and Block protocols, evaluation scores all valid complete-reference coordinates outside $\bm{M}_{\mathrm{in}}$, while our estimator and internal controls use the default $K=8$ context ensemble specified in Appendix~\ref{app:full-support-inference}.
Table~\ref{tab:observation-protocols} summarizes the observation and scoring protocols. Appendix~\ref{app:bfn-construction} specifies the training context--query constructors; the ocean validation/test overlays are defined above.

Ocean evaluation scores source-supported values withheld from the estimator, whereas controlled PDE evaluation scores valid unsupplied coordinates using complete references.
Reconstructions at authentically source-missing ocean locations lack independent quantitative references.

\subsection{Evaluation Metrics and Seed Aggregation}
\label{app:pooled-mse}
Training objectives and reconstruction comparisons use masked squared errors, with evaluation MSE pooled over the scored spatial-channel values.
For a finite collection $\mathcal D$ with estimator
states $\mathcal S^{(b)}$, target fields $\bm{x}_0^{(b)}$, and nonempty scoring masks
$A^{(b)}$, define
\begin{equation}
    \operatorname{MSE}_{A}(f;\mathcal D)
    =
    \frac{
      \sum_{b\in\mathcal D}
      \left\|
        A^{(b)}\odot
        \left(f(\mathcal S^{(b)})-\bm{x}_0^{(b)}\right)
      \right\|_2^2
    }{
      d_x\sum_{b\in\mathcal D}\left\|A^{(b)}\right\|_1
    }.
    \label{eq:app-evaluation-mse}
\end{equation}
Evaluation uses $A=\bm{M}_{\mathrm{eval}}$, and pooling over the complete collection gives equal weight to every scored spatial-channel value.
For our estimator and internal controls, the scored prediction is the final reconstruction under the dataset-specific inference protocol in Section~\ref{sec:method-inference} and Appendix~\ref{app:full-support-inference}.

PSNR is computed for each sample and then averaged over the evaluation collection.
For sample $n$, define
\[
e_n
=
\frac{
\left\|
A^{(n)}\odot
\left(\widehat{\bm{x}}^{(n)}-\bm{x}_0^{(n)}\right)
\right\|_2^2
}{
d_x\left\|A^{(n)}\right\|_1
}.
\]
The reported PSNR is
\begin{equation}
\operatorname{PSNR}
=
\frac{1}{N}\sum_{n=1}^{N}
10\log_{10}\!\left(\frac{R_n^2}{e_n}\right),
\qquad
R_n=
\begin{cases}
R_n^{\mathrm{ocean}}, & \text{ocean data},\\
2, & \text{PDE data}.
\end{cases}
\label{eq:app-evaluation-psnr}
\end{equation}
Here $R_n^{\mathrm{ocean}}$ is the sample-specific dynamic range of the ocean reference in the normalized, clipped evaluation space.
Lower MSE and higher PSNR indicate better reconstruction within a setting. The two metrics use the same scoring support but differ in aggregation, so PSNR is not a direct transformation of pooled MSE.

Our estimator and internal controls share supplied observations and perturbations within each comparison; context-ensemble comparisons additionally reuse paired subcontext masks.

The learned-model entries in Tables~\ref{tab:ocean64}, \ref{tab:matched}, and \ref{tab:pde64} report means and standard deviations of the evaluation metrics across models independently trained with different random seeds.
The intervention diagnostics in Appendix~\ref{app:remote-evidence} compute paired measurements on fixed scored blocks of the final full-mask reconstructions and summarize them over evaluation items and training-seed checkpoints.

\subsection{Estimator Controls and Prior Replacement}
\label{app:retraining-reporting}

Table~\ref{tab:matched} compares the Local Prior reference with Full Direct, Prior Direct, and Prior Residual, and includes two prior-replacement controls.
On the ocean products, Full Direct is a parameter-matched version of OAMP.
The Navier--Stokes Full Direct control in Figure~\ref{fig:ns-block-detail} and the Shallow Water Full Direct control in Appendix~\ref{app:inference-cost} are defined analogously using IDCD.
In each case, the total parameter count matches the frozen local predictor plus the trainable full-domain network in our two-stage estimator.
Full Direct receives $(\bm{x}_{\mathrm{ctx},t},\bm{M}_{\mathrm{ctx}},\bm{\Gamma})$ and uses the shared observation-aligned query supervision.
The OAMP and IDCD entries in the external-baseline tables are distinct from their parameter-matched Full Direct controls.

For the ocean comparisons and the Shallow Water Pixel 30\% comparison in Appendix~\ref{app:inference-cost}, Full Direct's training time is comparable to the combined training time of our two stages.
The second-stage time includes frozen-prior construction.
Shared mask-model pretraining, where applicable, is excluded from both sides of this comparison.

Prior Direct and Prior Residual share the frozen local predictor, full-domain backbone, input-construction procedure, and checkpoint-selection procedure.
Appendix~\ref{app:implementation-inputs} specifies their output forms and initialization; Section~\ref{sec:exp-estimator-organizations} interprets the comparison.

\paragraph{Prior-replacement controls.}
Global2Global uses a full-domain OAMP prediction as the reference, and Low-rank2Global uses a DINEOF reconstruction; their standalone results are reported in Table~\ref{tab:ocean64}.
We retain the same input channels and prior--corrector connection design, recompute $\bm{S}_{\mathrm{ctx}}$ and $\bm{S}_{\partial}$ from each replacement prior, and independently retrain the corresponding full-domain residual corrector.
Global2Global tests a learned two-stage cascade, while Low-rank2Global replaces dedicated local-prior learning with a low-rank reconstruction.

All rows in Table~\ref{tab:matched} use full-mask reconstruction under Appendix~\ref{app:full-support-inference} and the shared-input and metric conventions in Appendix~\ref{app:pooled-mse}.

\paragraph{Coordinate and input-feature controls.}
\label{app:component-ablations}
Table~\ref{tab:construction-controls} in Appendix~\ref{app:prior-construction-connection-ablations} reports the final-reconstruction results. Its two panels answer different questions. In Panel A, we remove the absolute-coordinate crop from the local predictor, learn a new local model under the same patch-level query objective, freeze it, and learn a new corrector. This tests whether position information in the shared local learning task helps the complete estimator. In Panel B, the patch predictor and consensus construction remain fixed. We remove one or both deterministic input maps and learn a new residual corrector for each variant, preserving the retained context and prior itself. All variants use full-mask reconstruction under the pairing conventions in Appendix~\ref{app:pooled-mse}.

\FloatBarrier

\section{Supplementary Experiments}
\label{app:supplementary-experiments}

This appendix reports controlled PDE reconstruction, higher-resolution ocean results, prior-construction and corrector-input ablations, remote-evidence interventions, and inference-cost and observation-retention analyses.

\subsection{Controlled PDE Reconstruction}
\label{app:pde-comparisons}
\label{sec:exp-pde}

These experiments complement the authentic ocean evaluation with controlled pixel- and block-observation settings and complete references.
Data preparation, observation protocols, and evaluation metrics are specified in Appendix~\ref{app:experimental-details}.
We compare with AmbientDiff, DINDiff, MissDiff, MSM, and IDCD \citep{zhou2026incomplete}, using shared training observations, evaluation items, supplied supports, and scoring coordinates.

\noindent\textbf{Primary pixel and block settings.}
\label{app:pde-primary}
Our estimator has the lowest MSE and highest PSNR among the compared methods in Shallow Water Pixel 30\%, Shallow Water Block 3/9, and Navier--Stokes Pixel 30\% (Table~\ref{tab:pde64}).

\begin{table}[H]
\centering
\footnotesize
\caption{PDE reconstruction at $64\times64$, evaluated on valid unsupplied coordinates. Results are means $\pm$ standard deviations across independent training seeds. Bold marks the best result in each column.}
\label{tab:main-pde}
\label{tab:pde64}

\vspace{3pt}
\setlength{\tabcolsep}{2pt}
\renewcommand{\arraystretch}{1.15}
\resizebox{\linewidth}{!}{%
\begin{tabular}{@{}lcccccc@{}}
\toprule
Method & \multicolumn{2}{c}{Shallow Water Pixel $30\%$} & \multicolumn{2}{c}{Shallow Water Block $3/9$} & \multicolumn{2}{c}{Navier--Stokes Pixel $30\%$} \\
\cmidrule(lr){2-3}\cmidrule(lr){4-5}\cmidrule(lr){6-7}
 & MSE$\downarrow$\,($\times10^{-3}$) & PSNR$\uparrow$\,(dB) & MSE$\downarrow$\,($\times10^{-3}$) & PSNR$\uparrow$\,(dB) & MSE$\downarrow$\,($\times10^{-3}$) & PSNR$\uparrow$\,(dB) \\
\midrule
AmbientDiff & \meanstd{0.903}{0.018} & \meanstd{37.168}{0.092} & \meanstd{13.777}{0.025} & \meanstd{26.516}{0.035} & \meanstd{6.338}{0.448} & \meanstd{28.146}{0.311} \\
DINDiff & \meanstd{0.866}{0.033} & \meanstd{37.233}{0.005} & \meanstd{17.656}{0.106} & \meanstd{26.170}{0.017} & \meanstd{7.995}{0.017} & \meanstd{27.074}{0.009} \\
MissDiff & \meanstd{1.177}{0.089} & \meanstd{35.779}{0.388} & \meanstd{16.839}{0.164} & \meanstd{25.798}{0.106} & \meanstd{8.740}{0.467} & \meanstd{26.728}{0.260} \\
MSM & \meanstd{4.025}{0.094} & \meanstd{30.352}{0.056} & \meanstd{18.660}{0.023} & \meanstd{25.017}{0.003} & \meanstd{17.756}{0.017} & \meanstd{23.613}{0.005} \\
IDCD & \meanstd{0.880}{0.111} & \meanstd{37.223}{0.275} & \meanstd{8.001}{0.055} & \meanstd{29.279}{0.241} & \meanstd{7.545}{0.085} & \meanstd{27.359}{0.056} \\
\midrule
Ours & \bestmeanstd{0.035}{0.007} & \bestmeanstd{51.628}{0.957} & \bestmeanstd{4.647}{0.456} & \bestmeanstd{34.554}{0.343} & \bestmeanstd{4.623}{0.046} & \bestmeanstd{29.603}{0.041} \\
\bottomrule
\end{tabular}%
}
\end{table}

The Shallow Water Block 3/9 result tests a contiguous missing region: with only three of nine spatial blocks supplied, MSE decreases from $8.001$ for the strongest external baseline to $4.647$ for ours (both in units of $10^{-3}$). Figures~\ref{fig:pde-task1-ns-pixel30} and~\ref{fig:pde-task1-sw-pixel30} show Navier--Stokes and Shallow Water Pixel 30\% reconstructions, respectively, and Figure~\ref{fig:pde-task1-sw-block3} shows Shallow Water Block 3/9.

\FloatBarrier
\noindent\textbf{Additional block settings.}
\label{app:block-six-results}
For Block 8/9, eight blocks of a regular $3\times3$ partition are supplied and the remaining block is scored.

\begin{table}[!b]
\centering
\small
\caption{PDE reconstruction under Block 8/9 observations at $64\times64$. Entries are point estimates. Bold marks the best result in each column.}
\label{tab:block-six-results}
\label{tab:pde_block89}
\label{tab:pde-block89}

\vspace{3pt}
\setlength{\tabcolsep}{3pt}
\renewcommand{\arraystretch}{1.15}
\begin{tabular*}{\linewidth}{@{\extracolsep{\fill}}lcccc@{}}
\toprule
Method & \multicolumn{2}{c}{Shallow Water Block $8/9$} & \multicolumn{2}{c}{Navier--Stokes Block $8/9$} \\
\cmidrule(lr){2-3}\cmidrule(lr){4-5}
 & MSE$\downarrow$\,($\times10^{-3}$) & PSNR$\uparrow$\,(dB) & MSE$\downarrow$\,($\times10^{-3}$) & PSNR$\uparrow$\,(dB) \\
\midrule
AmbientDiff & $8.608$ & $38.859$ & $33.725$ & $21.260$ \\
DINDiff & $5.530$ & $42.014$ & $34.068$ & $21.360$ \\
MissDiff & $16.510$ & $35.989$ & $53.201$ & $19.207$ \\
MSM & $16.099$ & $31.509$ & $45.947$ & $20.013$ \\
IDCD & $4.751$ & $41.694$ & $28.444$ & $22.054$ \\
\midrule
Ours & $\mathbf{3.481}$ & $\mathbf{45.044}$ & $\mathbf{25.326}$ & $\mathbf{22.591}$ \\
\bottomrule
\end{tabular*}
\end{table}

Our estimator achieves the lowest MSE and highest PSNR in both settings (Table~\ref{tab:pde-block89}).
Relative to IDCD, MSE decreases by 26.7\% on Shallow Water and 11.0\% on Navier--Stokes.
Qualitative comparisons with DINDiff on Shallow Water and AmbientDiff on Navier--Stokes appear in Figures~\ref{fig:pde-task1-sw-block8} and~\ref{fig:pde-task1-ns-block8}, respectively.

\Needspace{3\baselineskip}
\noindent\textbf{MSM under pixel observations.}
\label{app:msm_pixel}
The two Pixel 30\% PDE settings in Table~\ref{tab:pde64} show lower reconstruction accuracy for MSM.
Table~\ref{tab:msm-differences} contrasts its partial-measurement denoising and sampling pathway with our query-supervised reconstruction pathway.

\begin{table}[H]
\centering
\caption{Mechanistic comparison of MSM and our estimator.}
\label{tab:msm-differences}
\footnotesize
\setlength{\tabcolsep}{4pt}
\renewcommand{\arraystretch}{1.15}
\begin{tabularx}{\textwidth}{@{}
  >{\raggedright\arraybackslash}p{0.19\textwidth}
  >{\raggedright\arraybackslash}X
  >{\raggedright\arraybackslash}X@{}}
\toprule
Aspect & MSM \citep{park2026msm} & Ours \\
\midrule
Training observations & Partial measurements, including noisy or subsampled measurement settings & Incomplete physical-field observations \\
\addlinespace[4pt]
Supervised coordinates & Denoising targets lie on the measurement support supplied in noisy form & Query values are withheld from the network input and supervise local and full-domain predictions \\
\addlinespace[4pt]
Learned object & Partial-measurement denoisers and associated scores & Local clean-value predictions and a full-domain residual corrector \\
\addlinespace[4pt]
Aggregation object & Lifted partial scores or corresponding denoising estimates & Overlapping spatial-patch predictions reconciled into a field-valued consensus \\
\addlinespace[4pt]
Role of full-domain information & Propagated through sampling and observation-consistency updates & Supplied to a separately trained corrector together with the local reference and discrepancy features \\
\addlinespace[4pt]
Reconstruction procedure & Iterative stochastic sampling for generation or conditional reconstruction & A local-to-global reconstruction path per selected context at near-zero diffusion time; final fields are averaged when context ensembling is used \\
\bottomrule
\end{tabularx}
\end{table}
\FloatBarrier

Pixel and block observations provide different spatial information.
In the training-mask summaries, approximately 94\% of missing sites under Pixel 30\% have an observed eight-neighbor, compared with approximately 5\% under Block 3/9.
This geometry provides widespread local evidence for conditional reconstruction.
At the same time, pixel masks fragment jointly visible neighborhoods, whereas observed blocks retain contiguous local structure that may facilitate learning joint denoising relationships.
Pixel observations can therefore favor reconstruction without providing equally coherent partial-denoising examples.
This also differs from MSM's original face-image experiments, which used $32\times32$ block masks with approximately 40\% missing entries \citep{park2026msm}.

During inference, our MSM implementation applies a sampled mask $S$ to the current noisy iterate and aggregates predictions on that support.
The denoiser receives $S$, while the actual observation mask $M_{\mathrm{in}}$ enters through observation-consistency updates.
With both masks retaining approximately 30\% of coordinates and independent of one another, $\Pr(S_i=1,M_{\mathrm{in},i}=1)\approx0.09$.
This quantity describes support overlap; the network input remains a noisy iterate rather than clean measurements.
An observation-consistency update directly changes observed coordinates, and its influence on missing coordinates is propagated by later denoising calls.
Pixel subsampling further fragments the jointly selected observation support.

The published MSM formulation interprets its aggregated score as a composite-likelihood/product-of-experts construction \citep{park2026msm}.
Its stochastic-approximation guarantee compares finite-mask sampling with the corresponding full MSM composition, rather than establishing equality with the true joint-field score.
Increasing the number of partial-score evaluations can improve coverage and stochastic approximation, but does not by itself guarantee exact conditional dependencies.
Together, fragmented co-observation and this denoising-and-composition pathway may limit how effectively the distributed observations constrain point reconstruction in the pixel protocol.

\FloatBarrier

\subsection{Higher-Resolution Ocean Reconstruction}
\label{app:resolution-additional-settings}

Each learned method is trained and its checkpoint selected separately at $128\times128$.
Our estimator uses the default patch geometry in Appendix~\ref{app:patch-geometry}, and evaluation follows the source-supported scoring protocol in Appendix~\ref{app:observation-scoring-protocols}.

\begin{table}[H]
\centering
\small
\caption{Ocean reconstruction at $128\times128$ on withheld source-supported values. Entries are point estimates. Bold marks the best result in each column.}
\label{tab:ocean128}
\label{tab:ocean-128}

\vspace{3pt}
\setlength{\tabcolsep}{3pt}
\renewcommand{\arraystretch}{1.15}
\begin{tabular*}{\linewidth}{@{\extracolsep{\fill}}lcccccc@{}}
\toprule
Method & \multicolumn{2}{c}{Black Sea CHL} & \multicolumn{2}{c}{Baltic Sea NANO} & \multicolumn{2}{c}{Global Ocean SSS} \\
\cmidrule(lr){2-3}\cmidrule(lr){4-5}\cmidrule(lr){6-7}
 & MSE$\downarrow$\,($\times10^{-3}$) & PSNR$\uparrow$\,(dB) & MSE$\downarrow$\,($\times10^{-3}$) & PSNR$\uparrow$\,(dB) & MSE$\downarrow$\,($\times10^{-3}$) & PSNR$\uparrow$\,(dB) \\
\midrule
DINEOF & $21.806$ & $17.17$ & $38.067$ & $14.57$ & $51.323$ & $13.02$ \\
AmbientDiff & $12.594$ & $19.37$ & $30.786$ & $15.45$ & $55.249$ & $13.52$ \\
DINDiff & $43.359$ & $15.22$ & $42.796$ & $14.27$ & $72.747$ & $12.06$ \\
MissDiff & $31.779$ & $15.27$ & $51.621$ & $12.94$ & $50.593$ & $13.19$ \\
MSM & $30.211$ & $16.06$ & $34.102$ & $15.07$ & $45.435$ & $13.71$ \\
OAMP & $10.282$ & $20.20$ & $26.825$ & $15.81$ & $37.698$ & $14.43$ \\
\midrule
Ours & $\mathbf{6.935}$ & $\mathbf{22.00}$ & $\mathbf{21.102}$ & $\mathbf{16.87}$ & $\mathbf{30.230}$ & $\mathbf{15.42}$ \\
\bottomrule
\end{tabular*}
\end{table}

Table~\ref{tab:ocean-128} shows the lowest MSE and highest PSNR for our estimator on all three ocean products.
Relative to OAMP, the lowest-MSE baseline on each product, MSE decreases by 32.5\%, 21.3\%, and 19.8\% on Black Sea CHL, Baltic Sea NANO, and Global Ocean SSS, respectively.

\FloatBarrier

\subsection{Prior Construction and Connection Ablations}
\label{app:prior-construction-connection-ablations}

This subsection examines local-coordinate conditioning, corrector inputs, fusion weights, covering geometry, and patch scale. The comparisons reuse the same full supplied mask and Gaussian perturbation for each evaluation item.

\paragraph{Coordinate and input-feature controls.}
Table~\ref{tab:construction-controls} reports the final-reconstruction effects of local-coordinate conditioning and explicit corrector inputs; Appendix~\ref{app:retraining-reporting} specifies the corresponding retraining procedures. Removing local coordinates raises final MSE by 1.7\%, 2.9\%, and 14.2\% on Black Sea CHL, Baltic Sea NANO, and Global Ocean SSS, respectively (Panel A). Section~\ref{sec:exp-coordinates-discrepancies} interprets the input-feature results in Panel B.

\begin{table}[!b]
\centering
\footnotesize
\caption{Local-coordinate and corrector-input ablations for final full-mask reconstruction on $64\times64$ ocean fields. Bold marks the best result in each column within each panel.}
\label{tab:component-ablations}
\label{tab:construction-controls}
\label{tab:coordinate-ablation-full}
\label{tab:discrepancy-ablation-full}
\vspace{3pt}
\setlength{\tabcolsep}{0.8pt}
\renewcommand{\arraystretch}{1.15}
\begin{tabular*}{\linewidth}{@{\extracolsep{\fill}}>{\raggedright\arraybackslash}p{0.20\linewidth}cccccc@{}}
\toprule
Variant & \multicolumn{2}{c}{Black Sea CHL} & \multicolumn{2}{c}{Baltic Sea NANO} & \multicolumn{2}{c}{Global Ocean SSS} \\
\cmidrule(lr){2-3}\cmidrule(lr){4-5}\cmidrule(lr){6-7}
 & MSE$\downarrow$\,($\times10^{-3}$) & PSNR$\uparrow$\,(dB) & MSE$\downarrow$\,($\times10^{-3}$) & PSNR$\uparrow$\,(dB) & MSE$\downarrow$\,($\times10^{-3}$) & PSNR$\uparrow$\,(dB) \\
\midrule
\multicolumn{7}{l}{\textit{A: Local coordinate conditioning}} \\
With local coordinates & $\mathbf{7.164}$ & $\mathbf{21.79}$ & $\mathbf{19.345}$ & $\mathbf{17.37}$ & $\mathbf{26.798}$ & $\mathbf{15.93}$ \\
No local coordinates & $7.286$ & $21.66$ & $19.910$ & $17.17$ & $30.614$ & $15.33$ \\
\midrule
\multicolumn{7}{l}{\textit{B: Explicit prior--evidence representation}} \\
Full state & $\mathbf{7.164}$ & $\mathbf{21.79}$ & $\mathbf{19.345}$ & $\mathbf{17.37}$ & $\mathbf{26.798}$ & $\mathbf{15.93}$ \\
No $\bm{W}$ & $8.007$ & $21.14$ & $20.996$ & $16.99$ & $30.077$ & $15.47$ \\
No $\bm{S}_{\ctx}$ & $8.160$ & $21.05$ & $21.355$ & $16.89$ & $29.432$ & $15.48$ \\
No $\bm{S}_{\partial}$ & $8.227$ & $21.10$ & $22.046$ & $16.79$ & $30.202$ & $15.38$ \\
No $\bm{S}_{\ctx}$ or $\bm{S}_{\partial}$ & $9.800$ & $20.14$ & $23.772$ & $16.26$ & $31.467$ & $15.20$ \\
\bottomrule
\end{tabular*}
\end{table}

\paragraph{Consensus-rule comparison.}
\label{app:consensus-rule-comparison}
\label{app:local-prior-fusion-operators}
The consensus operator reconciles patch estimates that may disagree at overlapping locations. We ask whether unequal weights improve that reconciliation, and separately whether using overlapping views is preferable to tiling the field without overlap. These comparisons evaluate the local prior constructed from the full supplied mask before full-domain correction, so they isolate construction quality rather than the ability of a retrained corrector to compensate for a changed prior.
Panel A fixes the overlapping views and their predictions under the full supplied mask and varies only the fusion weights. Panel B compares non-overlapping and half-stride coverings under uniform fusion; unlike Panel A, it changes which patch predictions are available at each location and which observations can influence them.
Fusion choices and weighting settings are selected on validation data before testing.
For rule $f$, the scored prior is the consensus field $\bm{\Pi}_\delta^{\,f}$ constructed with $\bm{M}_{\mathrm{ctx}}=\bm{M}_{\mathrm{in}}$.

For a covering $\Pset^{(s)}$, let $\Pset_i^{(s)}=\{p\in\Pset^{(s)}:i\in\Omega_p\}$.
For a field of side length $L$ and patch width $h$, non-overlap origins are
$\{0,h,\ldots,L-h\}^2$ and half-stride overlap origins are
$\{0,h/2,\ldots,L-h\}^2$. For $L=64,h=32$ and $L=128,h=64$, these give
four non-overlapping and nine overlapping views. For the default Global Ocean
SSS setting $L=64,h=16$, the non-overlap origins are $\{0,16,32,48\}^2$
(16 views), and the overlap origins are $\{0,8,16,24,32,40,48\}^2$ (49 views).
The crop operators are in-bounds direct slices.
Dataset-specific mean multiplicity is computed as
\begin{equation}
    \overline m^{(s)}
    =
    \frac{\sum_{i\in\Omega}\V_i
      |\Pset_i^{(s)}|}{\sum_{i\in\Omega}\V_i}.
    \label{eq:app-mean-overlap-multiplicity}
\end{equation}

Within a fixed inference context, let $P_k$ denote the $k$th patch and let $\hat{x}_k(p)$ be its prediction at pixel $p$.
Here $k$ indexes patches only within this fusion definition; the inference context is held fixed.
The unified fusion rule is
\begin{equation}
\hat{x}_{\mathrm{fused}}(p)=
\frac{\displaystyle\sum_{k:p\in P_k}w_k(p)\hat{x}_k(p)}
{\displaystyle\max\!\left(\sum_{k:p\in P_k}w_k(p),\,\varepsilon_f\right)},
\qquad \varepsilon_f=10^{-6}.
\label{eq:app-stabilized-fusion}
\end{equation}
The weights are nonnegative and computed from the corresponding observable state. Uniform fusion uses $w_k(p)=1$; the resulting field is $\bm{\Pi}_{t,p}^{(s)}=\hat{x}_{\mathrm{fused}}(p)$.

\paragraph{Confidence weighting.}
Let $m_k(q)\in\{0,1\}$ indicate whether pixel $q$ in patch $P_k$ is a retained context observation.
We weight each patch by its retained context fraction:
\begin{equation}
c_k=\frac{1}{|P_k|}\sum_{q\in P_k}m_k(q),
\qquad
w_k(p)=\max\!\left(c_k^{\gamma},\,w_{\min}\right).
\label{eq:app-confidence-fusion-weight}
\end{equation}
Here $\gamma=\texttt{power}$ and $w_{\min}=\texttt{min\_weight}$.
The power is applied before the lower bound.
The resulting weight is constant within each patch and increases with its retained context fraction.

\paragraph{Context-error weighting.}
For the single-channel ocean data, the uncertainty proxy is the reconstruction MSE on retained context observations:
\[
u_k=
\frac{\displaystyle\sum_{q\in P_k}m_k(q)
\bigl(\hat{x}_k(q)-x_{\mathrm{obs}}(q)\bigr)^2}
{\displaystyle\sum_{q\in P_k}m_k(q)}.
\]
We then use
\begin{equation}
w_k(p)=\operatorname{clip}\!\left(
(u_k+\varepsilon_u)^{-\gamma},\,w_{\min},\,w_{\max}
\right).
\label{eq:app-uncertainty-fusion-weight}
\end{equation}
Here $\varepsilon_u=\texttt{eps}$; $\gamma$, $w_{\min}$, and $w_{\max}$ correspond to \texttt{power}, \texttt{min\_weight}, and \texttt{max\_weight}, respectively.
Only retained context entries enter $u_k$; ground-truth values at coordinates to be imputed are not used.
Patches without context observations are assigned $u_k=+\infty$ and receive the minimum weight.
The weight is constant within each patch. The table row ``Uncertainty-weighted fusion uses this retained-context error proxy.

\paragraph{Gaussian weighting.}
Let $\xi_k(p),\eta_k(p)\in[-1,1]$ denote normalized coordinates within patch $P_k$, and define
\[
\sigma=\max(\texttt{sigma\_scale},10^{-6}),
\qquad
g_k(p)=\exp\!\left(
-\frac{\xi_k(p)^2+\eta_k(p)^2}{2\sigma^2}
\right).
\]
The window is lower-bounded before normalization by its maximum:
\begin{equation}
\tilde g_k(p)=\max\!\left(g_k(p),w_{\min}\right),
\qquad
w_k(p)=\frac{\tilde g_k(p)}
{\max\!\left(\max_{q\in P_k}\tilde g_k(q),10^{-12}\right)}.
\label{eq:app-gaussian-fusion-weight}
\end{equation}
Here $w_{\min}=\texttt{min\_weight}$.
The weights vary within a patch, emphasizing its center over its edges; smaller \texttt{sigma\_scale} concentrates weights more strongly at the center.
The unified fusion operator then normalizes these weights across patches covering each pixel.

\suppressfloats[t]
\begin{table}[H]
\centering
\small
\caption{Local-prior reconstruction before global correction using full supplied masks on $64\times64$ ocean fields. Bold marks the lowest MSE within each dataset and panel.}
\label{tab:main-fusion}
\label{tab:consensus-rules}

\vspace{3pt}
\setlength{\tabcolsep}{3pt}
\renewcommand{\arraystretch}{1.15}
\begin{tabular*}{\linewidth}{@{\extracolsep{\fill}}lccc@{}}
\toprule
View family / fusion & Black Sea CHL & Baltic Sea NANO & Global Ocean SSS \\
 & MSE$\downarrow$\,($\times10^{-3}$) & MSE$\downarrow$\,($\times10^{-3}$) & MSE$\downarrow$\,($\times10^{-3}$) \\
\midrule
\multicolumn{4}{l}{\textit{A: identical overlap views; fusion weights vary}} \\
Confidence-weighted mean & $15.525$ & $31.106$ & $32.185$ \\
Gaussian-weighted mean & $15.569$ & $31.177$ & $32.203$ \\
Uncertainty-weighted fusion & $15.481$ & $30.922$ & $32.102$ \\
Uniform mean (ours) & $\mathbf{15.227}$ & $\mathbf{30.576}$ & $\mathbf{32.053}$ \\
\midrule
\multicolumn{4}{l}{\textit{B: covering and support union vary}} \\
Uniform non-overlap & $15.660$ & $31.532$ & $32.174$ \\
Half-stride overlap + uniform mean & $\mathbf{15.227}$ & $\mathbf{30.576}$ & $\mathbf{32.053}$ \\
\bottomrule
\end{tabular*}
\end{table}
\FloatBarrier

Uniform fusion gives the lowest local-prior MSE among the tested weighting rules on each product (Table~\ref{tab:main-fusion}, Panel A). Its margin over the best alternative is small for Global Ocean SSS ($32.053$ versus $32.102$ in the table's units), so the evidence supports the chosen simple rule within the tested family rather than a broad claim that adaptive weighting is harmful. Half-stride overlap also gives lower prior MSE than non-overlap on each product (Table~\ref{tab:main-fusion}, Panel B). This difference combines multiple changes: overlap supplies several predictions for some coordinates and alters the union of observations that can affect their estimates. It cannot be attributed to averaging alone.

\FloatBarrier
\paragraph{Patch-scale sensitivity.}
\label{app:patch-scale-64}
The patch size controls how much context each local prediction can see, while the stride and number of views affect coverage and computational work. We compare $32\times32$/stride-16 and $16\times16$/stride-8 coverings, corresponding to nine and 49 in-bounds views per $64\times64$ field.
Both use uniform overlap consensus and are evaluated through full-mask residual reconstruction.
Validation selects $32\times32$/stride-16 for Black Sea CHL and Baltic Sea NANO, and $16\times16$/stride-8 for Global Ocean SSS.

\begin{table}[H]
\centering
\footnotesize
\caption{Patch-scale comparison for final full-mask residual reconstruction on $64\times64$ ocean fields. Bold marks the better result in each column.}
\label{tab:patch_scale_64}
\label{tab:patch-scale}

\vspace{3pt}
\setlength{\tabcolsep}{0.8pt}
\renewcommand{\arraystretch}{1.15}
\begin{tabular*}{\linewidth}{@{\extracolsep{\fill}}lccccccc@{}}
\toprule
Patch / stride & \shortstack[b]{Views\\per field} & \multicolumn{2}{c}{Black Sea CHL} & \multicolumn{2}{c}{Baltic Sea NANO} & \multicolumn{2}{c}{Global Ocean SSS} \\
\cmidrule(lr){3-4}\cmidrule(lr){5-6}\cmidrule(lr){7-8}
 &  & MSE$\downarrow$\,($\times10^{-3}$) & PSNR$\uparrow$\,(dB) & MSE$\downarrow$\,($\times10^{-3}$) & PSNR$\uparrow$\,(dB) & MSE$\downarrow$\,($\times10^{-3}$) & PSNR$\uparrow$\,(dB) \\
\midrule
$32\times32$ / 16 & 9 & $\mathbf{7.164}$ & $\mathbf{21.79}$ & $\mathbf{19.345}$ & $\mathbf{17.37}$ & $28.297$ & $15.76$ \\
$16\times16$ / 8 & 49 & $7.542$ & $21.45$ & $21.405$ & $16.84$ & $\mathbf{26.798}$ & $\mathbf{15.93}$ \\
\bottomrule
\end{tabular*}
\end{table}

The smaller covering increases final MSE on Black Sea CHL ($7.164$ to $7.542$) and Baltic Sea NANO ($19.345$ to $21.405$), but decreases it on Global Ocean SSS ($28.297$ to $26.798$; Table~\ref{tab:patch-scale}). PSNR follows the same ordering. Thus the best tested scale depends on the product; the comparison changes patch support, covering geometry, and view count together, so it does not isolate any one of those factors.

\FloatBarrier

\subsection{Remote-Evidence Interventions on Full-Mask Reconstructions}
\label{app:remote-evidence}

The local-support argument in Appendix~\ref{app:local-support} predicts that changing context outside $U(B)$ cannot change the prior on block $B$. The experiment asks whether the \emph{corrected} prediction nevertheless changes, and whether removing such context raises or lowers error. We first use a complete remote-band deletion to test the structural prediction and measure output displacement. We then use equal-sized deletions across distance bands to compare error effects. Both diagnostics intervene on the final reconstruction produced by the default ocean full-mask protocol.

\paragraph{Paired inputs and diagnostic support.}
For each evaluation item and training-seed checkpoint, we compare the reconstruction from the full supplied mask $\bm{M}_{\mathrm{in}}$ with reconstructions obtained after prescribed observation deletions.
The unmodified and intervened inputs share the original near-zero perturbation.
Each input is evaluated through local-prior construction, feature computation, and full-domain residual correction, using all observations remaining after its specified deletion without an additional context split.
Scoring remains restricted to the original cached outer holdout constructed in Appendix~\ref{app:observation-scoring-protocols}.
Table~\ref{tab:remote-settings} summarizes the diagnostic windows, local coverings, eligibility criterion, and pairing protocol.

For each diagnostic window $W$, define
\begin{equation}
    B(W)=W\cap \bm{M}_{\mathrm{eval}}.
    \label{eq:remote-scored-block}
\end{equation}
When $B$ is understood, $W(B)$ denotes its associated diagnostic window. The
contributing-patch support union $U(B)$ for the scored set $B(W)$ is defined in Appendix~\ref{app:local-support}.
For one evaluation item, we suppress its sample index and let $s$ index the training-seed checkpoint.
The supplied observations outside the local-prior support form
\begin{equation}
R(B)=\{j:\bm{M}_{\mathrm{in},j}=1\}\setminus U(B).
\label{eq:remote-evidence-far-set}
\end{equation}
For $j\in R(B)$, its distance to the scored block is
\begin{equation}
\rho(j,B)=\min_{i\in B}\lVert j-i\rVert_2.
\label{eq:remote-distance-to-block}
\end{equation}
Sorting $R(B)$ by $\rho(j,B)$ and applying the mid-rank equal-frequency rule gives $R_N(B)$, $R_M(B)$, and $R_F(B)$.
Remote Near remains outside $U(B)$; ``near'' is relative only to the other remote observations.

\suppressfloats[t]
\begin{table}[!htbp]
\centering
\footnotesize
\caption{Diagnostic settings for remote-evidence interventions on full-mask ocean reconstructions.}
\label{tab:app-remote-settings}
\label{tab:remote-settings}

\vspace{3pt}
\setlength{\tabcolsep}{3pt}
\renewcommand{\arraystretch}{1.15}
\begin{tabularx}{\linewidth}{@{}l>{\raggedright\arraybackslash}X@{}}
\toprule
Item & Setting \\
\midrule
Resolution & $64\times64$ \\
Black Sea / Baltic bank & $W_{32}$: $32\times32$, stride 32; four windows per field \\
Global SSS bank & $W_{16}$: $16\times16$, stride 16; 16 windows per field \\
Eligibility & At least eight scored coordinates, $|B(W)|\geq8$ \\
Black Sea / Baltic local views & $32\times32$ patches, stride 16; nine views \\
Global SSS local views & $16\times16$ patches, stride 8; 49 views \\
Distance bands & Mid-rank equal-frequency remote terciles \\
Evaluation unit & Paired reconstructions for one evaluation item, scored block, and checkpoint \\
Pairing & Shared checkpoint, original supplied input, and near-zero perturbation; arms differ only by the prescribed deletion \\
\bottomrule
\end{tabularx}
\end{table}

\paragraph{Support invariance and prediction displacement.}
The structural diagnostic removes the complete Remote Far tercile $R_F(B)$ from the full supplied mask.
Here ``full'' denotes the unmodified input $\bm{M}_{\mathrm{in}}$.
For deletion set $A_a(B)\subseteq\{j:\bm{M}_{\mathrm{in},j}=1\}$, construct
\[
\begin{aligned}
\bm{M}_{\mathrm{ctx}}^a&=\bm{M}_{\mathrm{in}}\odot\bigl(1-\mathbf1_{A_a(B)}\bigr),\\
\bm{x}_{\ctx,\delta,s}^a&=\bm{M}_{\mathrm{ctx}}^a\odot\bm{x}_{\mathrm{in},\delta,s},\\
O_{\delta,s}^a&=(\bm{x}_{\ctx,\delta,s}^a,\bm{M}_{\mathrm{ctx}}^a),\\
\bm{x}_{\mathrm{in},\delta,s}&=\alpha_\delta\bm{x}_{\mathrm{in}}+\sigma_\delta\bm{\epsilon}_s.
\end{aligned}
\]
The unmodified state uses $A_{\mathrm{full}}(B)=\varnothing$, and the displacement diagnostic uses $A_{\mathrm{far}}(B)=R_F(B)$.
Each state is evaluated through prior construction, feature computation, and full-domain residual correction with $B$ fixed.
Let $\widehat{\bm{x}}_{\mathrm{out},s}^a$ denote its final reconstruction.
Because $B\subseteq\bm{M}_{\mathrm{eval}}$ is disjoint from $\bm{M}_{\mathrm{in}}$, restoring supplied observations does not change the prediction on $B$.
We measure the prior and final-output displacements
\begin{equation}
d_{\bm{\Pi},s}^{\mathrm{far}}(B)
=\frac{\left\|\bm{\Pi}_{\delta,B,s}^{\mathrm{far}}-\bm{\Pi}_{\delta,B,s}^{\mathrm{full}}\right\|_2^2}{d_x|B|},
\label{eq:remote-prior-displacement}
\end{equation}
and
\begin{equation}
d_{\widehat{\bm{x}},s}^{\mathrm{far}}(B)
=\frac{\left\|\widehat{\bm{x}}_{\mathrm{out},B,s}^{\mathrm{far}}-\widehat{\bm{x}}_{\mathrm{out},B,s}^{\mathrm{full}}\right\|_2^2}{d_x|B|}.
\label{eq:remote-output-displacement}
\end{equation}
\begin{table}[H]
\centering
\small
\caption{Local-prior and final-reconstruction displacement after deleting the complete Remote Far tercile from the full supplied mask.}
\label{tab:app-remote-structural}
\label{tab:remote-displacement}

\vspace{3pt}
\setlength{\tabcolsep}{3pt}
\renewcommand{\arraystretch}{1.15}
\begin{tabular*}{\linewidth}{@{\extracolsep{\fill}}lcccc@{}}
\toprule
Dataset & $\max d_{\bm{\Pi}}^{\mathrm{far}}$ & \shortstack{Mean $d_{\widehat{\bm{x}}}^{\mathrm{far}}$\\{[95\% CI]}} & \shortstack{Median\\$d_{\widehat{\bm{x}}}^{\mathrm{far}}$} & \shortstack{Max\\$d_{\widehat{\bm{x}}}^{\mathrm{far}}$} \\
 &  & $(\times10^{-5})$ & $(\times10^{-6})$ & $(\times10^{-2})$ \\
\midrule
Baltic Sea NANO & $0$ & \meanCI{6.46}{4.71,\,8.61} & $3.35$ & $1.617$ \\
Black Sea CHL & $0$ & \meanCI{5.80}{4.71,\,7.09} & $2.67$ & $1.156$ \\
Global Ocean SSS & $0$ & \meanCI{62.30}{56.96,\,68.02} & $28.00$ & $6.348$ \\
\bottomrule
\end{tabular*}
\end{table}

The local-prior displacement is zero for every eligible paired block evaluation (Table~\ref{tab:remote-displacement}).
Mean, median, and maximum final-output displacements are positive on all three datasets, with Global Ocean SSS having the largest reported mean.
The full-domain estimator therefore responds to observations outside the support that determines the local prior on the scored block.

\FloatBarrier
\paragraph{Matched-dose error changes.}
The local-near candidate set contains supplied observations inside the local-prior support union $U(B)$ but outside the current diagnostic window:
\begin{equation}
L_N(B)
=\left(\{j:\bm{M}_{\mathrm{in},j}=1\}\cap U(B)\right)\setminus W(B).
\label{eq:local-near-set}
\end{equation}
The matched deletion dose for each block is
\begin{equation}
m_B
=\min\left\{|R_N(B)|,\,|R_M(B)|,\,|R_F(B)|,\,|L_N(B)|\right\}.
\label{eq:matched-remote-dose}
\end{equation}
For each scored block, the deterministic compact selection rule removes $m_B$ observations from each candidate set, producing five paired input states for each checkpoint:
\begin{equation}
O_{\delta,s}^{\mathrm{full}},\quad
O_{\delta,s}^{\mathrm{RN}},\quad
O_{\delta,s}^{\mathrm{RM}},\quad
O_{\delta,s}^{\mathrm{RF}},\quad
O_{\delta,s}^{\mathrm{LN}}.
\label{eq:matched-remote-states}
\end{equation}
The matched-dose experiment uses equally sized deletions from the full supplied mask for each scored block, whereas the displacement diagnostic removes the complete Remote Far tercile.

For removal arm $a\in\{\mathrm{RN},\mathrm{RM},\mathrm{RF},\mathrm{LN}\}$, define the paired block-MSE change of the final reconstruction:
\begin{equation}
\Delta_{a,s}(B)
=\operatorname{MSE}_B\!\left(\widehat{\bm{x}}_{\mathrm{out},s}^a\right)
-\operatorname{MSE}_B\!\left(\widehat{\bm{x}}_{\mathrm{out},s}^{\mathrm{full}}\right).
\label{eq:remote-delta-mse}
\end{equation}
Measurements are computed for each eligible scored block in each evaluation item and for each training-seed checkpoint, then summarized across these paired evaluations.

\begin{table}[H]
\centering
\small
\caption{Final-reconstruction block-MSE changes under matched-dose deletion from the full supplied mask. Entries are means with 95\% confidence intervals; positive values indicate increased error after deletion. Bold marks means whose intervals exclude zero.}
\label{tab:app-remote-delta-mse}
\label{tab:remote-error}

\vspace{3pt}
\setlength{\tabcolsep}{3pt}
\renewcommand{\arraystretch}{1.15}
\begin{tabular*}{\linewidth}{@{\extracolsep{\fill}}lcccc@{}}
\toprule
 & \multicolumn{4}{c}{$\Delta$ block MSE $(\times10^{-5})$} \\
\cmidrule(lr){2-5}
Dataset & Remote Near & Remote Mid & Remote Far & Local Near \\
\midrule
Baltic Sea NANO & \meanCI{0.95}{-3.32,\,5.23} & \meanCI{\mathbf{4.66}}{1.08,\,8.56} & \meanCI{\mathbf{5.25}}{1.14,\,9.56} & \meanCI{\mathbf{50.11}}{32.63,\,69.56} \\
Black Sea CHL & \meanCI{0.53}{-1.57,\,2.77} & \meanCI{\mathbf{2.53}}{0.43,\,4.84} & \meanCI{\mathbf{3.72}}{0.65,\,6.75} & \meanCI{5.87}{-1.73,\,13.10} \\
Global Ocean SSS & \meanCI{\mathbf{6.76}}{2.99,\,10.54} & \meanCI{0.98}{-1.53,\,3.49} & \meanCI{\mathbf{-12.85}}{-18.64,\,-7.09} & \meanCI{\mathbf{64.12}}{55.49,\,73.16} \\
\bottomrule
\end{tabular*}
\end{table}

For Baltic Sea NANO, the intervals for Remote Mid, Remote Far, and Local Near error changes lie above zero, while the Remote Near interval crosses zero (Table~\ref{tab:remote-error}).
For Black Sea CHL, the Remote Mid and Remote Far intervals lie above zero, while Remote Near and Local Near cross zero.
For Global Ocean SSS, Remote Near and Local Near deletion increase mean block MSE, Remote Far deletion decreases it, and the Remote Mid interval crosses zero.
These full-mask interventions distinguish responsiveness beyond local support from the dataset-dependent accuracy effects of the removed observations. The remote bands contain different spatial locations even when their deletion counts match, and the mixed signs on Global Ocean SSS rule out a universal claim that farther context always improves this estimator's prediction.

\FloatBarrier

\subsection{Inference Cost and Observation Retention}
\label{app:inference-cost}

\paragraph{Comparison protocol.} We compare our estimator with parameter-matched Full Direct on Black Sea CHL and Shallow Water Pixel 30\% at $64\times64$. Full Direct is based on OAMP and IDCD, respectively, as defined in Appendix~\ref{app:retraining-reporting}. For each estimator, all inference configurations reuse the same set of independently trained checkpoints. For each sample, the supplied support $\bm{M}_{\mathrm{in}}$ and outer scoring support remain fixed across inference configurations. Context-ensemble comparisons additionally reuse one cached subcontext sequence, with an ensemble of size $K$ using its first $K$ entries. Full-mask and context-ensemble inference reuse the same near-zero perturbation and are scored on the same outer evaluation support. Full-mask uses all supplied observations, whereas Context $K=1$ uses one subsampled context. The default modes and ensemble size are selected on validation data as specified in Appendix~\ref{app:full-support-inference}.

Reconstruction latency is measured in milliseconds per request for one $64\times64$ field on a single NVIDIA A40 GPU in FP32. The measurement includes all local patch predictions, overlap fusion, feature construction, and full-domain correction. Inference subcontext masks, including BFN-generated masks, are cached, so mask generation is excluded from the timing interval. Ensemble contexts are batched for parallel prediction on the same GPU.

\suppressfloats[t]
\begin{table}[!t]
\centering
\footnotesize
\caption{Accuracy and request latency under full-mask and context-ensemble inference at $64\times64$. Entries are means $\pm$ standard deviations across independent training seeds. For latency, each seed contributes its median or 95th-percentile request time. Bold marks the best mean accuracy within each dataset.}
\label{tab:inference-cost}
\vspace{3pt}
\setlength{\tabcolsep}{2pt}
\renewcommand{\arraystretch}{1.15}
\begin{tabular*}{\linewidth}{@{\extracolsep{\fill}}llcccc@{}}
\toprule
Model & Inference & MSE$\downarrow$\,($\times10^{-3}$) & PSNR$\uparrow$\,(dB) & \shortstack[b]{Median latency\\(ms/field)} & \shortstack[b]{P95 latency\\(ms/field)} \\
\midrule
\multicolumn{6}{l}{\textit{A: Black Sea CHL}} \\
Full Direct & Full-mask & \meanstd{9.350}{0.306} & \meanstd{20.32}{0.12} & \meanstd{69.46}{0.20} & \meanstd{69.78}{0.31} \\
Full Direct & Context $K=1$ & \meanstd{13.922}{0.055} & \meanstd{18.61}{0.07} & \meanstd{69.41}{0.10} & \meanstd{69.76}{0.12} \\
Full Direct & Context $K=2$ & \meanstd{13.823}{0.065} & \meanstd{18.64}{0.07} & \meanstd{80.05}{0.19} & \meanstd{80.29}{0.26} \\
Full Direct & Context $K=4$ & \meanstd{13.762}{0.072} & \meanstd{18.66}{0.08} & \meanstd{105.30}{0.20} & \meanstd{105.51}{0.15} \\
Full Direct & Context $K=8$ & \meanstd{13.721}{0.083} & \meanstd{18.66}{0.08} & \meanstd{210.30}{0.41} & \meanstd{210.60}{0.50} \\
\addlinespace[2pt]
Ours & Full-mask & \meanstd{\mathbf{7.164}}{0.060} & \meanstd{\mathbf{21.79}}{0.04} & \meanstd{191.82}{1.64} & \meanstd{192.85}{2.84} \\
Ours & Context $K=1$ & \meanstd{12.016}{0.110} & \meanstd{19.37}{0.03} & \meanstd{192.06}{1.11} & \meanstd{196.95}{5.18} \\
Ours & Context $K=4$ & \meanstd{11.510}{0.015} & \meanstd{19.57}{0.01} & \meanstd{345.38}{0.01} & \meanstd{345.96}{0.07} \\
\midrule
\multicolumn{6}{l}{\textit{B: Shallow Water Pixel 30\%}} \\
Full Direct & Full-mask & \meanstd{0.79503}{0.16693} & \meanstd{37.77}{0.72} & \meanstd{69.10}{0.03} & \meanstd{69.35}{0.01} \\
Full Direct & Context $K=1$ & \meanstd{0.70450}{0.02675} & \meanstd{38.36}{0.15} & \meanstd{69.20}{0.14} & \meanstd{69.52}{0.08} \\
Full Direct & Context $K=2$ & \meanstd{0.68605}{0.02891} & \meanstd{38.46}{0.16} & \meanstd{88.06}{0.12} & \meanstd{88.36}{0.06} \\
Full Direct & Context $K=4$ & \meanstd{0.67749}{0.02930} & \meanstd{38.51}{0.16} & \meanstd{110.50}{0.06} & \meanstd{110.92}{0.10} \\
Full Direct & Context $K=8$ & \meanstd{0.67295}{0.03067} & \meanstd{38.53}{0.16} & \meanstd{210.54}{0.05} & \meanstd{210.83}{0.04} \\
Full Direct & Context $K=16$ & \meanstd{0.67009}{0.03109} & \meanstd{38.55}{0.16} & \meanstd{399.74}{0.29} & \meanstd{400.68}{0.15} \\
Full Direct & Context $K=32$ & \meanstd{0.66856}{0.03128} & \meanstd{38.55}{0.16} & \meanstd{434.50}{0.19} & \meanstd{434.86}{0.17} \\
\addlinespace[2pt]
Ours & Full-mask & \meanstd{0.05246}{0.01542} & \meanstd{50.19}{1.16} & \meanstd{190.39}{0.08} & \meanstd{190.99}{0.12} \\
Ours & Context $K=1$ & \meanstd{0.04998}{0.01149} & \meanstd{50.41}{0.98} & \meanstd{190.86}{0.07} & \meanstd{191.31}{0.08} \\
Ours & Context $K=4$ & \meanstd{0.03771}{0.00819} & \meanstd{51.40}{0.98} & \meanstd{349.09}{0.66} & \meanstd{349.53}{0.96} \\
Ours & Context $K=8$ & \meanstd{\mathbf{0.03521}}{0.00738} & \meanstd{\mathbf{51.63}}{0.96} & \meanstd{668.28}{1.28} & \meanstd{669.76}{1.19} \\
\bottomrule
\end{tabular*}
\end{table}
\paragraph{Accuracy under common latency ceilings.} The ceilings below apply to the across-seed mean of per-seed median reconstruction latency. For each ceiling, we compare the lowest mean MSE among the measured configurations in Table~\ref{tab:inference-cost} that satisfy it. On Black Sea, full-mask inference gives the lowest MSE for both estimators. Under a 211 ms ceiling, our full-mask estimator achieves $7.164\times10^{-3}$ MSE at 191.82 ms, compared with $9.350\times10^{-3}$ at 69.46 ms for the best eligible Full Direct configuration, a 23.4\% MSE reduction. Full Direct's full-mask configuration is faster, but increasing its context ensemble to $K=8$ raises latency to 210.30 ms without improving on its full-mask accuracy.

On Shallow Water Pixel 30\%, our estimator retains its advantage at lower ensemble sizes. Under a 211 ms ceiling, our $K=1$ configuration achieves $0.04998\times10^{-3}$ MSE at 190.86 ms, compared with $0.67295\times10^{-3}$ at 210.54 ms for Full Direct with $K=8$, a 92.6\% MSE reduction. Under a 400 ms ceiling, our $K=4$ configuration achieves $0.03771\times10^{-3}$ at 349.09 ms, compared with $0.67009\times10^{-3}$ at 399.74 ms for Full Direct with $K=16$, a 94.4\% reduction. Increasing our ensemble from $K=4$ to $K=8$ provides a further 6.6\% MSE reduction while increasing median latency from 349.09 to 668.28 ms.

\paragraph{Training and inference observation supports.} The two settings differ in both acquisition masks and internal partitions (Appendices~\ref{app:controlled-partitions}, \ref{app:bfn-construction}, and~\ref{app:observation-scoring-protocols}). Under Shallow Water Pixel 30\%, the stored acquisition mask retains approximately 30\% of the valid field, and training retains approximately 70\% of those observations as context. The nominal training context and query therefore occupy 21\% and 9\% of the valid field, respectively. At test time, full-mask inference supplies approximately 30\%, whereas each ensemble subcontext supplies approximately 21\%. The subcontexts are thus closer to the training inputs in observation density and pixel-mask structure.

Black Sea training instead partitions each sample's source-supported observations directly. Its retained context density varies across samples and guided proposals rather than being fixed by the BFN anchor-retention parameter (Appendix~\ref{app:bfn-construction}). At validation and test time, the empirical real-mask overlay protocol in Appendix~\ref{app:observation-scoring-protocols} constructs the cached supplied support $\bm{M}_{\mathrm{in}}$ and outer scoring support. Full-mask inference retains $\bm{M}_{\mathrm{in}}$, whereas context ensembling applies an additional guided-mask intersection within it.

\paragraph{Observation retention and spatial coverage.} We compute mask statistics over all test samples using their first eight cached subcontexts. Observation density uses the valid domain as its denominator, whereas subcontext retention and context-union coverage use the supplied support $\bm{M}_{\mathrm{in}}$. Ratios are computed within each sample before equal-weight averaging across samples; subcontext statistics additionally average over the eight contexts within each sample.

For sample $n$, let $Q_n$ denote its fixed outer scoring coordinates. The mean distance to an observation mask $A$ is
\[
d_n(A)=\frac{1}{|Q_n|}\sum_{i\in Q_n}\min_{j:A_j=1}\|i-j\|_2.
\]
We compute this distance within each sample and then average equally across samples, averaging over the eight contexts for the subcontext statistic. Distances are Euclidean distances on the $64\times64$ grid, measured in pixels rather than physical units.

\begin{table}[H]
\centering
\small
\caption{Test-time observation retention and spatial coverage on Black Sea CHL and Shallow Water Pixel 30\%. Entries are equally weighted means over test samples.}
\label{tab:observation-retention}
\vspace{3pt}
\setlength{\tabcolsep}{4pt}
\renewcommand{\arraystretch}{1.15}
\begin{tabularx}{\linewidth}{@{}>{\raggedright\arraybackslash}Xcc@{}}
\toprule
Mask statistic & Black Sea CHL & SW Pixel 30\% \\
\midrule
Full-mask observations / valid domain & 50.72\% & 29.96\% \\
Subcontext observations / valid domain & 27.70\% & 20.97\% \\
Supplied observations retained per subcontext & 51.23\% & 70.00\% \\
Supplied observations covered by the first 4 contexts & 72.93\% & 99.19\% \\
Supplied observations covered by the first 8 contexts & 78.05\% & 99.994\% \\
Mean nearest-observation distance: full-mask & 3.09 pixels & 1.15 pixels \\
Mean nearest-observation distance: subcontext & 7.54 pixels & 1.29 pixels \\
\bottomrule
\end{tabularx}
\end{table}

Black Sea subcontexts retain substantially less of the supplied evidence and create larger gaps around the scoring locations (Table~\ref{tab:observation-retention}). Even the first eight-context union covers only 78.05\% of the original input observations, while the mean nearest-observation distance increases from 3.09 to 7.54 pixels. In Shallow Water Pixel 30\%, the corresponding union covers 99.994\%, and the distance increases only from 1.15 to 1.29 pixels. Union coverage does not imply that any individual member conditions jointly on all supplied observations. Restoring observations at the output enforces their values but does not recover their omitted conditioning influence on predictions at unsupplied locations.

\paragraph{Interpreting the different ensemble effects.} Field structure may amplify these geometric differences. Black Sea chlorophyll observations have irregular spatial support and localized variation, for which removed nearby measurements may be difficult to replace with distant evidence. The Shallow Water subset shares radial dam-break structure, with variation in initial dam radius and evolution time. Spatially dispersed observations can repeatedly constrain this shared structure, potentially making moderate pixel subsampling less costly. Together with the closer match to training-context density, this provides an explanation consistent with the Shallow Water results.

Averaging helps relative to a single subcontext in both settings: increasing our ensemble from $K=1$ to $K=4$ reduces mean MSE by 4.2\% on Black Sea and 24.6\% on Shallow Water Pixel 30\% (Table~\ref{tab:inference-cost}). On Black Sea, this gain does not compensate for observation removal, and full-mask inference still has 37.8\% lower MSE than the $K=4$ ensemble. On Shallow Water, the $K=4$ ensemble instead improves over full-mask inference by 28.1\%. These reductions are consistent with averaging context-sensitive errors, while errors shared across contexts can persist when relevant observations are repeatedly omitted. Because all members share the same near-zero perturbation, the gains are not attributable to averaging independent diffusion-noise draws.

The measured retention and distance statistics establish the different observation costs of subsampling; training-input similarity and error averaging provide complementary interpretations of the resulting accuracy trends.

\FloatBarrier

\FloatBarrier
\section{Limitations}
\label{app:limitations}

Some block-missing reconstructions exhibit mild seam-like artifacts near observed--missing interfaces.
These local transitions can remain imperfect even when fine-scale structures within the missing region are recovered.
Improving cross-boundary consistency without suppressing local detail is a useful direction for future work.

\section{Reconstruction Visualizations}
\label{app:reconstruction-visualizations}
\label{app:visualizations}

The ocean visualizations use full-mask inference. In Figures~\ref{fig:ocean-decomposition}--\ref{fig:ocean128-sss}, black and red outlines mark supplied observations and withheld source-supported scoring locations, respectively; gray denotes outside-domain locations, and white in Reference denotes source-missing values.

\subsection{Ocean Reconstruction at \texorpdfstring{$64\times64$}{64 x 64}}
\label{app:ocean64-visualizations}

\begin{figure}[H]
\centering
\setlength{\abovecaptionskip}{6pt}
\input{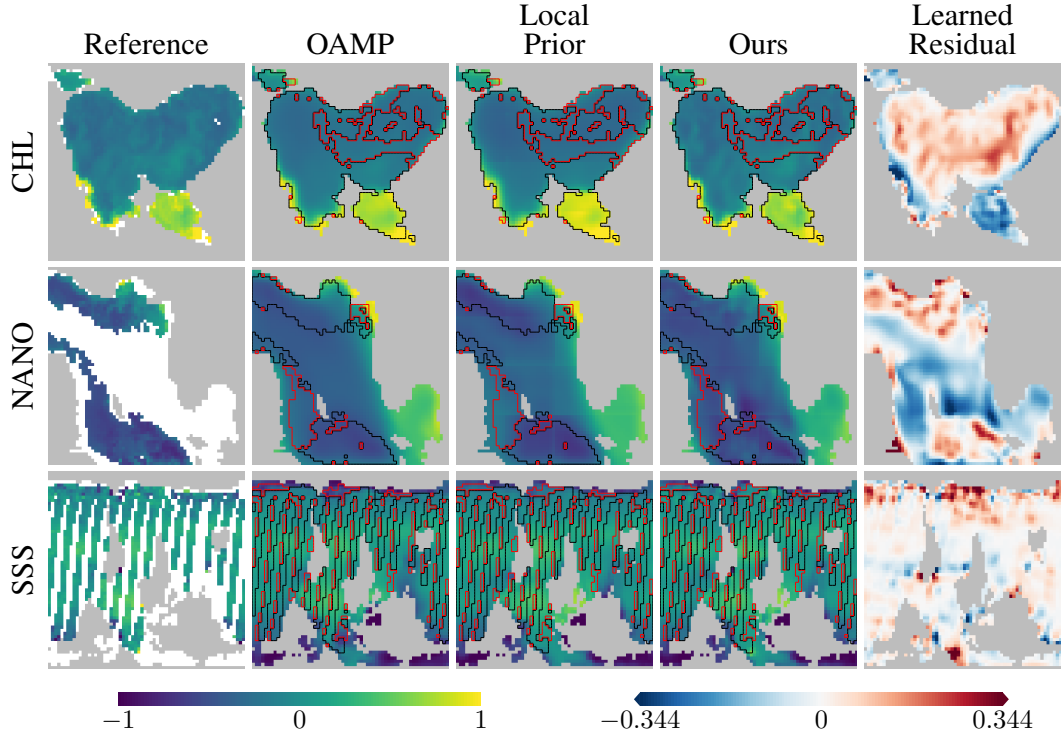}
\caption{Local-to-global decomposition for full-mask ocean reconstruction at $64\times64$. Model predictions are shown before restoring supplied observations, preserving the pointwise relation $\text{Ours}=\text{Local Prior}+\text{Learned Residual}$.}
\label{fig:ocean-decomposition}
\end{figure}

Figure~\ref{fig:ocean-decomposition} decomposes the model predictions used in Figure~\ref{fig:local-global-reconstruction}, for the same samples and masks, and also includes the external OAMP baseline from Table~\ref{tab:ocean64}. The local priors already contain large-scale field structure, while the learned residuals vary across the domain rather than applying a constant offset.

Figures~\ref{fig:ocean64-chl}--\ref{fig:ocean64-sss} compare Ours with parameter-matched Full Direct at $64\times64$, with supplied observations restored for display.
  \input{visualizations/ocean_selected_task2_chl.tex}%
  \input{visualizations/ocean_selected_task2_nano.tex}%
  \input{visualizations/ocean_selected_task2_sss.tex}%

\FloatBarrier
\subsection{Ocean Reconstruction at \texorpdfstring{$128\times128$}{128 x 128}}
\label{app:ocean128-visualizations}
Figures~\ref{fig:ocean128-chl}--\ref{fig:ocean128-sss} compare Reference and Ours at $128\times128$, with supplied observations restored for display.
  \input{visualizations/ocean_selected_task3_chl.tex}%
  \input{visualizations/ocean_selected_task3_nano.tex}%
  \input{visualizations/ocean_selected_task3_sss.tex}%

\FloatBarrier

\subsection{PDE Reconstruction under Pixel Observations}
\label{app:pde-pixel-visualizations}
\begin{figure}[H]
\centering
\setlength{\abovecaptionskip}{6pt}
\begingroup
\setlength{\parskip}{0pt}
{\VisualizationLabelFont Navier--Stokes Pixel 30\%}\par\smallskip
\noindent{\VisualizationLabelFont Reference}\par\nobreak
\noindent\hbox to\linewidth{%
\includegraphics[width=16mm]{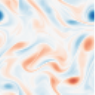}%
\hfill%
\includegraphics[width=16mm]{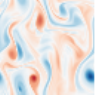}%
\hfill%
\includegraphics[width=16mm]{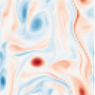}%
\hfill%
\includegraphics[width=16mm]{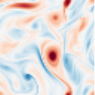}%
\hfill%
\includegraphics[width=16mm]{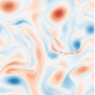}%
\hfill%
\includegraphics[width=16mm]{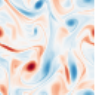}%
\hfill%
\includegraphics[width=16mm]{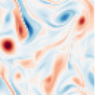}%
\hfill%
\includegraphics[width=16mm]{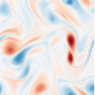}%
}\par
\vspace{1mm}
\noindent{\VisualizationLabelFont Ours}\par\nobreak
\noindent\hbox to\linewidth{%
\includegraphics[width=16mm]{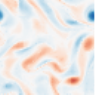}%
\hfill%
\includegraphics[width=16mm]{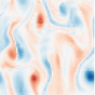}%
\hfill%
\includegraphics[width=16mm]{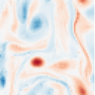}%
\hfill%
\includegraphics[width=16mm]{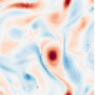}%
\hfill%
\includegraphics[width=16mm]{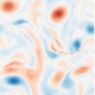}%
\hfill%
\includegraphics[width=16mm]{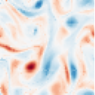}%
\hfill%
\includegraphics[width=16mm]{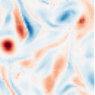}%
\hfill%
\includegraphics[width=16mm]{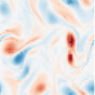}%
}\par
\vspace{1mm}
\noindent{\VisualizationLabelFont AmbientDiff}\par\nobreak
\noindent\hbox to\linewidth{%
\includegraphics[width=16mm]{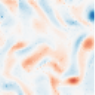}%
\hfill%
\includegraphics[width=16mm]{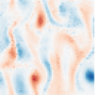}%
\hfill%
\includegraphics[width=16mm]{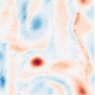}%
\hfill%
\includegraphics[width=16mm]{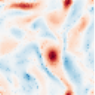}%
\hfill%
\includegraphics[width=16mm]{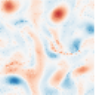}%
\hfill%
\includegraphics[width=16mm]{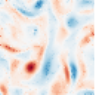}%
\hfill%
\includegraphics[width=16mm]{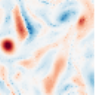}%
\hfill%
\includegraphics[width=16mm]{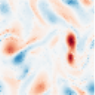}%
}\par
\endgroup
\caption{Navier--Stokes Pixel 30\% reconstructions at $64\times64$, comparing Ours with AmbientDiff.}
\label{fig:pde-task1-ns-pixel30}
\label{fig:pde-examples}
\end{figure}

\begin{figure}[H]
\centering
\setlength{\abovecaptionskip}{6pt}
\begingroup
\setlength{\parskip}{0pt}
{\VisualizationLabelFont Shallow Water Pixel 30\%}\par\smallskip
\noindent{\VisualizationLabelFont Reference}\par\nobreak
\noindent\hbox to\linewidth{%
\includegraphics[width=16mm]{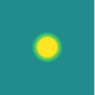}%
\hfill%
\includegraphics[width=16mm]{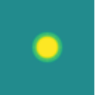}%
\hfill%
\includegraphics[width=16mm]{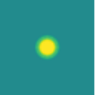}%
\hfill%
\includegraphics[width=16mm]{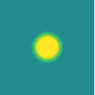}%
\hfill%
\includegraphics[width=16mm]{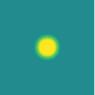}%
\hfill%
\includegraphics[width=16mm]{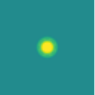}%
\hfill%
\includegraphics[width=16mm]{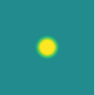}%
\hfill%
\includegraphics[width=16mm]{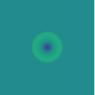}%
}\par
\vspace{1mm}
\noindent{\VisualizationLabelFont Ours}\par\nobreak
\noindent\hbox to\linewidth{%
\includegraphics[width=16mm]{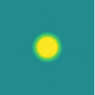}%
\hfill%
\includegraphics[width=16mm]{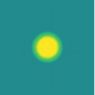}%
\hfill%
\includegraphics[width=16mm]{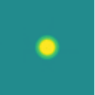}%
\hfill%
\includegraphics[width=16mm]{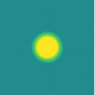}%
\hfill%
\includegraphics[width=16mm]{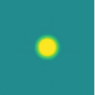}%
\hfill%
\includegraphics[width=16mm]{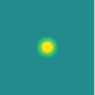}%
\hfill%
\includegraphics[width=16mm]{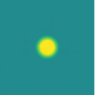}%
\hfill%
\includegraphics[width=16mm]{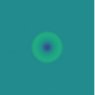}%
}\par
\vspace{1mm}
\noindent{\VisualizationLabelFont DINDiff}\par\nobreak
\noindent\hbox to\linewidth{%
\includegraphics[width=16mm]{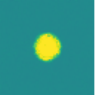}%
\hfill%
\includegraphics[width=16mm]{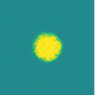}%
\hfill%
\includegraphics[width=16mm]{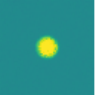}%
\hfill%
\includegraphics[width=16mm]{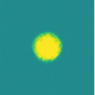}%
\hfill%
\includegraphics[width=16mm]{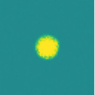}%
\hfill%
\includegraphics[width=16mm]{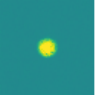}%
\hfill%
\includegraphics[width=16mm]{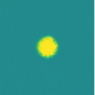}%
\hfill%
\includegraphics[width=16mm]{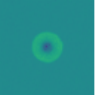}%
}\par
\endgroup
\caption{Shallow Water Pixel 30\% reconstructions at $64\times64$, comparing Ours with DINDiff.}
\label{fig:pde-task1-sw-pixel30}
\end{figure}

Figures~\ref{fig:pde-task1-ns-pixel30} and~\ref{fig:pde-task1-sw-pixel30} show Navier--Stokes and Shallow Water Pixel 30\% reconstructions, respectively. In Figures~\ref{fig:pde-task1-ns-pixel30}--\ref{fig:pde-task1-ns-block8}, Ours uses $K=8$ context ensembling. Supplied observations are restored for display, and each column compares the same sample across methods.

\FloatBarrier

\subsection{PDE Reconstruction under Block Observations}
\label{app:pde-block-visualizations}
\begin{figure}[!htbp]
\centering
\setlength{\abovecaptionskip}{6pt}
\begingroup
\setlength{\parskip}{0pt}
{\VisualizationLabelFont Shallow Water Block 3/9}\par\smallskip
\noindent{\VisualizationLabelFont Reference}\par\nobreak
\noindent\hbox to\linewidth{%
\includegraphics[trim=0.65bp 0.65bp 0.65bp 0.65bp,width=16mm]{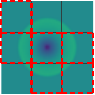}%
\hfill%
\includegraphics[trim=0.65bp 0.65bp 0.65bp 0.65bp,width=16mm]{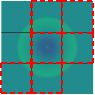}%
\hfill%
\includegraphics[trim=0.65bp 0.65bp 0.65bp 0.65bp,width=16mm]{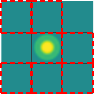}%
\hfill%
\includegraphics[trim=0.65bp 0.65bp 0.65bp 0.65bp,width=16mm]{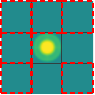}%
\hfill%
\includegraphics[trim=0.65bp 0.65bp 0.65bp 0.65bp,width=16mm]{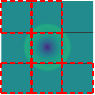}%
\hfill%
\includegraphics[trim=0.65bp 0.65bp 0.65bp 0.65bp,width=16mm]{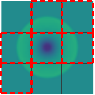}%
\hfill%
\includegraphics[trim=0.65bp 0.65bp 0.65bp 0.65bp,width=16mm]{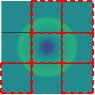}%
\hfill%
\includegraphics[trim=0.65bp 0.65bp 0.65bp 0.65bp,width=16mm]{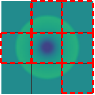}%
}\par
\vspace{1mm}
\noindent{\VisualizationLabelFont Ours}\par\nobreak
\noindent\hbox to\linewidth{%
\includegraphics[trim=0.65bp 0.65bp 0.65bp 0.65bp,width=16mm]{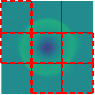}%
\hfill%
\includegraphics[trim=0.65bp 0.65bp 0.65bp 0.65bp,width=16mm]{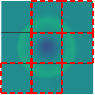}%
\hfill%
\includegraphics[trim=0.65bp 0.65bp 0.65bp 0.65bp,width=16mm]{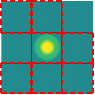}%
\hfill%
\includegraphics[trim=0.65bp 0.65bp 0.65bp 0.65bp,width=16mm]{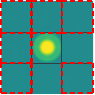}%
\hfill%
\includegraphics[trim=0.65bp 0.65bp 0.65bp 0.65bp,width=16mm]{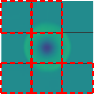}%
\hfill%
\includegraphics[trim=0.65bp 0.65bp 0.65bp 0.65bp,width=16mm]{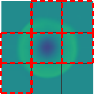}%
\hfill%
\includegraphics[trim=0.65bp 0.65bp 0.65bp 0.65bp,width=16mm]{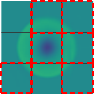}%
\hfill%
\includegraphics[trim=0.65bp 0.65bp 0.65bp 0.65bp,width=16mm]{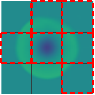}%
}\par
\vspace{1mm}
\noindent{\VisualizationLabelFont AmbientDiff}\par\nobreak
\noindent\hbox to\linewidth{%
\includegraphics[trim=0.65bp 0.65bp 0.65bp 0.65bp,width=16mm]{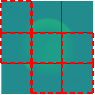}%
\hfill%
\includegraphics[trim=0.65bp 0.65bp 0.65bp 0.65bp,width=16mm]{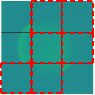}%
\hfill%
\includegraphics[trim=0.65bp 0.65bp 0.65bp 0.65bp,width=16mm]{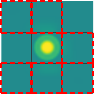}%
\hfill%
\includegraphics[trim=0.65bp 0.65bp 0.65bp 0.65bp,width=16mm]{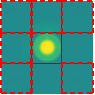}%
\hfill%
\includegraphics[trim=0.65bp 0.65bp 0.65bp 0.65bp,width=16mm]{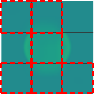}%
\hfill%
\includegraphics[trim=0.65bp 0.65bp 0.65bp 0.65bp,width=16mm]{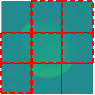}%
\hfill%
\includegraphics[trim=0.65bp 0.65bp 0.65bp 0.65bp,width=16mm]{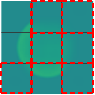}%
\hfill%
\includegraphics[trim=0.65bp 0.65bp 0.65bp 0.65bp,width=16mm]{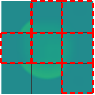}%
}\par
\endgroup
\caption{Shallow Water Block 3/9 reconstructions at $64\times64$, comparing Ours with AmbientDiff.}
\label{fig:pde-task1-sw-block3}
\end{figure}

\begin{figure}[!htbp]
\centering
\setlength{\abovecaptionskip}{6pt}
\begingroup
\setlength{\parskip}{0pt}
{\VisualizationLabelFont Shallow Water Block 8/9}\par\smallskip
\noindent{\VisualizationLabelFont Reference}\par\nobreak
\noindent\hbox to\linewidth{%
\includegraphics[trim=0.65bp 0.65bp 0.65bp 0.65bp,width=16mm]{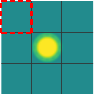}%
\hfill%
\includegraphics[trim=0.65bp 0.65bp 0.65bp 0.65bp,width=16mm]{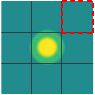}%
\hfill%
\includegraphics[trim=0.65bp 0.65bp 0.65bp 0.65bp,width=16mm]{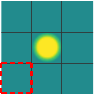}%
\hfill%
\includegraphics[trim=0.65bp 0.65bp 0.65bp 0.65bp,width=16mm]{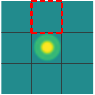}%
\hfill%
\includegraphics[trim=0.65bp 0.65bp 0.65bp 0.65bp,width=16mm]{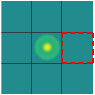}%
\hfill%
\includegraphics[trim=0.65bp 0.65bp 0.65bp 0.65bp,width=16mm]{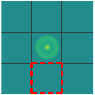}%
\hfill%
\includegraphics[trim=0.65bp 0.65bp 0.65bp 0.65bp,width=16mm]{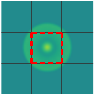}%
\hfill%
\includegraphics[trim=0.65bp 0.65bp 0.65bp 0.65bp,width=16mm]{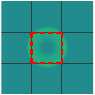}%
}\par
\vspace{1mm}
\noindent{\VisualizationLabelFont Ours}\par\nobreak
\noindent\hbox to\linewidth{%
\includegraphics[trim=0.65bp 0.65bp 0.65bp 0.65bp,width=16mm]{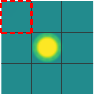}%
\hfill%
\includegraphics[trim=0.65bp 0.65bp 0.65bp 0.65bp,width=16mm]{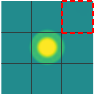}%
\hfill%
\includegraphics[trim=0.65bp 0.65bp 0.65bp 0.65bp,width=16mm]{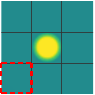}%
\hfill%
\includegraphics[trim=0.65bp 0.65bp 0.65bp 0.65bp,width=16mm]{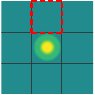}%
\hfill%
\includegraphics[trim=0.65bp 0.65bp 0.65bp 0.65bp,width=16mm]{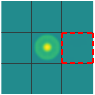}%
\hfill%
\includegraphics[trim=0.65bp 0.65bp 0.65bp 0.65bp,width=16mm]{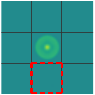}%
\hfill%
\includegraphics[trim=0.65bp 0.65bp 0.65bp 0.65bp,width=16mm]{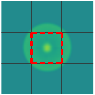}%
\hfill%
\includegraphics[trim=0.65bp 0.65bp 0.65bp 0.65bp,width=16mm]{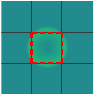}%
}\par
\vspace{1mm}
\noindent{\VisualizationLabelFont DINDiff}\par\nobreak
\noindent\hbox to\linewidth{%
\includegraphics[trim=0.65bp 0.65bp 0.65bp 0.65bp,width=16mm]{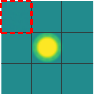}%
\hfill%
\includegraphics[trim=0.65bp 0.65bp 0.65bp 0.65bp,width=16mm]{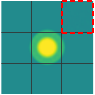}%
\hfill%
\includegraphics[trim=0.65bp 0.65bp 0.65bp 0.65bp,width=16mm]{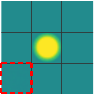}%
\hfill%
\includegraphics[trim=0.65bp 0.65bp 0.65bp 0.65bp,width=16mm]{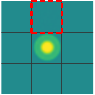}%
\hfill%
\includegraphics[trim=0.65bp 0.65bp 0.65bp 0.65bp,width=16mm]{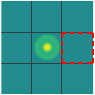}%
\hfill%
\includegraphics[trim=0.65bp 0.65bp 0.65bp 0.65bp,width=16mm]{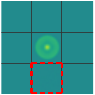}%
\hfill%
\includegraphics[trim=0.65bp 0.65bp 0.65bp 0.65bp,width=16mm]{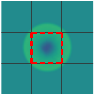}%
\hfill%
\includegraphics[trim=0.65bp 0.65bp 0.65bp 0.65bp,width=16mm]{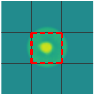}%
}\par
\endgroup
\caption{Shallow Water Block 8/9 reconstructions at $64\times64$, comparing Ours with DINDiff.}
\label{fig:pde-task1-sw-block8}
\end{figure}

\begin{figure}[!htbp]
\centering
\setlength{\abovecaptionskip}{6pt}
\begingroup
\setlength{\parskip}{0pt}
{\VisualizationLabelFont Navier--Stokes Block 8/9}\par\smallskip
\noindent{\VisualizationLabelFont Reference}\par\nobreak
\noindent\hbox to\linewidth{%
\includegraphics[trim=0.65bp 0.65bp 0.65bp 0.65bp,width=16mm]{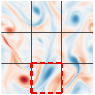}%
\hfill%
\includegraphics[trim=0.65bp 0.65bp 0.65bp 0.65bp,width=16mm]{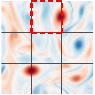}%
\hfill%
\includegraphics[trim=0.65bp 0.65bp 0.65bp 0.65bp,width=16mm]{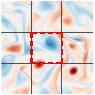}%
\hfill%
\includegraphics[trim=0.65bp 0.65bp 0.65bp 0.65bp,width=16mm]{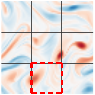}%
\hfill%
\includegraphics[trim=0.65bp 0.65bp 0.65bp 0.65bp,width=16mm]{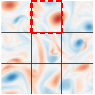}%
\hfill%
\includegraphics[trim=0.65bp 0.65bp 0.65bp 0.65bp,width=16mm]{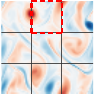}%
\hfill%
\includegraphics[trim=0.65bp 0.65bp 0.65bp 0.65bp,width=16mm]{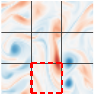}%
\hfill%
\includegraphics[trim=0.65bp 0.65bp 0.65bp 0.65bp,width=16mm]{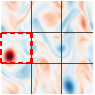}%
}\par
\vspace{1mm}
\noindent{\VisualizationLabelFont Ours}\par\nobreak
\noindent\hbox to\linewidth{%
\includegraphics[trim=0.65bp 0.65bp 0.65bp 0.65bp,width=16mm]{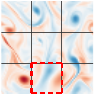}%
\hfill%
\includegraphics[trim=0.65bp 0.65bp 0.65bp 0.65bp,width=16mm]{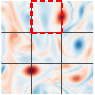}%
\hfill%
\includegraphics[trim=0.65bp 0.65bp 0.65bp 0.65bp,width=16mm]{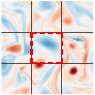}%
\hfill%
\includegraphics[trim=0.65bp 0.65bp 0.65bp 0.65bp,width=16mm]{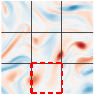}%
\hfill%
\includegraphics[trim=0.65bp 0.65bp 0.65bp 0.65bp,width=16mm]{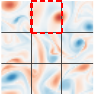}%
\hfill%
\includegraphics[trim=0.65bp 0.65bp 0.65bp 0.65bp,width=16mm]{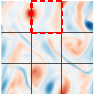}%
\hfill%
\includegraphics[trim=0.65bp 0.65bp 0.65bp 0.65bp,width=16mm]{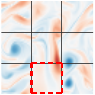}%
\hfill%
\includegraphics[trim=0.65bp 0.65bp 0.65bp 0.65bp,width=16mm]{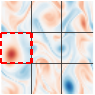}%
}\par
\vspace{1mm}
\noindent{\VisualizationLabelFont AmbientDiff}\par\nobreak
\noindent\hbox to\linewidth{%
\includegraphics[trim=0.65bp 0.65bp 0.65bp 0.65bp,width=16mm]{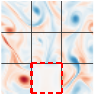}%
\hfill%
\includegraphics[trim=0.65bp 0.65bp 0.65bp 0.65bp,width=16mm]{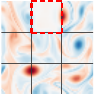}%
\hfill%
\includegraphics[trim=0.65bp 0.65bp 0.65bp 0.65bp,width=16mm]{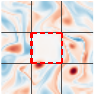}%
\hfill%
\includegraphics[trim=0.65bp 0.65bp 0.65bp 0.65bp,width=16mm]{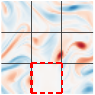}%
\hfill%
\includegraphics[trim=0.65bp 0.65bp 0.65bp 0.65bp,width=16mm]{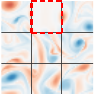}%
\hfill%
\includegraphics[trim=0.65bp 0.65bp 0.65bp 0.65bp,width=16mm]{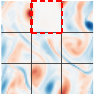}%
\hfill%
\includegraphics[trim=0.65bp 0.65bp 0.65bp 0.65bp,width=16mm]{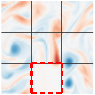}%
\hfill%
\includegraphics[trim=0.65bp 0.65bp 0.65bp 0.65bp,width=16mm]{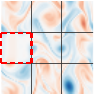}%
}\par
\endgroup
\caption{Navier--Stokes Block 8/9 reconstructions at $64\times64$, comparing Ours with AmbientDiff.}
\label{fig:pde-task1-ns-block8}
\end{figure}

Figures~\ref{fig:pde-task1-sw-block3}--\ref{fig:pde-task1-ns-block8} show Shallow Water and Navier--Stokes reconstructions under block observations. Solid lines mark the $3\times3$ acquisition partition, and red dashed outlines identify unsupplied blocks.

\FloatBarrier

\FloatBarrier

\end{document}